%% file: main.tex
\documentclass[10pt,twocolumn,letterpaper]{article}

\usepackage[pagenumbers]{wacv} 
\usepackage[accsupp]{axessibility}
\usepackage{amsmath,amssymb,amsfonts}
\usepackage{algorithmic}
\usepackage{graphicx}
\usepackage{textcomp}
\usepackage{xcolor}
\usepackage{microtype}
\usepackage{booktabs}
\usepackage{multirow}
\usepackage{subcaption}
\usepackage[table]{xcolor} 
\usepackage{array}
\usepackage{url}
\usepackage{listings} 
\graphicspath{{figures/}}
\DeclareGraphicsExtensions{.pdf,.png,.jpg}

\renewcommand{\arraystretch}{1.05}
\newcolumntype{L}[1]{>{\raggedright\arraybackslash}p{#1}}
\newcolumntype{C}[1]{>{\centering\arraybackslash}p{#1}}
\newcolumntype{R}[1]{>{\raggedleft\arraybackslash}p{#1}}

\definecolor{Accent1}{HTML}{1F77B4}
\definecolor{Accent2}{HTML}{FF7F0E}
\definecolor{Accent3}{HTML}{2CA02C}
\definecolor{Accent4}{HTML}{D62728}
\definecolor{Accent5}{HTML}{9467BD}

\definecolor{rscolor}{HTML}{FC8E62} 
\definecolor{oscolor}{HTML}{95EC69} 
\definecolor{generalcolor}{HTML}{0185F9} 

\newcommand{\rsb}{\textcolor{rscolor}{$\bullet$\,}}
\newcommand{\osb}{\textcolor{oscolor}{$\bullet$\,}}
\newcommand{\gb}{\textcolor{generalcolor}{$\bullet$\,}}

\definecolor{blockbg}{RGB}{245,245,245}

\definecolor{wacvblue}{rgb}{0.21,0.49,0.74}
\usepackage[pagebackref,breaklinks,colorlinks,allcolors=wacvblue]{hyperref}

\lstdefinestyle{geocode}{
  basicstyle=\ttfamily\footnotesize,
  breaklines=true,
  breakatwhitespace=false,
  columns=fullflexible,
  keepspaces=true,
  showstringspaces=false,
  tabsize=2,
  frame=single,
  aboveskip=4pt,
  belowskip=4pt,
  postbreak=\mbox{\textcolor{wacvblue}{\scriptsize$\hookrightarrow$}\space}
}
\definecolor{darkgreen}{RGB}{0,100,0} 

\def\wacvPaperID{2684} 
\def\confName{WACV}
\def\confYear{2026}

\title{Geo3DVQA: Evaluating Vision-Language Models for 3D Geospatial Reasoning from Aerial Imagery}

\author{%
  Mai Tsujimoto$^{1}$ \qquad
  Junjue Wang$^{1}$ \qquad
  Weihao Xuan$^{1,2}$ \qquad
  Naoto Yokoya$^{1,2,\dagger}$%
  \\
  \text{\normalsize $^{1}$The University of Tokyo \qquad
  $^{2}$RIKEN AIP \qquad
  $\dagger$Corresponding author}
}

\begin{document}
\maketitle
\begin{abstract}
  Three-dimensional geospatial analysis is critical for applications in urban planning, climate adaptation, and environmental assessment. Current methodologies depend on costly specialized sensors (e.g., LiDAR and multispectral), which restrict global accessibility. Existing sensor-based and rule-driven methods further struggle with tasks requiring the integration of multiple 3D cues, handling diverse queries, and providing interpretable reasoning. We hereby present Geo3DVQA, a comprehensive benchmark to evaluate vision-language models (VLMs) in height-aware 3D geospatial reasoning using RGB-only remote sensing imagery.  Unlike conventional sensor-based frameworks, Geo3DVQA emphasizes realistic scenarios that integrate elevation, sky view factors, and land cover patterns. The benchmark encompasses 110k curated question–answer pairs spanning 16 task categories across three complexity levels: single-feature inference, multi-feature reasoning, and application-level spatial analysis. The evaluation of ten state-of-the-art VLMs highlights the difficulty of RGB-to-3D reasoning. GPT-4o and Gemini-2.5-Flash achieve around 30\% overall short-answer accuracy, whereas domain-specific fine-tuning of Qwen2.5-VL-7B substantially improves performance to nearly 50\%, representing a gain of over 20 percentage points compared to the base model. Geo3DVQA thus introduces a new frontier for scalable, accessible, and interpretable 3D geospatial analyses. The code and data will be released upon publication at \url{https://github.com/mm1129/Geo3DVQA}.
  \end{abstract}

\input{sections/introduction}
\input{sections/related_work_re}
\input{sections/dataset}
\input{sections/benchmark_experiments}

\input{sections/conclusion}

\section*{Acknowledgements}
This work was supported by JST NEXUS, Japan Grant Number JPMJNX25CA.
Weihao Xuan is supported by the RIKEN Junior Research Associate (JRA) program.

\bibliographystyle{ieeetr}
\bibliography{references}

\subsection*{Author Note (v2)}
We identified a minor inconsistency in the GeoNRW land cover class name mapping used in our land cover related evaluation, which only affected the land cover type and land use/land cover question categories.
This version corrects the mapping and clarifies the evaluation policy for land cover related tasks, and all affected short-answer results were re-evaluated and updated accordingly (Appendix Table~\ref{tab:shortanswer_v1_v2_summary}).
These corrections slightly affected the numerical results but did not alter the main conclusions or overall trends.

\appendix
\input{sections/appendix2_reorganized}

\end{document}

%% file: sections/introduction.tex
\section{Introduction}
\begin{figure*}[!t]
        \centering
        \includegraphics[width=0.82\textwidth]{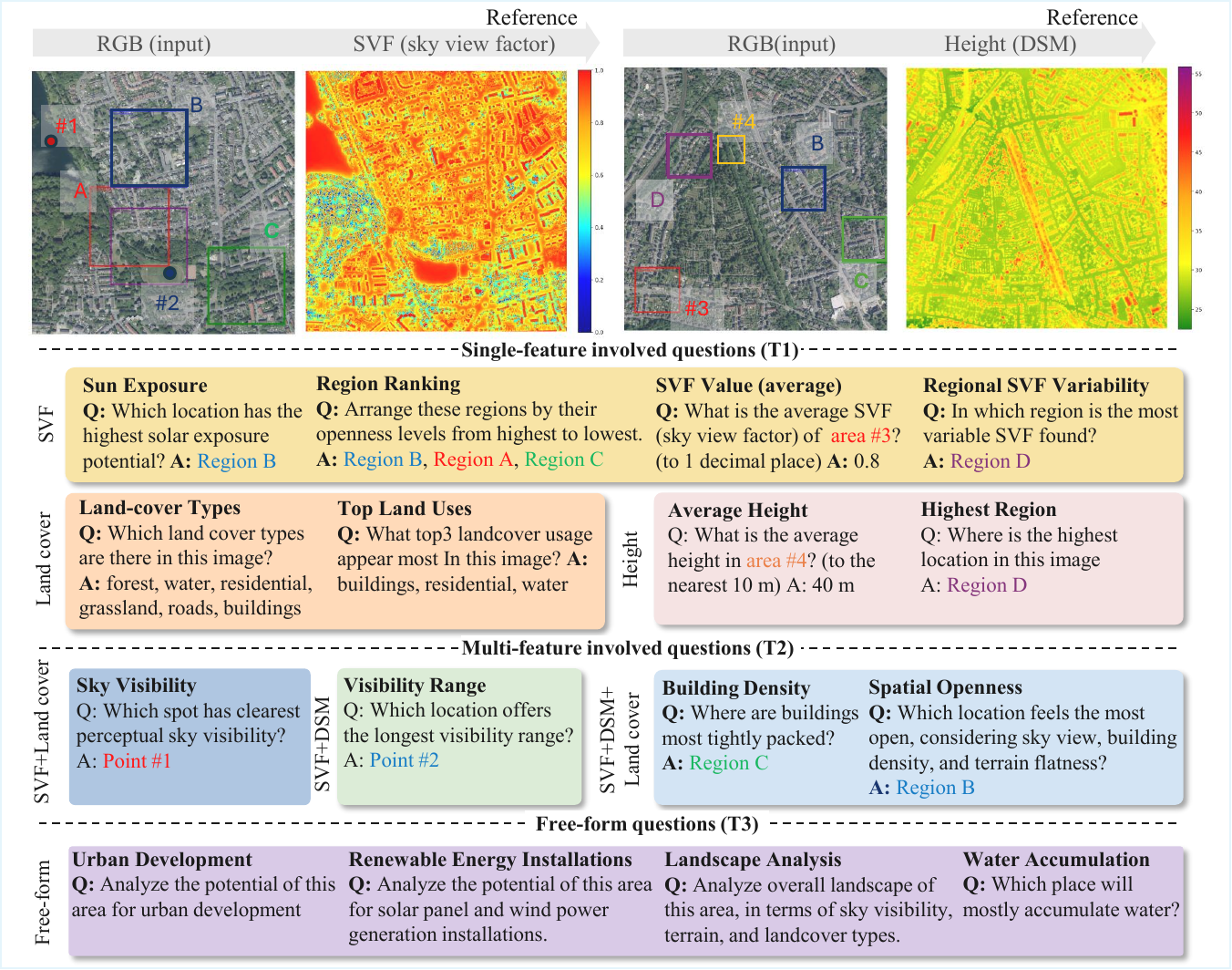} 
        \caption{\textbf{Geo3DVQA Task Taxonomy and Evaluation Framework.} 
        Our benchmark comprises 16 task categories organized into three complexity tiers to evaluate RGB-to-3D spatial reasoning capabilities.
        \textbf{Tier 1 (Single-feature)}: Direct inference of individual properties across SVF, land cover, and height from RGB patterns. 
        \textbf{Tier 2 (Multi-feature)}: Composite reasoning tasks requiring the integration of multiple spatial attributes (e.g., sky visibility combines SVF with building penalties).
        \textbf{Tier 3 (Application-level)}: Free-form reasoning for real-world applications in urban planning, renewable energy, landscape analysis, and water accumulation. 
        Ground truth generation uses multimodal reference data (SVF, DSM, land cover), whereas the models are evaluated using only RGB imagery. 
        The colored regions indicate the locations of multiple-choice options, with answers highlighted for visualization.}
        \label{fig:qa_categories}
    \end{figure*}

Recent advancements in vision-language models (VLMs)~\cite{liu2023visual,li2024llava,bai2025qwen2} have exhibited noteworthy capabilities in two-dimensional (2D) visual understanding. However, their potential for three-dimensional (3D) spatial reasoning from widely available RGB imagery remains largely unexplored. This gap is particularly important given the growing importance of 3D geospatial analysis in addressing contemporary global challenges.

The Intergovernmental Panel on Climate Change (IPCC) projects an increasing frequency and intensity of climate extremes under global warming scenarios~\cite{arias2021climate}. These risks are particularly amplified in urban areas, where three-dimensional morphology drives local microclimates and intensifies Urban Heat Island effects~\cite{unger2004intra,middel2018sky}. Accurate assessment of spatial properties, such as sky view factors, building heights, and urban morphology, is essential for applications ranging from heat mitigation to disaster risk assessment, energy planning, and quality-of-life improvements.

Nevertheless, current approaches to three-dimensional geospatial analyses face significant accessibility barriers. Traditional methods depend on costly specialized sensors such as LiDAR (often costing \$100{,}000 or more) and multispectral sensors~\cite{Li2020Automatic,Lu2021SGTBN,Takhtkeshha2024Multispectral}. Beyond acquisition costs, these approaches require considerable technical expertise for sensor operation, data processing workflows, and cross-feature integration. These limitations restrict advanced geospatial reasoning to well-funded institutions, creating a substantial accessibility gap in areas where climate adaptation analyses are most urgently needed.

Conversely, aerial RGB imagery is widely accessible through platforms such as Google Earth and national mapping agencies at minimal or no cost. This contrast naturally leads to the question: \emph{Can vision-language models perform sophisticated 3D spatial reasoning directly from ubiquitous RGB imagery alone?}
We studied their ability to learn and infer imperceptible 3D properties directly from the observable RGB patterns. Among these properties, the Sky View Factor (SVF) serves as a fundamental metric: it represents the ratio (0--1) of the visible sky from a point and is essential for urban microclimate and solar radiation analysis~\cite{lindberg2018umep}.

To advance the development of VLMs with robust 3D spatial reasoning, we introduce \textbf{Geo3DVQA}, a dedicated VQA (Visual Question Answering) benchmark specifically designed to evaluate height-aware spatial reasoning from aerial RGB imagery alone.
As shown in Figure~\ref{fig:qa_categories}, our framework systematically evaluates the models' ability to extract 3D spatial properties from 2D visual cues. Here we employed a concise three-tier taxonomy (single-feature, multi-feature, and application-level). 

Geo3DVQA focuses on coarse, decision-oriented, height-aware reasoning (e.g., relative comparisons and bin-level distinctions). It is designed to be realistic for RGB-only inputs where precise 3D recovery from monocular RGB is inherently ill-posed, while coarse height distinctions remain feasible through learned 2D cues, such as shadows, occlusions, and perspective~\cite{Cai2020Monocular,Lu2021SGTBN,Lourenço2021Intel,liu2020im2elevation,liasis2016satellite}.

Our main contributions are as follows:\\
1. We introduce Geo3DVQA, a dedicated benchmark for RGB-to-3D geospatial reasoning with a clear three-tier taxonomy, containing 110k carefully constructed question–answer pairs across \textbf{16} task categories. \\ 
2. We systematically evaluated ten leading VLMs and revealed fundamental limitations in height-aware spatial reasoning under RGB-only inputs. \\
3. 
We established baselines via domain-specific instruction tuning, achieving +22.3 points in overall accuracy with improvements in height and SVF tasks.

%% file: sections/related_work_re.tex
\section{Related Work}
\label{sec:related_work}

\subsection{Vision-Language Models for Remote Sensing and the Height-Aware Gap}
Recent advances in vision-language models (VLMs) have enabled strong visual understanding and natural language reasoning across general-purpose tasks~\cite{liu2023visual,li2024llava,bai2025qwen2}.
In remote sensing, large-scale resources have catalyzed progress in captioning, grounding, and VQA research. RSVQA~\cite{lobry2020rsvqa} pioneered VQA (visual question answering) from RS data using OSM-derived (OpenStreetMap) supervision, HRVQA~\cite{li2024hrvqa} scaled VQA to high-resolution imagery, and VRSBench~\cite{li2024vrsbench} broadened evaluations beyond VQA to detailed captioning and visual grounding. RSIEval (RSGPT)~\cite{hu2025rsgpt} further supports evaluation of rich remote-sensing captions. DisasterM3~\cite{wang2025disasterm3} targets bi-temporal disaster assessment with optical/SAR pairs, DynamicVL~\cite{xuan2025dynamicvl} benchmarks MLLMs for long-term multi-temporal city understanding, and TEOChat~\cite{irvin2024teochat} extends to temporal instruction following.

Systems such as GeoChat~\cite{kuckreja2024geochat} and GEOBench-VLM~\cite{Danish_2025_ICCV} standardize RS instruction following and evaluation, and EarthVQA~\cite{wang2024earthvqa} addresses 2D geospatial relational reasoning.
Despite these advances, existing datasets and models primarily focus on 2D recognition and relational reasoning rather than 3D remote sensing inference from 2D RGB inputs. This creates a significant gap in geospatial understanding, a challenge our study seeks to address.

\begin{table}[t]
\centering
\resizebox{\linewidth}{!}{
\begin{tabular}{l l r}
\toprule
Dataset & Primary Focus & Scale \\
\midrule
RSVQA~\cite{lobry2020rsvqa} & VQA with OSM supervision (2D) & --\\
HRVQA~\cite{li2024hrvqa} & High-resolution VQA (2D) & 1.07M \\
VRSBench~\cite{li2024vrsbench} & Versatile VL tasks (2D) & 123K \\
RSIEval (RSGPT) & RS-VLM caption evaluation (2D) & -- \\
GeoChat~\cite{kuckreja2024geochat} & Unified instruction following & 318K \\
TEOChat~\cite{irvin2024teochat} & Temporal EO understanding & 554K \\
DisasterM3~\cite{wang2025disasterm3} & Disaster assessment (bi-temporal) & 123K \\
\midrule
Geo3DVQA (Ours) & \textbf{Height-aware 3D inference (RGB-only)} & 110K QA \\
\bottomrule
\end{tabular}}
\caption{Representative remote sensing benchmarks and our focus.}
\end{table}

\subsection{Inferring 3D Properties from RGB}
For monocular height estimation, a rich line of work estimates the height or DSM from single-view imagery.
IM2HEIGHT~\cite{mou2018im2height} pioneered learning-based height regression from monocular images, while subsequent approaches leveraged shadow geometry and sun elevation for building height estimation~\cite{liasis2016satellite} and extended to IM2ELEVATION-style pipelines for DSM/height prediction~\cite{liu2020im2elevation}.
Most of these methods target continuous value map generation with explicit height supervision.

The SVF can be delivered either from DSM/LiDAR or street-view fisheye imagery with semantic segmentation~\cite{lindberg2018umep,liang2017automatic}.
Estimating continuous SVF from RGB data alone is a challenge. Some attempts have been made to use satellite shadow cues~\cite{hodul2016estimation}, but robust inference generally relies on 3D data.
Furthermore, practical applications require an integrated understanding of multiple spatial properties beyond isolated SVF estimation. This need motivated our investigation into VLMs' ability to infer height-aware properties from RGB patterns. Our evaluation bridges 2D semantic understanding and 3D geometric reasoning through a language-based assessment without specialized sensors.

%% file: sections/dataset.tex
\section{Geo3DVQA Dataset}
\label{sec:dataset}

\vspace{-0.1cm}
\subsection{Dataset Construction and Composition}

Geo3DVQA is built on GeoNRW~\cite{baier2020geonrw}, which provides orthorectified aerial imagery, LiDAR elevation models, and semantic segmentation for North Rhine-Westphalia, Germany. The GeoNRW dataset contains 7,783 image triplets that were curated at 1~m spatial resolution, covering 10 land cover classes.

\begin{figure}[!t]
    \centering
    \includegraphics[width=\columnwidth]{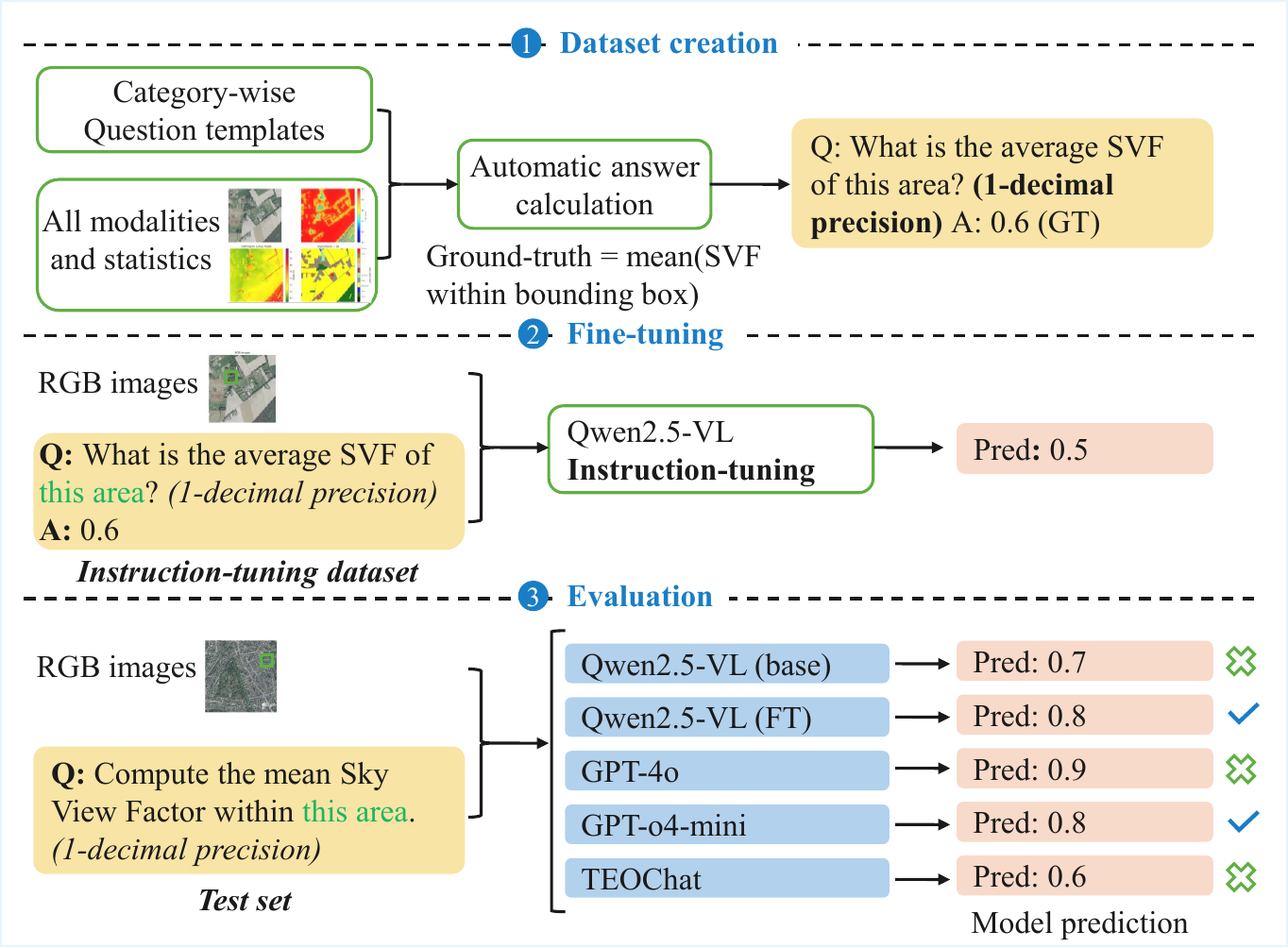}
    \caption{\textbf{Geo3DVQA Dataset Construction and Evaluation Pipeline.} (1) Dataset Creation: Question templates were generated for each category, with ground-truth answers automatically derived from multimodal data (SVF, DSM, land cover) and their statistics. (2) Fine-tuning: Qwen2.5-VL learns height-aware spatial reasoning from RGB inputs using the constructed dataset. (3) Evaluation: Models are tested using only RGB inputs, with predictions compared against the ground truth. The fine-tuned model leverages domain knowledge from step (2).}
    \label{fig:dataset_overview}
\end{figure}

\noindent \textbf{Spatial data processing.} Digital Surface Models (DSM) provide elevation information vital for spatial analysis~\cite{Guth2021Digital,Algancı2018Accuracy}. From the DSM data, we computed the SVF values using the Urban Multi-scale Environmental Predictor (UMEP) toolkit with ray casting~\cite{lindberg2018umep}. This approach is widely adopted in urban climate analysis and demonstrates strong agreement with fisheye imagery and high-resolution 3D models (R$^2{>}0.91$, RMSE $\approx$ 0.07)~\cite{Jiao2019Evaluation,dirksen2019sky,liang2017automatic}. Our processing pipeline (Fig.~\ref{fig:dataset_overview}) aligned these multimodal reference data (i.e., SVF, elevation, and segmentation data), extracted spatial indicators, and generated question–answer pairs.

Although the SVF mainly quantifies vertical openness, it can overestimate perceived openness because horizontal obstructions are not fully captured~\cite{miao2020review,middel2018sky,dirksen2019sky}. To address this limitation, we employed a composite \emph{sky visibility} metric that integrates SVF with land cover information, providing a more realistic representation of perceived openness~\cite{Daramola2019Analysis, Xia2021Sky, bradley2001method}. The reliability of these metrics and the derived results was checked via statistical validation and multimodal visual inspection.

\begin{figure*}[!t]
    \includegraphics[width=0.9\textwidth]{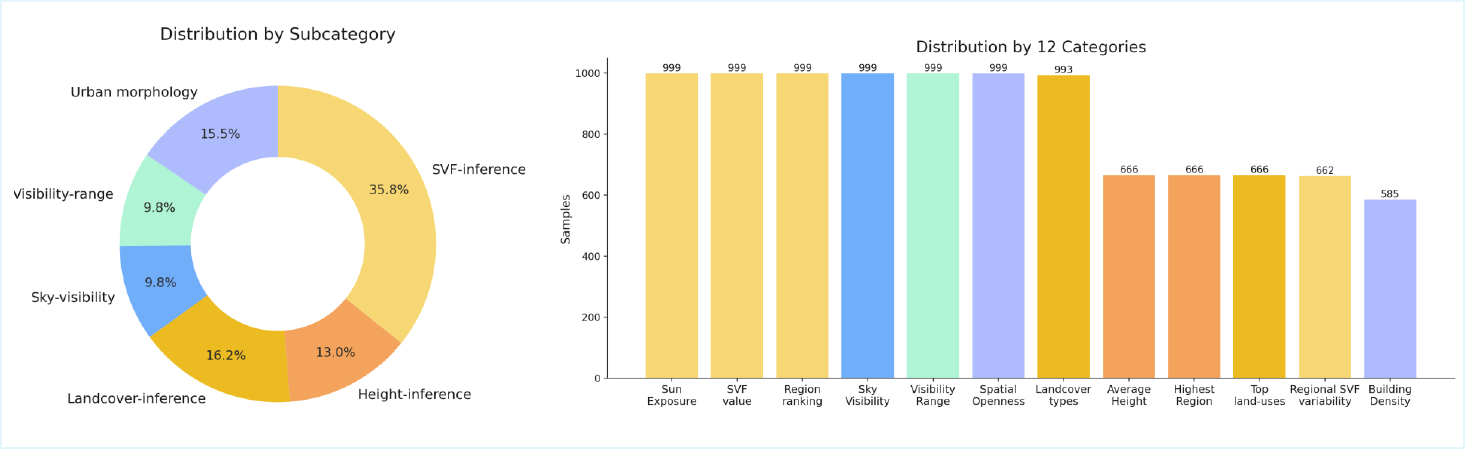}
    \caption{Task composition of Geo3DVQA evaluation set (left). Subcategory-level distribution (right panel). Distribution across 12 task categories. Tier 3 was excluded (116 questions in total).} 
    \label{fig:dataset_semantic_stats}
\end{figure*}

Geo3DVQA comprises 110k QA pairs across 16 task categories. The question formats included multiple-choice, ranking, numerical estimation, and open-ended questions. Quality control ensures balanced answer distributions and spatial diversity (Figure.~\ref{fig:dataset_semantic_stats} and Appendix Fig.~\ref{fig:dataset_statistics}).

\noindent\textbf{Task Design and Generation.}
We organize tasks into three tiers. \textit{T1} focuses on single-feature inference directly computable from a
single modality: SVF inference (sun exposure, region ranking, regional SVF variability, mean SVF), segmentation inference (land cover type, land use), and DSM inference (height average, highest region). \textit{T2} requires multi-feature reasoning that integrates SVF, DSM, and land cover to produce a unified score for sky visibility, spatial openness, building density, and visibility range. Scoring adheres to established geospatial analysis methodologies from prior studies (e.g., ~\cite{miao2020review,bradley2001method,Daramola2019Analysis,Jaehyun_Ha_2016, li2020two, hoechstetter2008effects}). For instance, the visibility range is computed from the viewshed, SVF, and terrain to balance sightline reach, sky access, and terrain blocking. Spatial openness combines SVF statistics, terrain flatness, and the complement of building density, which together reflect sky access, gentle relief, and low obstructions. The full details and scientific rationale for each metric, as well as the sensitivity study for weights and prompts, are provided in Appendix~\ref{sec:appendix_scientific_rationale}, Appendix~\ref{sec:appendix_weight_sensitivity}, and Appendix~\ref{sec:appendix_prompt_paraphrasing}. 

 \textit{T3} covers application-oriented free-form analyses (urban development, renewable energy, landscape, water), where answers are generated from statistics distilled from the reference modalities for QA construction.
During the evaluation, the models received only RGB, and spatial queries used normalized coordinates (x\%, y\%) following TEOChat~\cite{irvin2024teochat}. 

\subsection{Dataset Creation Pipeline}

\textbf{Short-answer (T1/T2).} After preparing the required inputs (SVF, DSM, and segmentation) and passing suitability checks, we generated questions from category-specific templates. Correct answers and distractors were sampled from a shared candidate pool to ensure balanced scores and spatial diversity. Option labels were randomized to avoid bias, and sampled Q\&As were manually verified for correctness and scientific validity.

\textbf{Free-form (T3).} We compiled scene-level statistics from the reference modalities and used them to generate answers to rubric-constrained prompts with GPT-4.1-mini. The drafts were automatically checked against the statistics and then reviewed by human annotators, and only validated pairs were retained in the evaluation set. The templates use a structured format for observations that describe SVF, elevation, and land cover, as well as conclusions that provide actionable recommendations. Training and test separation are enforced by strict geographic splits.

\begin{figure*}[t]
    \centering
    \includegraphics[width=\linewidth]{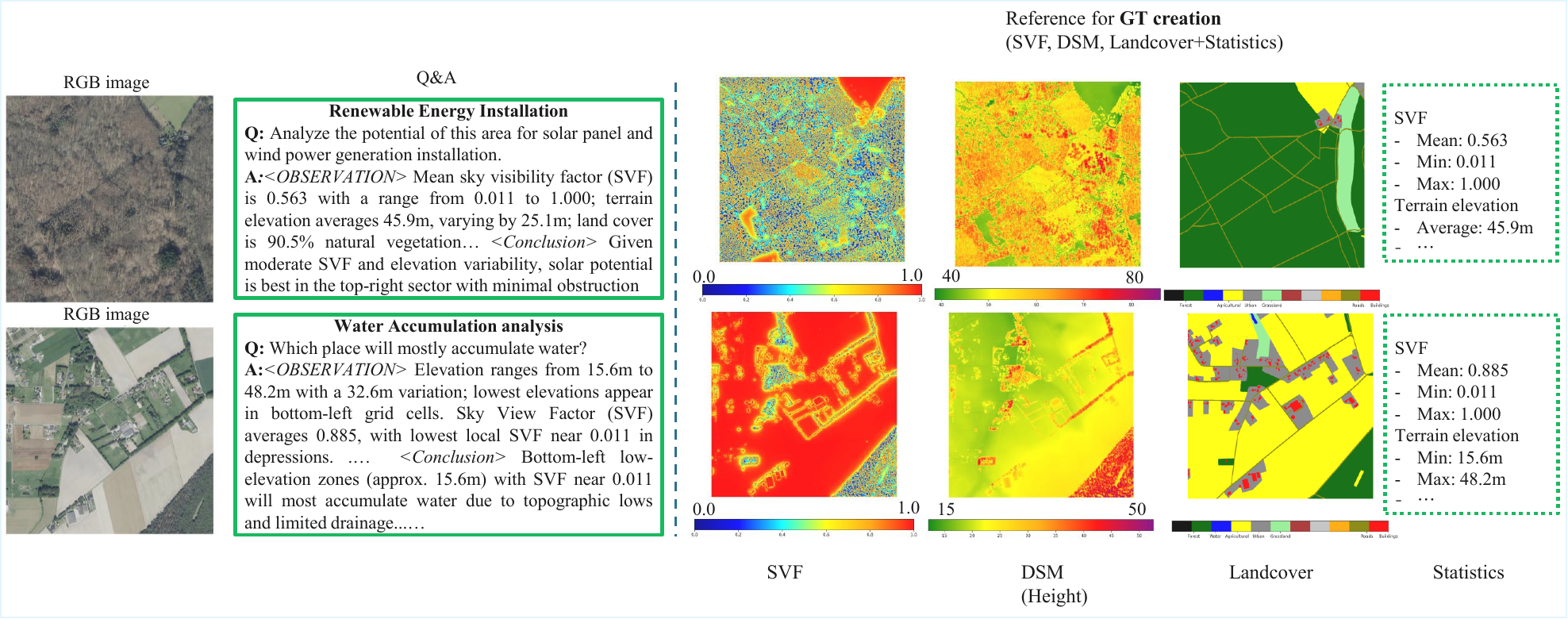}
    \vspace{-1em}
    \caption{The shortened examples of Tier-3 free-form Q\&A with \textbf{RGB input}  and reference for ground truth. Statistics were first calculated from the reference modalities (SVF, DSM, segmentation) and input into GPT-4.1-mini to generate answers corresponding to pre-defined categories. These answers were validated by cross-checking them against the reference statistics using GPT-5.  Q\&As that passed the validation are then verified by humans, and the final Q\&A pairs are used for evaluation.}
    \label{fig:freeform_examples_inputs_and_modalities}
\end{figure*}

Fig.~\ref{fig:freeform_examples_inputs_and_modalities} shows open-ended tasks where renewable energy assessments combine sky view factors, elevation, and land cover, whereas water accumulation focuses on low-elevation, low-SVF regions. Additional visualizations of the Tier-3 answer space, including a word cloud, are provided in Appendix~\ref{sec:appendix_freeform_eval}.

\vspace{0.2em}
\noindent \textit{Supporting details.} Appendix~\ref{sec:appendix_scientific_rationale} provides the scientific rationale for each metric, and Appendix~\ref{sec:appendix_eval_prompts} details the task taxonomy, coordinate system, and prompt design used for QA construction. Appendix~\ref{sec:appendix_dataset_statistics} reports additional dataset statistics and near-equal label frequencies for each option.

%% file: sections/benchmark_experiments.tex
\begin{table*}[!t]
    \centering
    \begingroup\setlength{\tabcolsep}{4.5pt}\renewcommand{\arraystretch}{0.9}\setlength{\aboverulesep}{0.6ex}\setlength{\belowrulesep}{0.6ex}\setlength{\abovetopsep}{0.4ex}\setlength{\belowbottomsep}{0.4ex}\small
    \begin{tabular}{@{}l|ccccc|cc@{}}
    \toprule
    \multirow{2}{*}{\textbf{Model}} & \multicolumn{5}{c|}{\textbf{Short-answer (\%)}} & \multicolumn{2}{c}{\textbf{Free-form (1--5)}} \\
    \cmidrule(lr){2-6} \cmidrule(lr){7-8}
    & \textbf{SVF} & \textbf{Height} & \textbf{Land Use/LC} & \textbf{Multi} & \textbf{Overall} & \textbf{Total} & \textbf{Conclusion} \\
    \midrule
    \multicolumn{1}{l|}{\textit{\gb Commercial models}} & & & & & & & \\
    *o4-mini & 20.2 & 11.7 & 23.4 & 23.7 & 20.5 & 2.05 & 2.41 \\
    GPT-4o & 30.7 & 27.8 & 32.3 & 32.4 & 31.0 & 2.27 & 2.57 \\
    GPT-4.1-mini & 28.5 & 23.7 & 43.3 & 31.5 & 31.0 & 2.53 & 3.02 \\
    Gemini-2.5-Flash & 32.9 & 19.4 & 45.9 & 35.6 & 33.8 & 1.96 & 2.34 \\
    \midrule
    \multicolumn{1}{l|}{\textit{\rsb Remote sensing VLMs}} & & & & & & & \\
    TEOChat & 17.5 & 21.4 & 10.1 & 31.5 & 20.3 & 1.45 & 1.66 \\
    GeoChat & 20.2 & 21.0 & 26.5 & 25.4 & 22.6 & 1.15 & 1.67 \\
    \midrule
    \multicolumn{1}{l|}{\textit{\osb Open-source models}} & & & & & & & \\
    LLaVA-OneVision & 22.4 & 24.5 & 26.1 & 19.3 & 22.4 & 1.48 & 2.01 \\
    InternVL3-8B & 21.8 & 24.9 & 32.9 & 22.0 & 24.0 & 2.22 & 2.62 \\
    Qwen2.5-VL-3B & 21.3 & 22.5 & 30.2 & 24.4 & 23.6 & 1.38 & 2.11 \\
    Qwen2.5-VL-7B Base & 23.5 & 21.3 & 45.3 & 22.6 & 26.4 & 2.04 & 2.23 \\
    \midrule
    \multicolumn{1}{l|}{\textit{\textbf{Fine-tuned models}}} & & & & & & & \\
    Qwen2.5-VL-7B FT (10K) & 34.24 & 33.56 & 65.1 & 39.30 & 40.4 & 2.69 & 3.01 \\
    Qwen2.5-VL-7B FT (100K) & \underline{\textbf{44.61}} & \underline{\textbf{41.07}} & \underline{\textbf{71.3}} & \underline{\textbf{45.45}} & \underline{\textbf{48.7}} & \underline{\textbf{2.89}} & \underline{\textbf{3.11}} \\
    \textit{(FT $-$ Base)} & \textcolor{green}{$\uparrow$21.1} & \textcolor{green}{$\uparrow$19.8} & \textcolor{green}{$\uparrow$26.0} & \textcolor{green}{$\uparrow$22.9} & \textcolor{green}{$\uparrow$22.3} & \textcolor{green}{$\uparrow$0.85} & \textcolor{green}{$\uparrow$0.88} \\
    \midrule
    \end{tabular}
    \endgroup
    \caption{Aggregated Evaluation Across Major Categories. Accuracy (\%) and free-form scores (1--5 scale) in the last two columns. SVF, Height, and Land Use/LC are single-feature inference (T1). Multi represents multi-feature inference (T2). Overall covers all short-answer categories (T1+T2). Free-form evaluation (T3) uses rubric-based scoring for structured-reasoning tasks. FT(10 K) and FT(100 K) in the result tables denote fine-tuning with 10 K and 100 K instruction-tuning samples, respectively. `FT--Base' indicates the performance improvements achieved by fine-tuning Qwen-2.5-VL on 100K Q\&As (`Qwen2.5-VL-7B FT (100K)') compared to the baseline results (`Qwen2.5-VL-7B Base'). \\ 
    \footnotesize *Format errors (repeated questions without answers): o4-mini 47.2\%, Gemini-2.5-Flash 16.3\%, others $<$2\%.}
    \label{tab:agg_eval_new_transposed}
    
\end{table*}
\begin{table*}[tb]
    \centering
    \footnotesize
    \begingroup\setlength{\tabcolsep}{4pt}\renewcommand{\arraystretch}{0.85}\setlength{\aboverulesep}{0.4ex}\setlength{\belowrulesep}{0.4ex}\setlength{\abovetopsep}{0.3ex}\setlength{\belowbottomsep}{0.3ex}
    \begin{tabular}{l|cccc|cc}
    \toprule
     & \multicolumn{4}{c|}{\textbf{SVF-aware}} & \multicolumn{2}{c}{\textbf{Height-aware}} \\
    \cmidrule(lr){2-5} \cmidrule(lr){6-7}
    \textbf{Model} & \textbf{SVF val.} & \textbf{Reg. rank} & \textbf{SVF var.} & \textbf{Sun exp.} & \textbf{Hgt. avg.} & \textbf{Highest} \\
    \midrule
    \multicolumn{1}{l|}{\textit{\gb Commercial models}} & & & & & & \\
    GPT-4o & 11.7 & 30.6 & 36.7 & 35.0 & 26.4 & 29.1 \\
    GPT-4.1-mini & 1.9 & 33.8 & 37.3 & 31.3 & 18.9 & 28.4 \\
    Gemini-2.5-Flash & 7.2 & 36.2 & 37.3 & 39.9 & 9.0 & 29.9 \\
    \midrule
    \multicolumn{1}{l|}{\textit{\rsb Remote sensing VLMs}} & & & & & & \\
    TEOChat & 0.7 & 15.1 & 34.4 & 22.4 & 18.9 & 23.9 \\
    GeoChat & 6.5 & 16.7 & 35.7 & 25.4 & 17.3 & 24.6 \\
    \midrule
    \multicolumn{1}{l|}{\textit{\osb Open-source models}} & & & & & & \\
    LLaVA-OneVision & 3.9 & 18.4 & 33.4 & 27.2 & 16.8 & 32.1 \\
    InternVL3-8B & 2.9 & 19.0 & 33.5 & 26.3 & 18.5 & 31.2 \\
    Qwen2.5-VL-3B & 5.5 & 19.3 & 32.2 & 23.2 & 18.0 & 27.0 \\
    Qwen2.5-VL-7B Base & 6.8 & 19.5 & 34.0 & 28.6 & 9.9 & 32.7 \\
    \midrule
    \multicolumn{1}{l|}{\textit{\textbf{Fine-tuned models}}} \\
    Qwen2.5-VL-7B FT (10K) & 27.9 & 29.6 & 39.4 & 35.3 & 34.8 & 32.3 \\
    Qwen2.5-VL-7B FT (100K) & \underline{\textbf{42.7}} & \underline{\textbf{39.7}} & \underline{\textbf{51.4}} & \underline{\textbf{43.9}} & \underline{\textbf{42.2}} & \underline{\textbf{39.9}} \\
    \textit{(FT $-$ Base)} & \textcolor{green}{$\uparrow$35.9} & \textcolor{green}{$\uparrow$20.2} & \textcolor{green}{$\uparrow$17.4} & \textcolor{green}{$\uparrow$15.3} & \textcolor{green}{$\uparrow$32.3} & \textcolor{green}{$\uparrow$7.2} \\
    \bottomrule
    \end{tabular}%
    \endgroup
    \caption{Category-wise Accuracy (Part I): SVF-aware and Height-aware categories. Abbreviations: SVF val. = SVF value, Reg. rank = region ranking, SVF var. = regional SVF variability, Sun exp. = sun exposure, Hgt. avg. = height average, Highest = highest region \\
    \footnotesize \emph{Note.} Part II (Table~\ref{tab:category_acc_b}) continues with land cover and multi-factor categories.}
    \label{tab:category_acc_a}
\end{table*}

\begin{table*}[tb]
    \centering
    \footnotesize
    \begingroup\setlength{\tabcolsep}{4pt}\renewcommand{\arraystretch}{0.85}\setlength{\aboverulesep}{0.4ex}\setlength{\belowrulesep}{0.4ex}\setlength{\abovetopsep}{0.3ex}\setlength{\belowbottomsep}{0.3ex}
    \begin{tabular}{l|cc|cccc}
    \toprule
     & \multicolumn{2}{c|}{\textbf{Land cover}} & \multicolumn{4}{c}{\textbf{Multi-factor}} \\
    \cmidrule(lr){2-3} \cmidrule(lr){4-7}
    \textbf{Model} & \textbf{Top uses} & \textbf{LC type} & \textbf{Spatial open.} & \textbf{Sky vis.} & \textbf{Bldg. dens.} & \textbf{Vis. range} \\
    \midrule
    \multicolumn{1}{l|}{\textit{\gb Commercial models}} & & & & & & \\
    GPT-4o & 39.9 & 27.2 & 41.4 & 36.3 & 39.7 & 24.2 \\
    GPT-4.1-mini & 29.1 & 52.8 & 41.2 & 32.8 & 43.8 & 22.9 \\
    Gemini-2.5-Flash & 29.6 & 55.5 & 45.1 & 38.1 & 46.3 & 26.8 \\
    \midrule
    \multicolumn{1}{l|}{\textit{\rsb Remote sensing VLMs}} & & & & & & \\
    TEOChat & 24.8 & 0.3 & 20.3 & 38.8 & 28.2 & 26.0 \\
    GeoChat & 24.8 & 27.6 & 22.0 & 24.0 & 26.3 & 26.1 \\
    \midrule
    \multicolumn{1}{l|}{\textit{\osb Open-source models}} & & & & & & \\
    LLaVA-OneVision & 28.4 & 24.9 & 32.7 & 11.6 & 31.5 & 19.9 \\
    InternVL3-8B & 18.6 & 41.2 & 31.2 & 15.8 & 31.8 & 22.4 \\
    Qwen2.5-VL-3B & 27.5 & 31.6 & 30.1 & 19.7 & 32.5 & 24.4 \\
    Qwen2.5-VL-7B Base & 29.1 & 55.3 & 32.2 & 13.1 & 39.8 & 22.0 \\
    \midrule
    \multicolumn{1}{l|}{\textit{\textbf{Fine-tuned models}}} \\
    Qwen2.5-VL-7B FT (10K) & 47.9 & 76.5 & 40.6 & 48.5 & 45.0 & 26.8 \\
    Qwen2.5-VL-7B FT (100K) & \underline{\textbf{57.8}} & \underline{\textbf{80.1}} & \underline{\textbf{47.6}} & \underline{\textbf{55.3}} & \underline{\textbf{50.4}} & \underline{\textbf{32.7}} \\
    \textit{(FT $-$ Base)} & \textcolor{green}{$\uparrow$28.7} & \textcolor{green}{$\uparrow$24.8} & \textcolor{green}{$\uparrow$15.3} & \textcolor{green}{$\uparrow$42.2} & \textcolor{green}{$\uparrow$10.6} & \textcolor{green}{$\uparrow$10.7} \\
    \bottomrule
    \end{tabular}%
    \endgroup
    \caption{Category-wise Accuracy (Part II): Land cover and multi-factor categories. Abbreviations: Top uses = top land uses, LC type = land cover type, Spatial open. = spatial openness, Sky vis. = sky visibility, Bldg. dens. = building density, Vis. range = visibility range. \\
    \footnotesize Most multiple-choice questions have four options (25\% baseline), but `regional SVF variability' has three options and `region ranking' varies by question nature.}
    \label{tab:category_acc_b}
\end{table*}

\section{Benchmark Experiments}
\label{sec:benchmark}

\noindent \textbf{Implementation Setting.}
To comprehensively assess 3D spatial reasoning capabilities, we benchmarked ten VLMs. These cover open-source models (LLaVA-OneVision~\cite{li2024llava}, InternVL3-8B~\cite{zhu2025internvl3}, Qwen2.5-VL variants~\cite{bai2025qwen2}), commercial systems (GPT-4o~\cite{hurst2024gpt}, o4-mini, GPT-4.1-mini, Gemini-2.5-Flash), and domain-specific remote sensing VLMs (TEOChat~\cite{irvin2024teochat}, GeoChat~\cite{kuckreja2024geochat}). For fine-tuning, we investigated two configurations: a \emph{10K} and a \emph{100K} set of short-answer Q\&As, both mixed with 1 K free-form Q\&As. 

 \noindent \textbf{Evaluation Metrics.}
For short-answer QA, we report per-category accuracy across the 12 tasks defined in T1+T2~\cite{li2024llava}. Multi-label categories rely on Jaccard similarity with a 0.8 threshold. 
Height estimations require magnitude-aware tolerances to capture relative and absolute errors. Free-form responses were scored on a 1–5 rubric in four application domains, guided by structured ``observation'' and ``conclusion'' outputs. The observation prompt explicitly asked for SVF, elevation, and land cover descriptions, as shown in Table~\ref{tab:freeform_specialized}. 
To test generalization, we evaluated free-form QA using prompts that differed from those used during fine-tuning.

All models were decoded with nearly deterministic settings (temperature of 0.0 or the lowest commercial API setting). Appendix~\ref{sec:appendix_implementation} details the implementation setup, Appendix~\ref{sec:appendix_metrics} explains the metric rationale and sensitivity checks, and Appendix~\ref{sec:appendix_evaluation_protocols} describes the full evaluation protocol.

\noindent \textbf{Detailed Evaluation Protocol}
\label{sec:detailed_eval_protocol}
The pipeline enforces structured output formats while maintaining parsing tolerance to whitespace and reasoning chains. We shuffled multiple-choice options to reduce positional bias and only reported format errors when no valid answer could be recovered from the response.

\subsection{Comparative Results}

\noindent \textbf{Domain gap and scaling effects.}
The aggregated results in \autoref{tab:agg_eval_new_transposed} reveal clear domain gaps in 3D geospatial reasoning for all models. Commercial models generally outperform their open-source counterparts because of their larger pre-training scale and stronger alignment. However, scaling the parameters within the family (Qwen 3B to 7B Base) yielded limited gains without 3D geospatial data (23.6\% to 26.4\%). Notably, o4-mini exhibited 47.2\% format errors (repeated questions without answers), while Gemini-2.5-Flash showed 16.3\% errors, compared to $<$2\% for other models. These format failures are likely due to thinking conventions and conversational defaults, and detailed analysis is provided in Appendix~\ref{sec:appendix_format_errors}.

\noindent \textbf{Fine-tuned models achieve substantial improvements.}
 The fine-tuned Qwen2.5-VL-7B secured the best overall short-answer accuracy (48.7\%) and the highest free-form scores (Total: 2.89, Conclusion: 3.11 in Table~\ref{tab:agg_eval_new_transposed}). Increasing the size of the instruction-tuning corpus from 10 K to 100 K improved short-answer accuracy by 8.3 pp and slightly increased free-form scores (+0.20 in total and +0.10 in conclusion). This result supports the hypothesis that scaling short-answer supervision improves both the short-answer and downstream reasoning tasks.
 Table~\ref{tab:qwen_comprehensive_comparison} details the effects of combining 100 K short-answer training with free-form QA, which constitutes our main `FT' configuration. This combination slightly lifted the overall short-answer accuracy from 48.0\% to 48.7\%, which shows the advantages of supervision learned through open-ended questions.

\noindent \textbf{Category-wise performance variations.}
The SVF-related inference showed substantial gains (e.g., SVF value: +35.9 pp), which indicates that sky view factors are learnable from RGB patterns (Tables ~\ref{tab:category_acc_b} and ~\ref{tab:category_acc_a}). Land use questions also showed sizable gains (e.g., top land uses: +28.7 pp), confirming that domain adaptation enhances 2D visual learning. On the other hand, height-related categories remained challenging despite our coarse decision-oriented design (relative comparisons and bin-level distinctions). This may be due to the ambiguity inherent in inferring 3D spatial properties from monocular 2D RGB inputs, where scale and depth cues are fundamentally underdetermined~\cite{Cai2020Monocular,Lu2021SGTBN,Lourenço2021Intel}. The multi-factor categories improved to 45.5\% (+22.9 pp) after instruction tuning, which reflects better integration of multiple spatial cues (Table~\ref{tab:agg_eval_new_transposed}).
\vspace{-0.1cm}

\begin{table}[tb]
    \centering
    \footnotesize
    \begingroup\setlength{\tabcolsep}{5pt}\renewcommand{\arraystretch}{0.9}
    \begin{tabular}{@{}l|ccc@{}}
    \toprule
    \textbf{Model} & \textbf{SVF} & \textbf{Land Cover} & \textbf{Elevation} \\
    \midrule
    \multicolumn{1}{l|}{\textit{\gb Commercial models}} & & & \\
    o4-mini & 2.03 & 2.12 & 1.73 \\
    Gemini-2.5-Flash & 1.50 & 2.05 & 1.22 \\
    GPT-4o & 2.12 & 2.27 & 1.62 \\
    GPT-4.1-mini & 2.29 & 2.57 & 2.39 \\
    \midrule
    \multicolumn{1}{l|}{\textit{\rsb Remote sensing VLMs}} & & & \\
    TEOChat & 1.25 & 1.50 & 1.38 \\
    GeoChat & 1.20 & 1.23 & 1.19 \\
    \midrule
    \multicolumn{1}{l|}{\textit{\osb Open-source models}} & & & \\
    LLaVA-OneVision & 1.10 & 1.78 & 1.39 \\
    InternVL3-8B & 1.81 & 2.22 & 1.92 \\
    Qwen2.5-VL-3B & 1.29 & 1.36 & 1.27 \\
    Qwen2.5-VL-7B Base & 1.57 & 2.18 & 1.73 \\
    \midrule
    \multicolumn{1}{l|}{\textit{\textbf{Fine-tuned models}}} & & & \\
    Qwen2.5-VL-7B FT (10K) & 3.22 & 2.80 & 2.69 \\
    Qwen2.5-VL-7B FT (100K) & \underline{\textbf{3.40}} & \underline{\textbf{2.91}} & \underline{\textbf{2.85}} \\
    \textit{(FT $-$ Base)} & \textcolor{green}{$\uparrow$1.83} & \textcolor{green}{$\uparrow$0.73} & \textcolor{green}{$\uparrow$1.12} \\
    \bottomrule
    \end{tabular}%
    \endgroup
    \caption{Free-form Analysis (Part I): Specialized feature agreement (1--5 scale).}
    \label{tab:freeform_specialized}
\end{table}

\begin{table}[tb]
    \centering
    \footnotesize
    \begingroup\setlength{\tabcolsep}{4pt}\renewcommand{\arraystretch}{0.9}
    \begin{tabular}{@{}l|cccc@{}}
    \toprule
    \textbf{Model} & \textbf{Urban} & \textbf{Energy} & \textbf{Landscape} & \textbf{Water} \\
    \midrule
    \multicolumn{1}{l|}{\textit{\gb Commercial models}} & & & & \\
    o4-mini & 1.97 & 2.21 & 2.07 & 1.97 \\
    Gemini-2.5-Flash & 2.10 & 2.07 & 2.28 & 1.38 \\
    GPT-4o & 2.14 & 2.52 & 2.38 & 2.03 \\
    GPT-4.1-mini & 2.38 & 2.86 & 2.28 & 2.59 \\
    \midrule
    \multicolumn{1}{l|}{\textit{\rsb Remote sensing VLMs}} & & & & \\
    TEOChat & 1.45 & 1.34 & 1.83 & 1.17 \\
    GeoChat & 1.14 & 1.00 & 1.45 & 1.00 \\
    \midrule
    \multicolumn{1}{l|}{\textit{\osb Open-source models}} & & & & \\
    LLaVA-OneVision & 1.62 & 1.24 & 1.66 & 1.41 \\
    InternVL3-8B & 2.17 & 2.52 & 2.28 & 1.90 \\
    Qwen2.5-VL-3B & 1.31 & 1.21 & 1.86 & 1.14 \\
    Qwen2.5-VL-7B Base & 2.03 & 2.14 & 2.24 & 1.76 \\
    \midrule
    \multicolumn{1}{l|}{\textit{\textbf{Fine-tuned models}}} & & & & \\
    Qwen2.5-VL-7B FT (10K) & 2.62 & 3.00 & 2.38 & \underline{\textbf{2.76}} \\
    Qwen2.5-VL-7B FT (100K) & \underline{\textbf{2.93}} & \underline{\textbf{3.17}} & \underline{\textbf{2.76}} & 2.69 \\
    \textit{(FT $-$ Base)} & \textcolor{green}{$\uparrow$0.90} & \textcolor{green}{$\uparrow$1.03} & \textcolor{green}{$\uparrow$0.52} & \textcolor{green}{$\uparrow$0.93} \\
    \bottomrule
    \end{tabular}%
    \endgroup
    \caption{Free-form Analysis (Part II): Category-specific performance (1--5 scale). \\
    \footnotesize Urban = Urban Development, Energy = Renewable Energy, Water = Water Accumulation.}
    \label{tab:freeform_categories}
\end{table}

\begin{table}[tb]
    \centering
    \begingroup\setlength{\tabcolsep}{3.5pt}\renewcommand{\arraystretch}{0.85}\footnotesize
    \begin{tabular}{@{}l|cccc@{}}
    \toprule
    \textbf{Category} & \textbf{Base} & \textbf{10K+free} & \textbf{100K} & \textbf{100K+free} \\
    \multicolumn{1}{l|}{\textit{Major categories}} & & & & \\
    SVF inference & 23.5 & 34.2 & 43.9 & 44.6 \\
    Height inference & 21.3 & 33.6 & 38.0 & 41.1 \\
    Land cover & 45.3 & 65.1 & 72.0 & 71.3 \\
    Multi-factor & 22.6 & 39.3 & 45.5 & 45.5 \\
    \midrule
    \textbf{Overall} & \textbf{26.4} & \textbf{40.4} & \textbf{48.0} & \textbf{48.7} \\
    \bottomrule
    \end{tabular}
    \endgroup
    \caption{Comprehensive comparison of Qwen2.5-VL-7B variants showing synergistic effects of combining short-answer and free-form training. All values indicate accuracy (\%). `100 K' here means purely 100 K short-answer Q\&As, and 10 K+free and 100 K+free are `10 K' and `100 K' recipes in other tables respectively, mixed with 1 K free-form Q\&As.}
    \label{tab:qwen_comprehensive_comparison}
\end{table}
\noindent \textbf{Upper-bound and Two-stage Baseline Analysis.}
To rigorously assess the value of the end-to-end VLM approach, we conducted additional experiments using inputs other than RGB. 
First, we evaluated a configuration with access to various ground truth modalities (DSM/SVF/segmentation) at inference. Using only the necessary modalities for each question tended to improve category-specific accuracy. The best overall accuracy reached \textbf{57.4\%} (Height Inference: 69.25\%) with agent-style routing, which selected the necessary modalities for each question.
Second, we evaluated a two-stage pipeline that used three pretrained UNet-based monocular predictors on GeoNRW to estimate the DSM/SVF/segmentation from RGB, followed by agent-style inference. 
This achieved only 33.9\% overall accuracy (Height Inference: 14.97\%), which reflects the error propagation in the two-stage pipeline.
Future work could combine improved RGB-to-DSM estimation with VLMs to bridge the gap toward upper-bound performance.
Please refer to Appendix~\ref{sec:appendix_modality_ablations_full} for detailed modality ablation tables.

\subsection{Detailed Analysis}

\noindent \textbf{VLM spatial reasoning challenges and architectural constraints.}
The results point to the fundamental limitations of current vision-language model architectures in spatial tasks requiring coordinate-level understanding. For instance, VLMs achieve only 32.7\% accuracy in the visibility range, which highlights the constraints of patch-based vision encoders. Even with advanced high-resolution techniques such as AnyRes (LLaVA-OneVision) and dynamic resolution processing (Qwen2.5-VL), images are still partitioned into discrete patches rather than interpreted in continuous coordinates~\cite{zhu2025internvl3,Jiao2023DilateFormer:,tang2025data}. This leads to systematic errors, such as misinterpreting patch boundaries as actual coordinate positions~\cite{Zhang2023Vision-Language}, implying that the challenge is rooted in architectural design rather than insufficient training. In contrast, sky visibility achieved 55.3\%, likely because SVF + land cover reasoning aligns more naturally with patch-level recognition than SVF + DSM inference, given that land cover varies less frequently within a patch than elevation.

\noindent \textbf{Evidence for height estimation learning from RGB patterns.}
Domain-specific training still improved the performance on height-aware tasks by +19.8 pp (Table~\ref{tab:agg_eval_new_transposed}). This confirms that despite the challenges~\cite{hou2023enhancing}, VLMs can learn some aspects of 3D reasoning from RGB patterns through supervised learning. Prior work on monocular height estimation~\cite{liasis2016satellite,liu2020im2elevation} has demonstrated that deep learning models can learn height estimation from 2D visual cues. Our substantial performance gains demonstrate the effective conversion of established mappings from visual cues to geometric properties within the VLM architecture when sufficient training data and task-specific supervision are available.

\noindent \textbf{Error patterns and robustness analysis.}
Models without fine-tuning typically exhibit weak spatial grounding, generate vague answers in free-form tasks, and often fail to integrate multiple spatial cues (e.g., SVF and elevation) effectively. In the multi-label land cover category, domain-specific models such as TEOChat and GeoChat often collapse into a single label, indicating limitations in following instructions that may be rooted in VQA training with single answers. 

\noindent \textbf{Performance comparison with specialized methods.}
VLMs achieved 41.1\% and 44.6\% accuracy on height- and SVF-related tasks, respectively (Table~\ref{tab:agg_eval_new_transposed}), which reveals substantial gaps compared with specialized sensor-based approaches. However, VLMs offer unique advantages including natural language interfaces and integrated reasoning across multiple spatial attributes. Our free-form evaluation achieved a 3.11/5 conclusion score and demonstrates compositional reasoning potential that specialized estimators cannot achieve. Although a direct comparison with traditional monocular depth estimation would be valuable, our work focused on evaluating VLMs within a unified framework to provide a consistent assessment of diverse spatial reasoning tasks. We will leave the explicit baseline comparisons with specialized estimators plus rule-based QAs for future work.

\noindent \textbf{Prompt sensitivity analysis.}
To test the model's robustness to prompt phrasing, we reevaluated our best fine-tuned model using a version of the benchmark in which all question texts were paraphrased by GPT-4, and semantic equivalence was checked on 1\% of the samples. The accuracy dropped from 48.7\% to 39.7\%, whereas the format error rate remained very low (0.09\%). This performance still exceeds both the Qwen-base model (26.4\%) and the best inference-only model (33.8\%), indicating that the fine-tuned model retains advantages even under prompt variations. Full results are in Appendix~\ref{sec:appendix_prompt_paraphrasing}. 
\subsection{Limitations and Future Directions}

\noindent \textbf{Current limitations.}
Despite domain-specific improvements, our best models achieved only $<$50\% accuracy on complex spatial reasoning tasks and remain insufficient for reliable deployment. Additionally, the evaluation was confined to North Rhine-Westphalia, which limits generalization claims. A systematic cross-regional evaluation is left for future work. Furthermore, performance varies significantly across task types, and precision for coordinate-level tasks remains problematic, with an accuracy rate of 32.7\%. Although option labels were balanced and choices were randomized during dataset creation to reduce positional bias, a dedicated sensitivity study of multiple-choice option ordering is left for future work. Finally, our geospatial specialization may impact general vision-language performance and requires future assessment of this trade-off.

\noindent \textbf{Future research directions.}
In summary, our error analysis highlights the main challenges of Geo3DVQA and suggests several future research directions.
	1)  \textit{Architectural innovations}: Development of spatial encoders with 3D structure understanding and coordinate-aware attention mechanisms to better integrate 2D patches with continuous spatial positions.
	2)	\textit{Generalization}: Broader geographic and seasonal training data for real-world robustness.
	3)	\textit{Explainability}: Attention visualization and counterfactuals to reveal the actual model reasoning processes for height-aware inference and failure modes.
Pursuing these directions could lead to robust 3D reasoning from RGB without relying on costly specialized sensor data and, in turn, make real-world deployment more scalable.

%% file: sections/conclusion.tex
\vspace{-0.1em}
\section{Conclusion}
\vspace{-0.4em}
\label{sec:conclusion}
We introduced Geo3DVQA, the first VQA benchmark that evaluates height-aware 3D spatial reasoning directly from RGB imagery. Our evaluation across 16 categories and three complexity tiers demonstrates that domain-specific instruction tuning substantially improves reasoning from RGB images alone. Specifically, fine-tuned Qwen2.5-VL-7B attained 48.7\% overall accuracy compared to 26.4\% for the base model, and surpassed the 31.0\% achieved by GPT-4o. Improvements were particularly pronounced in SVF-aware and multi-factor reasoning tasks, as well as in land cover tasks.

Despite this achievement, precise coordinate-level estimation remains a challenge. This necessitates targeted architectural innovations to achieve reliable spatial precision in these models. As a unified 3D inference benchmark based solely on RGB imagery, Geo3DVQA highlights both the current strengths and remaining limitations while identifying concrete opportunities for advancing accessible 3D geospatial understanding without specialized sensors.

%% file: sections/appendix2_reorganized.tex
\begingroup
\sloppy
\setlength{\emergencystretch}{3em}

\section*{Overview of Supplementary Materials}
\addcontentsline{toc}{section}{Overview of Supplementary Materials}

 Appendix~\ref{sec:appendix_implementation} details implementation (models, decoding, infrastructure), Appendix~\ref{sec:appendix_evaluation_protocols} describes evaluation protocols and scoring rules (with sensitivity analyses), Appendix~\ref{sec:appendix_comprehensive_results} presents extended results and analyses, Appendix~\ref{sec:appendix_metrics} formalizes the spatial metrics and methodology, Appendix~\ref{sec:appendix_examples} provides extended qualitative discussions and worked examples, and Appendix~\ref{sec:appendix_data_code} documents data/code availability and release scope.

\section{Technical Implementation Details}
\label{sec:appendix_implementation}

We summarize the technical implementation details for reproducibility, including the experimental setup, dataset construction methodology, evaluation protocols, and supplementary reference materials.

\subsection{Experimental Implementation}
\label{sec:appendix_experimental_setup}

\subsubsection{Model Selection and Training Configurations}
\label{sec:appendix_model_selection}

Our model selection represents the current state-of-the-art capabilities across different paradigms: commercial models provide performance benchmarks, open-source models enable systematic analysis of architectural constraints (high-resolution processing via AnyRes, dynamic resolution handling), and domain-specific models offer comparisons with specialized approaches for temporal and spatial reasoning in remote sensing contexts.


\noindent \textbf{Detailed Instruction-tuning Setups.} Learning rate: 3e-5, batch size: 1 (effective 8 with gradient accumulation), optimizer: AdamW, epochs: 3, schedule: cosine, warmup: 10\% of total steps, weight decay: 0.01, gradient clipping: 1.0, adaptation: LoRA (rank 8, target modules: all transformer attention and feedforward layers).

\subsubsection{Infrastructure and Reproducibility}
\label{sec:appendix_infrastructure}


\noindent \textbf{Software Environment.}
\begin{itemize}
\item Python: 3.9.7
\item PyTorch: 2.2.2+cu121
\item Transformers: 4.21.1
\item CUDA: 12.1
\item CUDNN: 8
\item Additional libraries: NumPy 1.23.5, Pandas 1.3.3, Scikit-learn 1.0.2, aenum, nptyping, rasterio, nvidia-pyindex, openmim
\end{itemize}


\subsubsection{Inference and Decoding Configuration}
\label{sec:appendix_inference_config}

\noindent \textbf{Open-source (HuggingFace) models.}
Unless otherwise noted, inference uses greedy decoding:
\begin{itemize}
  \item \texttt{do\_sample=False}, \texttt{num\_beams=1}
  \item \texttt{temperature=None}, \texttt{top\_p=None}, \texttt{top\_k=None} (ignored when \texttt{do\_sample=False})
  \item \texttt{max\_new\_tokens=256}
  \item Evaluation loop wrapped with \texttt{with torch.no\_grad():} and \texttt{model.generate(**inputs, ...)}
\end{itemize}

\noindent \textbf{Commercial APIs.}
To ensure deterministic conditions across all models, we configured endpoints with temperature set to 0.0 (or the minimum allowable value) and disabled sampling, while maintaining token limits comparable to open-source experiments. However, achieving complete uniformity across API providers presents inherent challenges due to provider-specific constraints. Context window limits, internal token budget allocations, and distributed inference mechanisms vary across different models. Furthermore, commercial models (e.g., GPT-4o, Gemini) may exhibit subtle non-deterministic behavior even at temperature 0.0, attributed to internal optimizations and distributed execution architectures~\cite{ouyang2025empirical}~\cite{renze2024effect}. While we minimized these variations through consistent use of minimum temperature settings and comparable token budgets, readers should note that minor reproducibility differences may emerge from these API-level factors, representing inherent limitations in commercial-to-open-source model comparisons.

\subsubsection{Model Endpoints and Checkpoints}
\label{sec:appendix_models}

We list all models, endpoints, and checkpoints used for evaluation and fine-tuning.
\begin{itemize}
  \item GPT-4o: OpenAI API (\texttt{gpt-4o-2024-08-06})
  \item GPT-4.1-mini: OpenAI API (\texttt{gpt-4.1-mini-2025-04-14})
  \item GPT-5: OpenAI API (\texttt{gpt-5-2025-08-07})
  \item GPT-4: OpenAI API (\texttt{gpt-4-0613})
  \item Qwen2.5-VL-7B (Base): \texttt{Qwen/Qwen2.5-VL-7B-Instruct} (HuggingFace)
  \item Qwen2.5-VL-7B (FT-100K): LoRA-adapted checkpoint from 100K QA fine-tuning (ours)
  \item GeoChat: \texttt{MBZUAI/GeoChat-7B} (HuggingFace)
  \item TeoChat: \texttt{TeoChat/TeoChat-7B} (HuggingFace)
\end{itemize}


\subsubsection{Question Generation Framework}
\label{sec:appendix_question_generation}

Our question-generation framework incorporates several key design principles to ensure systematic coverage of reasoning capabilities while maintaining diversity and practical relevance.

\noindent \textbf{Difficulty Stratification Methodology.} Questions were systematically categorized into difficulty levels through a multi-stage classification process. The stratification framework ensures a broad evaluation of model capabilities across different complexity levels, from basic spatial queries to complex multi-feature reasoning tasks and complete deep reasoning analyses.

\noindent \textbf{Modality-Specific Design Principles.} Each task category is designed to leverage specific data modalities effectively (SVF, DSM, RGB, and semantic segmentation), ensuring that multi-feature integration provides meaningful advantages over single-feature approaches. The design process followed established geospatial analysis methodologies while adapting them for vision-language model evaluation.


\noindent \textbf{Template-Based Generation Strategy.} Following established methodologies from prior VLM benchmarking research, we employed a systematic template-based approach with manually curated question templates for each category. During dataset construction, a GPT-4 family model (default: \texttt{gpt-4o-mini}) optionally paraphrased these base templates to increase linguistic diversity while preserving task semantics and answer formats. Our template system includes: (1) Spatial Comparison Templates for location (region or coordinate-based) comparative analysis; (2) Quantitative Analysis Templates for precise value extraction and calculation; (3) Grid-Based Templates for structured spatial analysis tasks; and (4) Free-form Description Templates for comprehensive landscape analysis.


\subsubsection{Free-form Generation Implementation}
\label{sec:appendix_freeform_generation}

\noindent \textbf{Statistical Feature Extraction.} For Tier 3 free-form questions, we computed comprehensive scene-level statistics from all four modalities (SVF, DSM, RGB, and semantic segmentation). The statistical extraction process includes:

\textbf{Summary Statistics}: Mean, standard deviation, minimum, maximum, and quartiles for continuous data modalities (SVF, DSM).

\textbf{Land Cover Analysis}: Ratio calculations for different semantic classes, spatial distribution patterns, and transition zone analysis.


\textbf{Color Space Analysis}: RGB histogram distributions, dominant color identification, and color diversity metrics.

\textbf{Grid Analysis}: 3x3 grid-based analysis for each metric related to the questions, including statistical computation and reference generation for optimal locations (e.g., SVF, DSM, land cover, suitable places for the question).

\noindent \textbf{Open-ended Q\&As Creation Workflow.} The structured statistics served as input to GPT-4.1-mini to generate contextually relevant free-form questions and answers. The processing workflow follows these steps:
\begin{enumerate}
\item Statistical feature compilation into a structured JSON-style format from SVF, DSM, segmentation, and RGB inputs
\item Template-based question generation using pre-defined categories and base prompts
\item Answer generation based on statistical evidence and domain knowledge using GPT-4.1-mini and human expert prompt refinement
\item Iterative refinement with an LLM-based verifier that checks answers against the SVF/DSM/segmentation/RGB statistics and a 2$\times$2 multimodal panel, combined with rule-based consistency checks
\item Human expert validation on a sampled subset for accuracy and consistency with colorized SVF, DSM, RGB, and segmentation data
\end{enumerate}


\subsubsection{Quality Assurance and Validation}
\label{sec:appendix_quality_assurance}

\noindent \textbf{Bias Mitigation Implementation.} Our generation framework includes systematic bias detection and mitigation strategies to ensure a fair evaluation across different spatial patterns and urban environments. 
The bias mitigation process includes:
\begin{itemize}
\item Systematic sampling across different urban morphologies
\item Geographic diversity within the GeoNRW dataset coverage area
\item Statistical validation of answer distribution patterns
\end{itemize}

\noindent \textbf{Automated Validation Procedures.} The automated validation system implements multiple checking mechanisms.

\textbf{Answer Consistency Verification.} We use automated rule based and LLM based verifiers to check answer consistency. These verifiers compare \texttt{<OBSERVATION>} and \texttt{<CONCLUSION>} against precomputed SVF, DSM, and land cover statistics. They also check against a 2$\times$2 multimodal panel that includes RGB, SVF, DSM, and segmentation data. We supplement these automated checks with human inspection of a sampled subset.







\section{Evaluation Methodology and Protocols}
\label{sec:appendix_evaluation_protocols}


{\noindent \textbf{Deterministic decoding and repeats.}}
Unless otherwise noted, we used deterministic decoding (temperature$=0$; no sampling). 

{\noindent \textbf{Multiple-choice option order.}}
Choices are shuffled at the generation time by a bias-free shuffler that prevents positional skew. 


\subsubsection{Short-answer Scoring Methodology}
\label{sec:appendix_short_answer_scoring}

{\noindent \textbf{Normalization and category-specific criteria.}}
We lowercase the predictions, trim whitespace, and canonicalize comma-separated lists. The category-specific rules are as follows:
\begin{itemize}
  \item \textbf{landcover\_type}: Jaccard similarity $\geq 0.8$ between predicted and ground-truth label sets (with \textit{roads} and \textit{bare\_soil} merged as \textit{ground\_surface}, and cases involving \textit{commercial} excluded for version 2 to mitigate confusion in land-cover terms.)
  \item \textbf{land\_use}: order-independent exact set match
  \item \textbf{height\_average}: mixed tolerance with 10\,m quantization; exact for 0\,m; $\pm 10$\,m if $\leq 30$\,m; $\pm 30\%$ if $> 30$\,m
  \item \textbf{hard\_pixel}: absolute error $\leq 0.05$
\end{itemize}

For multi-label categories (e.g., \textit{landcover\_type}), we used Jaccard similarity (intersection-over-union) between the predicted label set and the ground-truth set, and counted an answer as correct when the similarity was at least 0.8. Jaccard is a standard set-similarity metric in classification and segmentation evaluation and is supported in common ML libraries for multilabel settings (e.g., scikit-learn's \texttt{jaccard\_score})~\cite{scikitlearnJaccard_score}. This threshold strikes a balance between strictness and leniency, being stricter than the IoU@0.5 threshold used in the VOC style and the microsoft coco's mAP@[.50:.95], but not as unforgiving as an exact set match~\cite{everingham2010pascal, lin2014microsoft}.

For height estimation (\textit{height\_average}), numeric answers were compared to the ground truth with 10 m quantization and magnitude aware tolerances. We require exact match for 0 m. For ground truth values up to 30 m, we allow \(\pm 10\) m deviation. For values above 30 m, we use \(\pm 30\%\) relative tolerance.

This mixed absolute and relative rule reflects measurement error characteristics in remote sensing elevation and canopy height products. Satellite DEMs such as SRTM frequently report absolute vertical errors on the order of 5 to 10 m~\cite{rodriguez2006global}. Canopy height estimation methods such as GEDI exhibit substantial relative errors (25 to 45\%)~\cite{lang2022global}. Moreover, relative percentage errors are known to be ill behaved near zero~\cite{hyndman2006another}. Hence, absolute tolerances are preferred for small magnitudes, and relative tolerances are appropriate for larger values.

For SVF value estimation (\textit{hard\_pixel}), we applied an absolute error threshold of \(\leq 0.05\). This threshold was calibrated to the SVF value range [0.0, 1.0] with typical quantization at 0.1 increments. In relative terms, this represents approximately 5--10\% tolerance (0.05/0.5--1.0), which is comparable to the height estimation tolerances when accounting for the different value ranges and measurement characteristics. The different tolerance criteria across tasks (absolute for SVF, mixed absolute/relative for height) reflect the distinct error characteristics of the underlying data sources: SVF values are normalized ratios with bounded ranges, whereas height measurements exhibit scale-dependent uncertainties that require magnitude-aware processing.

\smallskip
\noindent \emph{Revision note (v2).}
We fixed an inconsistency in the GeoNRW land-cover class name mapping used for land-cover-related evaluation (primarily affecting labels 7--9). This issue caused confusion between \textit{road} and \textit{bare\_soil} and, in some cases, unintended inclusion of the \textit{commercial} label in land-cover scoring. 
To ensure a well-defined and reproducible evaluation protocol, we applied deterministic label canonicalization for land-cover evaluation: \textit{road} and \textit{bare\_soil} were merged into \textit{ground\_surface}, and samples involving \textit{commercial} were excluded. We also used a Jaccard similarity threshold of 0.8 for multi-label scoring, instead of 0.7, as it better balances strictness and flexibility.
All experiments were re-evaluated under this corrected setting, and the reported numbers were updated accordingly. While some absolute scores changed slightly, the overall trends and qualitative conclusions remain unchanged. For example, the overall fine-tuned model performance changed from 49.6\% to 48.7\%, the baseline from 24.8\% to 26.4\%, and Gemini-2.5-Flash from 33.0\% to 33.8\%. The result differences are summarized in Table~\ref{tab:shortanswer_v1_v2_summary}.

\smallskip
\noindent \emph{Robustness analysis on refined benchmark (v2.1).}
To further validate the robustness of our approach, we conducted experiments on a refined version of the benchmark, denoted as v2.1.
This version addresses the land cover mapping inconsistencies mentioned above by applying corrected GeoNRW class mapping with detailed label names (without canonicalization).
Furthermore, to evaluate the model's capability in understanding relative geometry, we introduced a \textit{relative\_height} category, an NDSM-style task that estimates the relative elevation of a region compared to its surroundings. To ensure task quality, we enforced stricter margins for the generation of distractors and standardized the floor height assumption to 2.8 m, which better reflected local building standards.
As shown in Table~\ref{tab:rebench_v21_summary}, the fine-tuned model demonstrates significant performance gains even on this rigorous split, achieving over 50\% accuracy and outperforming the base model by a larger margin ($+25.08$ points) compared to the original benchmark.
This confirms that our fine-tuning strategy is effective and robust for label noise correction and task diversification.

\begin{table}[t]
    \centering
    \footnotesize
    \setlength{\tabcolsep}{2.5pt}
    \begin{tabular}{@{}l|ccccc@{}}
    \toprule
    \textbf{Model} & \textbf{Overall} & \textbf{SVF inf.} & \textbf{Height inf.} & \textbf{LU/LC} & \textbf{Multi} \\
    \midrule
    Qwen2.5-VL-7B Base & 25.48 & 25.92 & 23.95 & 32.00 & 21.31 \\
    Qwen2.5-VL-7B FT & 50.56 & 49.03 & 46.03 & 63.30 & 48.77 \\
    \textit{(FT $-$ Base)} & \textcolor{green}{$\uparrow$25.08} & \textcolor{green}{$\uparrow$23.11} & \textcolor{green}{$\uparrow$22.08} & \textcolor{green}{$\uparrow$31.30} & \textcolor{green}{$\uparrow$27.46} \\
    \bottomrule
    \end{tabular}
    \caption{v2.1 evaluation summary (3K test). The values represent accuracy (\%). Major categories: SVF inf. = SVF inference, Height inf. = height inference, LU/LC = land use / land cover type, Multi = multi-factor inference.}
    \label{tab:rebench_v21_summary}
\end{table}

\begin{table}[t]
    \centering
    \scriptsize
    \begingroup\setlength{\tabcolsep}{3.5pt}\renewcommand{\arraystretch}{0.95}
    \resizebox{\columnwidth}{!}{%
    \begin{tabular}{@{}l|ccc@{}}
    \toprule
    \textbf{Category key} & \textbf{Random (\%)} & \textbf{Base (\%)} & \textbf{FT (\%)} \\
    \midrule
    sun\_exposure & 25.00 & 30.32 & 49.46 \\
    SVF\_value & 10.00 & 3.97 & 40.79 \\
    region\_ranking & 16.67 & 21.75 & 44.21 \\
    regional\_svf\_variability & 25.00 & 35.56 & 50.37 \\
    height\_average & -- & 14.74 & 44.91 \\
    highest\_region & 25.00 & 28.93 & 38.57 \\
    relative\_height & -- & 28.18 & 54.30 \\
    land\_use & -- & 34.63 & 53.71 \\
    landcover\_type & -- & 29.45 & 72.60 \\
    sky\_visibility & 25.00 & 14.84 & 57.95 \\
    visibility\_range & 25.00 & 20.63 & 36.71 \\
    openness\_assessment & 25.00 & 38.41 & 60.51 \\  
    urban\_density & 25.00 & 33.74 & 53.99 \\
    \bottomrule
    \end{tabular}%
    }
    \endgroup
    \caption{v2.1 evaluation, detailed results by category (3K test). The values indicate accuracy (\%). The category names follow the evaluation script keys used in v2.1. Random denotes the chance level when applicable, and ``--'' indicates it is not applicable (e.g., non-uniform label spaces).}
    \label{tab:rebench_v21_categories}
\end{table}

\subsubsection{Free-form Evaluation Protocol}
\label{sec:appendix_freeform_protocol}
Open-ended responses were scored by a well-prompted GPT evaluator with temperature$=0.0$ using a rubric that was repeatedly adjusted by human validation. Domain scores (\emph{SVF}, \emph{Land cover}, \emph{Elevation}) are assigned \emph{NaN} only when the ground truth lacks that domain entirely. We saved raw evaluator outputs and per-question summaries for auditability.


\subsubsection{Modality Ablations and Two-Stage Pipeline Analysis}
\label{sec:appendix_modality_ablations_full}

\noindent We conduct modality ablation experiments to quantify how each input type and training recipe contributes to Geo3DVQA performance.
These analyses complement the main benchmark results by contrasting our end-to-end VLM with (i) an upper-bound oracle that has access to ground-truth DSM/SVF at inference and (ii) a conventional two-stage pipeline that first predicts DSM/SVF and segmentation from RGB and then answers questions from these predictions.

\paragraph{Terminology (definitions) for Modality Ablations}
\begin{itemize}
    \item \textbf{Modality}: The type of input image provided to the model. Options include:
    \begin{itemize}
        \item \textbf{RGB}: Optical aerial imagery capturing surface appearance
        \item \textbf{DSM}: Digital Surface Model representing elevation data
        \item \textbf{SVF}: Sky View Factor measuring visible sky proportion (0--1)
        \item \textbf{Seg}: Semantic segmentation map of land cover classes
    \end{itemize}
    \item \textbf{Legend}: A color bar overlay on DSM colormaps showing minimum and maximum elevation values. Because colormaps alone do not convey an absolute scale, adding legends substantially improves the height estimation accuracy.
    \item \textbf{Arrays (numeric image)}: Raw numeric representation of DSM/SVF data (DSM in meters, SVF as 0--1 values) without colormap conversion. This contrasts with the colormap-based visualization.
    \item \textbf{Colormap}: Visualization that converts numeric DSM/SVF values into color gradients for visual interpretation by VLMs.
    \item \textbf{Necessary/needed modalities}: A routing strategy where each question is analyzed (via rules + AI) to determine which modalities are required, providing only the relevant inputs rather than all available data.
\end{itemize}

\noindent \textbf{Summary (inference only).}
\noindent Table~\ref{tab:modality_ablation_infer_summary_appx} reports overall and domain-wise accuracies for non-fine-tuned inference under different input settings, using 3000 QA samples from the test set.
The metrics cover overall accuracy, height inference, land use/land cover (LULC), Sky View Factor (SVF) inference, and multi-factor reasoning, and provide a high-level view of how each modality combination affects performance.
\begin{table*}[tb]
    \centering
    \footnotesize
    \begingroup\setlength{\tabcolsep}{6pt}\renewcommand{\arraystretch}{0.9}
    \resizebox{\textwidth}{!}{%
    \begin{tabular}{@{}l l|ccccc@{}}
        \toprule
        \textbf{Model} & \textbf{Modality at inference} & \textbf{Overall} & \textbf{Height inf.} & \textbf{LULC} & \textbf{SVF inf.} & \textbf{Multi inf.} \\
        \midrule
        \multicolumn{7}{l}{\textit{Qwen2.5-VL-7B Base (no FT)}} \\
        Qwen2.5-VL-7B Base & all & 25.5 & 19.8 & \textbf{44.7} & 22.8 & 21.0 \\
        Qwen2.5-VL-7B Base & dsm (no legend) & 22.5 & 20.9 & 28.2 & 22.4 & 20.1 \\
        Qwen2.5-VL-7B Base & svf & 22.2 & 20.3 & 26.1 & \textbf{23.7} & 18.0 \\
        Qwen2.5-VL-7B Base & rgb & 24.8 & 19.0 & 38.0 & 23.1 & \textbf{22.4} \\
        Qwen2.5-VL-7B Base & rgb+dsm (no legend) & 23.8 & 18.2 & 34.8 & 22.4 & 22.1 \\
        Qwen2.5-VL-7B Base & rgb+dsm (legend) & \textbf{27.8} & \textbf{50.0} & 35.7 & 21.8 & 22.3 \\
        Qwen2.5-VL-7B Base & dsm (legend) & 25.6 & \textbf{50.0} & 21.1 & 23.4 & 19.9 \\
        \midrule
        \multicolumn{7}{l}{\textit{Qwen2.5-VL-7B Base (arrays, no FT)}} \\
        Qwen2.5-VL-7B Base & all (arrays) & \textbf{25.1} & 20.6 & \textbf{44.4} & 21.8 & \textbf{21.2} \\
        Qwen2.5-VL-7B Base & dsm (arrays) & 20.5 & \textbf{21.9} & 15.8 & 22.7 & 18.7 \\
        Qwen2.5-VL-7B Base & svf (arrays) & 19.3 & 19.2 & 11.3 & \textbf{23.2} & 17.5 \\
        \midrule
        \multicolumn{7}{l}{\textit{GPT-4o (no FT)}} \\
        GPT-4o & all & \textbf{28.1} & 25.4 & \textbf{23.5} & 32.1 & 29.6 \\
        GPT-4o & dsm (legend) & 23.7 & \textbf{29.9} & 2.8 & 29.4 & 27.3 \\
        GPT-4o & rgb & 26.8 & 24.1 & 18.6 & 30.0 & \textbf{31.8} \\
        GPT-4o & svf & 24.4 & 25.7 & 3.0 & \textbf{33.8} & 24.6 \\
        \bottomrule
    \end{tabular}%
    }
    \endgroup
    \caption{Modality ablation (inference only) — Summary metrics. Accuracy (Overall), height inference, LULC (land cover \& land use), SVF inference, and multi-factor inference (\%).}
    \label{tab:modality_ablation_infer_summary_appx}
\end{table*}

\noindent \textbf{Detailed categories (inference only).}
\noindent Table~\ref{tab:modality_ablation_infer_appx} expands this analysis to all task categories, grouping SVF-aware, height-aware, land-cover, and multi-factor metrics to show where each modality is most effective.
\begin{table*}[tb]
    \centering

\caption*{(a) SVF-aware and Height-aware categories}
\begingroup\setlength{\tabcolsep}{2.5pt}\renewcommand{\arraystretch}{0.85}
\begin{tabular}{@{}ll|cccccc@{}}
\toprule
\textbf{Model} & \textbf{Modality} & \textbf{SVF val.} & \textbf{Reg. rank} & \textbf{R.SVF.v} & \textbf{Sun.e} & \textbf{Hgt.avg} & \textbf{Highest} \\
\midrule
\multicolumn{8}{l}{\textit{Qwen2.5-VL-7B Base (no FT)}} \\
Qwen2.5-VL-7B Base & all & \textbf{10.6} & 18.9 & 36.0 & 24.9 & 5.9 & 33.7 \\
Qwen2.5-VL-7B Base & dsm w/o leg. & 8.5 & 18.5 & 32.3 & 27.8 & 8.0 & 33.7 \\
Qwen2.5-VL-7B Base & svf & 9.9 & 20.3 & 33.9 & 27.4 & 7.5 & 33.2 \\
Qwen2.5-VL-7B Base & rgb & 5.7 & 17.8 & 35.5 & \textbf{30.2} & 3.7 & 34.2 \\
Qwen2.5-VL-7B Base & rgb+dsm (no leg.) & 9.9 & 18.9 & 32.8 & 24.2 & 1.6 & 34.8 \\
Qwen2.5-VL-7B Base & rgb+dsm (leg.) & 9.5 & 17.8 & 32.3 & 23.8 & 64.2 & \textbf{35.8} \\
Qwen2.5-VL-7B Base & dsm (legend) & \textbf{10.6} & 17.4 & 33.9 & 28.1 & \textbf{64.7} & 35.3 \\
Qwen2.5-VL-7B Base & all (arrays) & 4.6 & 19.9 & 32.8 & 26.0 & 7.5 & 33.7 \\
Qwen2.5-VL-7B Base & dsm (arrays) & 6.4 & 20.3 & \textbf{37.6} & 27.4 & 9.6 & 34.2 \\
Qwen2.5-VL-7B Base & svf (arrays) & 7.8 & \textbf{22.4} & 36.6 & 25.6 & 4.8 & 33.7 \\
\midrule
\multicolumn{8}{l}{\textit{GPT-4o (no FT)}} \\
GPT-4o & all & 15.2 & 28.5 & 45.2 & \textbf{35.2} & 16.6 & 34.2 \\
GPT-4o & dsm (legend) & 15.6 & \textbf{30.2} & 43.5 & 26.7 & \textbf{20.9} & \textbf{39.0} \\
GPT-4o & rgb & 11.7 & 28.1 & 41.9 & \textbf{35.2} & 20.3 & 27.8 \\
GPT-4o & svf & \textbf{27.6} & 29.5 & \textbf{45.7} & 32.4 & 17.1 & 34.2 \\
\bottomrule
\end{tabular}

\endgroup

\vspace{1em}

\caption*{(b) Land cover and Multi-factor categories}
\begingroup\setlength{\tabcolsep}{2.5pt}\renewcommand{\arraystretch}{0.85}
\begin{tabular}{@{}ll|cccccc@{}}
\toprule
\textbf{Model} & \textbf{Modality} & \textbf{Land use} & \textbf{LC type} & \textbf{Open.} & \textbf{Sky vis.} & \textbf{Bldg.d.} & \textbf{Vis.rng} \\
\midrule
\multicolumn{8}{l}{\textit{Qwen2.5-VL-7B Base (no FT)}} \\
Qwen2.5-VL-7B Base & all & 26.6 & \textbf{56.8} & 28.0 & 14.2 & 35.3 & 19.2 \\
Qwen2.5-VL-7B Base & dsm w/o leg. & 8.5 & 41.4 & 28.3 & 12.8 & 31.2 & 20.6 \\
Qwen2.5-VL-7B Base & svf & 3.7 & 41.1 & \textbf{30.8} & 12.1 & 26.5 & 18.9 \\
Qwen2.5-VL-7B Base & rgb & 12.2 & 55.4 & 30.5 & 12.8 & \textbf{42.9} & 19.6 \\
Qwen2.5-VL-7B Base & rgb+dsm (no leg.) & 14.9 & 48.2 & 29.8 & 14.2 & 36.5 & \textbf{21.4} \\
Qwen2.5-VL-7B Base & rgb+dsm (leg.) & 15.4 & 49.3 & 29.4 & 13.5 & 40.0 & 20.3 \\
Qwen2.5-VL-7B Base & dsm (legend) & 6.4 & 31.1 & 30.5 & \textbf{14.9} & 26.5 & 21.0 \\
Qwen2.5-VL-7B Base & all (arrays) & \textbf{30.3} & 53.9 & 29.4 & 12.5 & 38.2 & 19.6 \\
Qwen2.5-VL-7B Base & dsm (arrays) & 1.1 & 25.7 & 26.9 & 14.6 & 27.1 & 17.8 \\
Qwen2.5-VL-7B Base & svf (arrays) & 4.8 & 15.7 & 28.3 & 10.3 & 28.8 & 17.8 \\
\midrule
\multicolumn{8}{l}{\textit{GPT-4o (no FT)}} \\
GPT-4o & all & \textbf{23.9} & \textbf{23.2} & \textbf{40.9} & 29.9 & \textbf{42.4} & 21.7 \\
GPT-4o & dsm (legend) & 5.8 & 0.7 & 35.8 & 24.2 & 35.3 & \textbf{25.6} \\
GPT-4o & rgb & 21.8 & 16.4 & 37.3 & \textbf{32.7} & 41.8 & 24.9 \\
GPT-4o & svf & 6.9 & 0.4 & 38.0 & 23.1 & 37.6 & 18.1 \\
\bottomrule
\end{tabular}

\endgroup
\caption{Modality ablation (inference only) — Detailed categories grouped by SVF-aware, height-aware, land cover, and multi-factor tasks. Abbreviations: SVF val. = SVF value, Reg. rank = region ranking, R.SVF.v = regional SVF variability, Sun.e = sun exposure, Hgt.avg = height average, Highest = highest region, Land use = top land uses, LC type = land cover type, Open. = spatial openness, Sky vis. = sky visibility, Bldg.d. = building density, Vis.rng = visibility range. Qwen2.5-VL-7B Base (no FT) and GPT-4o are evaluated under different input modalities. ``LULC (land cover \& land use)'' denotes combined land use/land cover accuracy.}
\label{tab:modality_ablation_infer_appx}
\end{table*}

\noindent \textbf{Summary (fine-tuned) and Two-Stage Comparison.}
Table~\ref{tab:modality_ablation_ft_summary_appx} provides a comprehensive breakdown of accuracy across fine-tuned configurations. To rigorously evaluate the benefits of an end-to-end approach, we benchmarked a \textbf{Two-Stage Pipeline} in which a U-Net (trained on GeoNRW) first predicts DSM, SVF, and segmentation maps from RGB, followed by a rule-based agent fine-tuned on ground truth data. This two-stage baseline achieved only 33.9\% overall accuracy, lagging significantly behind our end-to-end VLM (50.1\% with RGB FT) and oracle upper bound (57.4\%). This substantial gap is primarily driven by error propagation in height inference (15.0\% vs. 44.1\% for the end-to-end VLM), underscoring the resilience of the unified VLM architecture.

\begin{table*}[tb]
    \centering
    \footnotesize
    \begingroup\setlength{\tabcolsep}{6pt}\renewcommand{\arraystretch}{0.9}
    \resizebox{\textwidth}{!}{%
    \begin{tabular}{@{}l l|ccccc@{}}
        \toprule
        \textbf{Recipe} & \textbf{Modality at inference} & \textbf{Overall} & \textbf{Height inf.} & \textbf{LULC} & \textbf{SVF inf.} & \textbf{Multi inf.} \\
        \midrule

        \multicolumn{7}{l}{\textit{FT with arrays for DSM/SVF (used colormap at inference)}} \\
        FT (arrays) & all & 44.8 & 18.7 & 75.0 & 40.2 & \textbf{47.1} \\
        FT (arrays) & dsm w/o legend & 39.2 & 20.3 & 50.0 & 37.9 & 44.3 \\
        FT (arrays) & rgb & 45.0 & 27.5 & \textbf{77.6} & 38.2 & 45.4 \\
        FT (arrays) & svf & 39.1 & 15.2 & 42.7 & \textbf{41.6} & 44.4 \\
        FT (arrays) & dsm with legend & \textbf{46.3} & \textbf{32.4} & 76.9 & 39.0 & 47.0 \\
        \midrule
        \multicolumn{7}{l}{\textit{FT with RGB}} \\
        FT (rgb) & rgb & 50.1 & 44.1 & \textbf{83.8} & 41.4 & 47.3 \\
        FT (rgb) & dsm+rgb with legend & \textbf{50.5} & \textbf{50.5} & 79.7 & \textbf{41.8} & \textbf{47.4} \\
        \midrule
        \multicolumn{7}{l}{\textit{FT with RGB+DSM (visualized)}} \\
        FT (rgb+dsm) & all & \textbf{48.4} & 48.2 & 76.2 & 39.5 & \textbf{46.6} \\
        FT (rgb+dsm) & dsm & 42.7 & 48.7 & 41.2 & \textbf{40.9} & 44.0 \\
        FT (rgb+dsm) & dsm+rgb & 47.3 & \textbf{50.0} & 75.8 & 37.9 & 44.4 \\
        FT (rgb+dsm) & rgb & 46.8 & 44.9 & \textbf{76.6} & 37.9 & 44.5 \\
        FT (rgb+dsm) & svf & 39.0 & 19.3 & 43.7 & 40.2 & 44.1 \\
        \midrule
        \multicolumn{7}{l}{\textit{FT with necessary modalities (agent-selected)}} \\
        FT (necessary) & all & 53.4 & 58.8 & 88.0 & 41.9 & 49.0 \\
        FT (necessary) & dsm & 45.1 & 64.2 & 42.7 & 39.2 & 47.5 \\
        FT (necessary) & rgb & 41.6 & 27.5 & 55.1 & 37.5 & 47.4 \\
        FT (necessary) & seg & 46.0 & 26.7 & \textbf{90.6} & 35.0 & 47.0 \\
        FT (necessary) & svf & 43.6 & 41.2 & 41.2 & 43.1 & 47.1 \\
        FT (necessary) & dsm+rgb & 48.4 & 63.6 & 63.0 & 39.0 & 48.0 \\
        FT (necessary-only) & dsm+seg+svf & 54.1 & 60.4 & 88.3 & 43.1 & 49.2 \\
        FT (necessary-only) & necessary-only (no rgb) & \textbf{57.4} & \textbf{69.3} & 89.6 & \textbf{46.0} & \textbf{51.6} \\
        \midrule
        \multicolumn{7}{l}{\textit{Two-Stage Baseline (Prediction $\to$ Agent)}} \\
        Two-Stage Pipeline & RGB $\to$ Pred. DSM/SVF & 33.9 & 15.0 & 40.8 & 33.6 & 39.8 \\
        \bottomrule
    \end{tabular}%
    }
    \endgroup
    \caption{Modality ablation (fine-tuned) — Summary metrics. Accuracy (overall), height inference, LULC, SVF inference, and multi-factor inference (\%). The Two-Stage Pipeline serves as a baseline for explicit depth estimation approaches. Results rounded to the first decimal place show minimal or no differences in v2 recalculation.}
    \label{tab:modality_ablation_ft_summary_appx}
\end{table*}

\noindent \textbf{Detailed categories (fine-tuned).}
\noindent Table~\ref{tab:modality_ablation_ft_appx} reports the complete category-level results for the fine-tuned settings with different modality inputs during fine-tuning and inference.
\begin{table*}[tb]
    \centering

 \caption*{(a) SVF-aware and Height-aware categories}
 \begingroup\setlength{\tabcolsep}{2.5pt}\renewcommand{\arraystretch}{0.85}
\begin{tabular}{@{}ll|cccccc@{}}
\toprule
\textbf{Recipe} & \textbf{Modality} & \textbf{SVF val.} & \textbf{Reg. rank} & \textbf{R.SVF.v} & \textbf{Sun.e} & \textbf{Hgt.avg} & \textbf{Highest} \\
\midrule
\multicolumn{8}{l}{\textit{FT (arrays for DSM/SVF at inference)}} \\
FT (arrays) & all & \textbf{39.2} & 33.1 & 44.1 & 43.1 & 0.0 & 37.4 \\
FT (arrays) & dsm & 32.9 & 33.8 & 42.5 & 40.6 & 1.1 & \textbf{39.6} \\
FT (arrays) & rgb & 35.3 & 31.3 & 45.2 & 39.9 & 20.3 & 34.8 \\
FT (arrays) & svf & 31.8 & \textbf{37.7} & \textbf{48.4} & \textbf{45.2} & 0.0 & 30.5 \\
FT (arrays) & dsm (legend) & 35.0 & 33.8 & 45.7 & 42.0 & \textbf{25.7} & 39.0 \\
\midrule
\multicolumn{8}{l}{\textit{FT with RGB}} \\
FT (rgb) & rgb & \textbf{40.3} & 35.6 & 45.7 & 43.1 & 49.7 & 38.5 \\
FT (rgb) & dsm+rgb & 37.1 & \textbf{37.0} & \textbf{47.3} & \textbf{43.8} & \textbf{62.0} & \textbf{39.0} \\
\midrule
\multicolumn{8}{l}{\textit{FT with RGB+DSM (visualized)}} \\
FT (rgb+dsm) & all & 39.7 & 31.9 & 45.1 & \textbf{41.4} & 56.2 & \textbf{40.2} \\
FT (rgb+dsm) & dsm & 38.7 & \textbf{36.6} & 48.2 & \textbf{41.4} & 58.8 & 38.7 \\
FT (rgb+dsm) & dsm+rgb & 39.7 & 29.8 & 42.0 & 39.7 & \textbf{60.3} & 39.7 \\
FT (rgb+dsm) & rgb & \textbf{40.8} & 29.5 & 44.0 & 38.7 & 54.6 & 35.1 \\
FT (rgb+dsm) & svf & 35.3 & 34.6 & \textbf{50.8} & \textbf{41.4} & 10.3 & 28.4 \\
\midrule
\multicolumn{8}{l}{\textit{FT with necessary modalities (agent-selected)}} \\
FT (necessary) & all & \textbf{42.8} & 38.4 & 42.5 & 39.9 & 74.3 & 43.3 \\
FT (necessary) & dsm & 35.7 & 33.8 & 45.2 & 36.3 & 83.4 & 44.9 \\
FT (necessary) & rgb & 33.2 & 32.0 & 41.4 & 41.6 & 24.1 & 31.0 \\
FT (necessary) & seg & 30.0 & 23.8 & 41.4 & 38.8 & 23.0 & 30.5 \\
FT (necessary) & svf & 38.9 & 39.9 & 48.9 & 44.1 & 49.2 & 33.2 \\
FT (necessary) & dsm+rgb & 35.3 & 33.8 & 44.1 & 36.7 & 85.0 & 42.3 \\
FT (necessary-only) & dsm+seg+svf & 39.6 & 41.6 & 38.4 & 51.1 & 77.0 & 43.9 \\
FT (necessary-only) & necessary-only (no rgb) & 37.8 & \textbf{44.1} & \textbf{50.5} & \textbf{45.9} & \textbf{93.1} & \textbf{45.5} \\
 \bottomrule
 \end{tabular}
 \endgroup

 \vspace{1em}

 \caption*{(b) Land cover and Multi-factor categories}
 \begingroup\setlength{\tabcolsep}{2.5pt}\renewcommand{\arraystretch}{0.85}
 \begin{tabular}{@{}ll|cccccc@{}}
 \toprule
 \textbf{Recipe} & \textbf{Modality} & \textbf{Land use} & \textbf{LC type} & \textbf{Open.} & \textbf{Sky vis.} & \textbf{Bldg.d.} & \textbf{Vis.rng} \\
 \midrule
 \multicolumn{8}{l}{\textit{FT (arrays for DSM/SVF at inference)}} \\
 FT (arrays) & all & 54.3 & 88.9 & 43.0 & \textbf{52.7} & \textbf{51.8} & \textbf{38.8} \\
 FT (arrays) & dsm & 22.9 & 68.2 & 41.2 & 51.3 & 45.9 & 36.3 \\
 FT (arrays) & rgb & \textbf{55.3} & \textbf{92.5} & 41.6 & 51.6 & 49.4 & 36.7 \\
 FT (arrays) & svf & 22.3 & 56.4 & \textbf{47.3} & 52.0 & 43.5 & 37.4 \\
 FT (arrays) & dsm (legend) & 53.7 & \textbf{92.5} & 40.9 & \textbf{52.7} & 51.8 & 38.4 \\
 \midrule
 \multicolumn{8}{l}{\textit{FT with RGB}} \\
 FT (rgb) & rgb & \textbf{65.4} & \textbf{96.1} & 43.7 & \textbf{55.2} & \textbf{54.1} & 35.2 \\
 FT (rgb) & dsm+rgb & 60.1 & 92.9 & \textbf{45.9} & \textbf{55.2} & \textbf{54.1} & \textbf{35.6} \\
 \midrule
 \multicolumn{8}{l}{\textit{FT with RGB+DSM (visualized)}} \\
 FT (rgb+dsm) & all & 57.1 & 89.0 & 41.2 & \textbf{54.3} & \textbf{53.1} & 34.9 \\
 FT (rgb+dsm) & dsm & 19.4 & 55.8 & 41.9 & 50.2 & 38.6 & \textbf{41.1} \\
 FT (rgb+dsm) & dsm+rgb & 59.7 & 86.6 & 39.5 & 53.6 & 39.1 & 38.4 \\
 FT (rgb+dsm) & rgb & \textbf{56.6} & \textbf{90.1} & 38.5 & 51.9 & 50.3 & 33.6 \\
 FT (rgb+dsm) & svf & 19.4 & 59.9 & \textbf{42.3} & 51.9 & 42.5 & 37.3 \\
 \midrule
 \multicolumn{8}{l}{\textit{FT with necessary modalities (agent-selected)}} \\
 FT (necessary) & all & 71.8 & 98.9 & 46.2 & 54.1 & \textbf{55.9} & 39.9 \\
 FT (necessary) & dsm & 18.1 & 59.3 & 47.0 & 55.9 & 44.1 & 41.3 \\
 FT (necessary) & rgb & 20.2 & 78.6 & 40.5 & 53.4 & 54.1 & 37.4 \\
 FT (necessary) & seg & \textbf{76.6} & \textbf{100.0} & 43.0 & 51.3 & 54.1 & 38.4 \\
 FT (necessary) & svf & 19.7 & 55.7 & 45.9 & 52.7 & 52.9 & 38.1 \\
 FT (necessary) & dsm+rgb & 29.8 & 85.4 & 47.0 & 53.4 & 51.8 & 40.2 \\
 FT (necessary-only) & dsm+seg+svf & 70.0 & 100.0 & 47.3 & 55.2 & 47.7 & 44.1 \\
 FT (necessary-only) & necessary-only (no rgb) & 75.0 & 98.9 & \textbf{53.4} & \textbf{56.6} & \textbf{55.9} & \textbf{44.1} \\
 \bottomrule
 \end{tabular}
 \endgroup
 \caption{Modality ablation (fine-tuned Qwen2.5-VL-7B) — Detailed categories grouped by SVF-aware, height-aware, land cover, and multi-factor tasks.}
\label{tab:modality_ablation_ft_appx}
\end{table*}

\noindent \textbf{Key observations on modality alignment.}
Across both non-fine-tuned and fine-tuned settings, the highest accuracy is achieved when the input modality matches the queried attribute. For example, calibrated DSMs improved accuracy for height-aware questions, SVF images improved accuracy for SVF tasks, and segmentation, RGB, or all modality combinations improved accuracy for land-cover and appearance-based reasoning.
In non-fine-tuned inference, adding a legend to DSM inputs roughly doubles the accuracy of height inference for the Qwen2.5-VL-7B base model (from about 20\% to 50.0\% in Table~\ref{tab:modality_ablation_infer_summary_appx}), and it would be because the legend resolves absolute scale ambiguities that cannot be recovered from colormaps alone.
Fine-tuning further amplifies this benefit. The agent-style recipe (`FT (necessary)'), which routes each question to its required modalities, achieves strong overall accuracy and dominates the performance for the height-aware categories (Table~\ref{tab:modality_ablation_ft_appx}).

\noindent \textbf{Oracle performance with necessary-only routing.}
The upper-bound capability of the system appears when RGB is excluded and only the ground-truth modalities required for each question are provided (`necessary-only (no rgb)' in Table~\ref{tab:modality_ablation_ft_summary_appx}).
This configuration achieves the best overall accuracy of \textbf{57.4\%}, with particularly strong height-aware performance (height inference 69.3\%,  height-average 93.1\%) and near-perfect land-cover type accuracy (98.9\%) in Table~\ref{tab:modality_ablation_ft_summary_appx},\ref{tab:modality_ablation_ft_appx}.
These results confirm the importance of strict alignment between modality and question type. DSMs should be used for height-aware reasoning, segmentation maps for semantic land cover-related tasks, and SVFs for sky view factor questions.

\paragraph{Visualizations versus raw arrays}
We observed that using DSM or SVF as direct numeric arrays (without colormap rendering) generally degraded performance compared with calibrated colormap visualizations, and overall accuracy for DSM-only and SVF-only settings dropped from roughly 22\% to 19--20\% in Table~\ref{tab:modality_ablation_infer_summary_appx}.
This pattern supports the hypothesis that current multimodal transformers are optimized for extracting features from visual patterns rather than raw numerical grids, validating our choice of calibrated colormaps.
A small exception occurs for height inference with `dsm' in the legend-free setting where arrays slightly outperform colormaps, but DSM with legends still yields much higher height-aware accuracy overall.

\paragraph{Role of VLM-based reasoning}
Although raw height values can be obtained from sensors, many benchmark questions require interpreting spatial relationships and composing multiple cues.
Our fine-tuned VLM attains a high free-form conclusion score (3.29/5 or 3.11/5 in Table~\ref{tab:freeform2_general_appendix}), which demonstrates that the VLM can synthesize spatial relationships and explain height-dependent phenomena in context.
Such compositional reasoning is difficult to replicate with purely rule-based or retrieval-oriented systems, highlighting the value of VLMs for 3D geospatial question answering.

\paragraph{Sensitivity to RGB-to-DSM estimation errors}
A critical insight emerges when comparing the oracle setup with the two-stage pipeline: using explicit RGB-to-DSM predictors results in significantly lower accuracy (33.9\%) than using ground-truth modalities (57.4\%). This suggests that current VLMs are sensitive to estimation artifacts in upstream predictions, which further supports the end-to-end learning approach. In this approach, the model can potentially learn robust features directly from RGB or fuse noisy inputs more effectively.

\paragraph{Generalization risks of colormap dependence}
The strong gains from legends and specific colormap choices also expose potential generalization risks, as models become accustomed to particular visualization schemes.
Our plasma (SVF) and terrain (DSM) colormaps follow domain conventions, but substantial changes in color mapping or legend design could degrade performance.
Future work should investigate colormap-invariant encoders and hybrid representations that jointly embed raw numeric arrays and their visualizations, thereby reducing reliance on any single color palette while preserving the benefits of calibrated legends.

\section{Comprehensive Results Analysis}
\label{sec:appendix_comprehensive_results}

\subsubsection{Free-form Evaluation Results}
\label{sec:appendix_freeform_eval}

We evaluated \emph{free-form} categories under a controlled protocol: training used 1000 free-form QA pairs, and evaluation used a disjoint set of 100 free-form QA pairs covering four categories (Urban Development, Renewable Energy, Landscape Analysis, Water Accumulation). An automated rubric-based assessor (GPT-4) scored the responses on a 1--5 scale along the general and domain axes. Subsequently, a sampled subset was evaluated by human annotators to validate the automated scoring. 

\noindent \textbf{Note on evaluator/template affinity.} Because our prompt templates were developed using GPT-4.1-mini and the rubric judge is a GPT-4–family model, the evaluation may exhibit shared-family bias: the judge could assign slightly higher scores to outputs that match stylistic patterns common in the same model family. Accordingly, we treat rubric-based metrics as relative rather than absolute, and iteratively refined the rubric until its judgments aligned with human annotators on a validation subset. The full rubric and evaluation prompts will be released in our project repository to ensure transparency. Nevertheless, the Geo3DVQA fine-tuned Qwen2.5-VL-7B surpasses GPT-4.1-mini and other baselines across most categories, indicating that our main conclusions are robust to this potential bias.

\noindent \textbf{Land-cover label leniency in free-form scoring (v2).} To mitigate confusion in land-cover phrasing, we added an evaluator instruction that treats \textit{road} and \textit{bare\_soil} as equivalent to \textit{ground\_surface} and ignores \textit{commercial} mentions in land-cover scoring. As expected, the land-cover-related scoring changes only slightly, and most models show a small increase in all categories, which may partly reflect a generic ``leniency'' effect of the judge prompt. Importantly, the relative comparisons remain nearly unchanged. In Table~\ref{tab:freeform2_general_appendix}, scores outside parentheses use this lenient instruction and values in parentheses denote the original scores.

\begin{table}[h]
    \centering
    \footnotesize
    \resizebox{\columnwidth}{!}{%
    \begin{tabular}{@{}lcccc@{}}
    \toprule
    \textbf{Model} & \textbf{Total} & \textbf{Observation} & \textbf{Logic} & \textbf{Conclusion} \\
    \midrule
    Gemini-2.5-Flash & 2.07 (1.96) & 2.00 (1.92) & 2.68 (2.66) & 2.47 (2.34) \\
    GPT-4o & 2.29 (2.27) & 2.28 (2.27) & 3.17 (3.23) & 2.64 (2.57) \\
    GPT-4.1-mini & 2.62 (2.53) & 2.62 (2.52) & 3.44 (3.37) & 3.09 (3.02) \\
    o4-mini & 2.09 (2.05) & 2.03 (1.98) & 2.72 (2.69) & 2.51 (2.41) \\
    LLaVA-one-vision & 1.68 (1.48) & 1.00 & 1.91 (1.89) & 2.09 (2.01) \\
    InternVL3-8B & 2.34 (2.22) & 2.16 (2.08) & 2.99 (2.94) & 2.67 (2.62) \\
    Qwen2.5-VL-3B & 1.51 (1.38) & 1.00 & 1.92 (1.90) & 2.16 (2.11) \\
    Qwen2.5-VL-7B Base & 2.09 (2.04) & 2.07 (2.06) & 2.80 (2.72) & 2.39 (2.23) \\
    TeoChat & 1.45 & 1.39 & 1.91 & 1.66 \\
    GeoChat & 1.15 & 1.00 & 1.68 & 1.67 \\
    Qwen2.5-VL-7B FT & \textbf{3.05 (2.89)} & \textbf{3.06 (2.91)} & \textbf{3.55 (3.41)} & \textbf{3.29 (3.11)} \\
    \textit{\small (FT $-$ Base)} & \small$\uparrow$0.96 (0.85) & \small$\uparrow$0.99 (0.85) & \small$\uparrow$0.75 (0.69) & \small$\uparrow$0.90 (0.88) \\
    \bottomrule
    \end{tabular}
    }
    \caption{General criteria (free-form questions). Higher is better. Values outside parentheses use the land-cover label leniency instruction, and the values in parentheses denote the original scores.}
    \label{tab:freeform2_general_appendix}
\end{table}

\begin{table}[h]
    \centering
    \footnotesize
    \begingroup\setlength{\tabcolsep}{5pt}\renewcommand{\arraystretch}{0.9}
    \resizebox{\columnwidth}{!}{%
    \begin{tabular}{@{}l|ccc@{}}
    \toprule
    \textbf{Model} & \textbf{SVF} & \textbf{Land Cover} & \textbf{Elevation} \\
    \midrule
    \multicolumn{1}{l|}{\textit{\gb Commercial models}} & & & \\
    o4-mini & 2.06 (2.03) & 2.19 (2.12) & 1.77 (1.73) \\
    Gemini-2.5-Flash & 1.63 (1.50) & 2.21 (2.05) & 1.26 (1.22) \\
    GPT-4o & 2.19 (2.12) & 2.34 (2.27) & 1.66 (1.62) \\
    GPT-4.1-mini & 2.37 (2.29) & 2.60 (2.57) & 2.46 (2.39) \\
    \midrule
    \multicolumn{1}{l|}{\textit{\rsb Remote sensing VLMs}} & & & \\
    TEOChat & 1.25 & 1.50 & 1.38 \\
    GeoChat & 1.20 & 1.23 & 1.19 \\
    \midrule
    \multicolumn{1}{l|}{\textit{\osb Open-source models}} & & & \\
    LLaVA-OneVision & 1.17 (1.10) & 1.82 (1.78) & 1.44 (1.39) \\
    InternVL3-8B & 1.90 (1.81) & 2.28 (2.22) & 1.99 (1.92) \\
    Qwen2.5-VL-3B & 1.39 (1.29) & 1.45 (1.36) & 1.35 (1.27) \\
    Qwen2.5-VL-7B Base & 1.63 (1.57) & 2.20 (2.18) & 1.84 (1.73) \\
    \midrule
    \multicolumn{1}{l|}{\textit{\textbf{Fine-tuned models}}} & & & \\
    Qwen2.5-VL-7B FT (10K) & 3.22 & 2.80 & 2.69 \\
    Qwen2.5-VL-7B FT (100K) & \textbf{3.59 (3.40)} & \textbf{3.09 (2.91)} & \textbf{3.02 (2.85)} \\
    \textit{(FT $-$ Base)} & \textcolor{green}{$\uparrow$1.96 ($\uparrow$1.83)} & \textcolor{green}{$\uparrow$0.89 ($\uparrow$0.73)} & \textcolor{green}{$\uparrow$1.18 ($\uparrow$1.12)} \\
    \bottomrule
    \end{tabular}%
    }
    \endgroup
    \caption{Specialized feature agreement (free-form questions, 1--5 scale). Values outside parentheses use the land-cover label leniency instruction (roads/bare\_soil $\rightarrow$ ground\_surface, commercial mentions ignored), and values in parentheses denote the original scores. 
    }
\end{table}

\begin{figure}[t]
    \centering
    \includegraphics[width=0.96\linewidth]{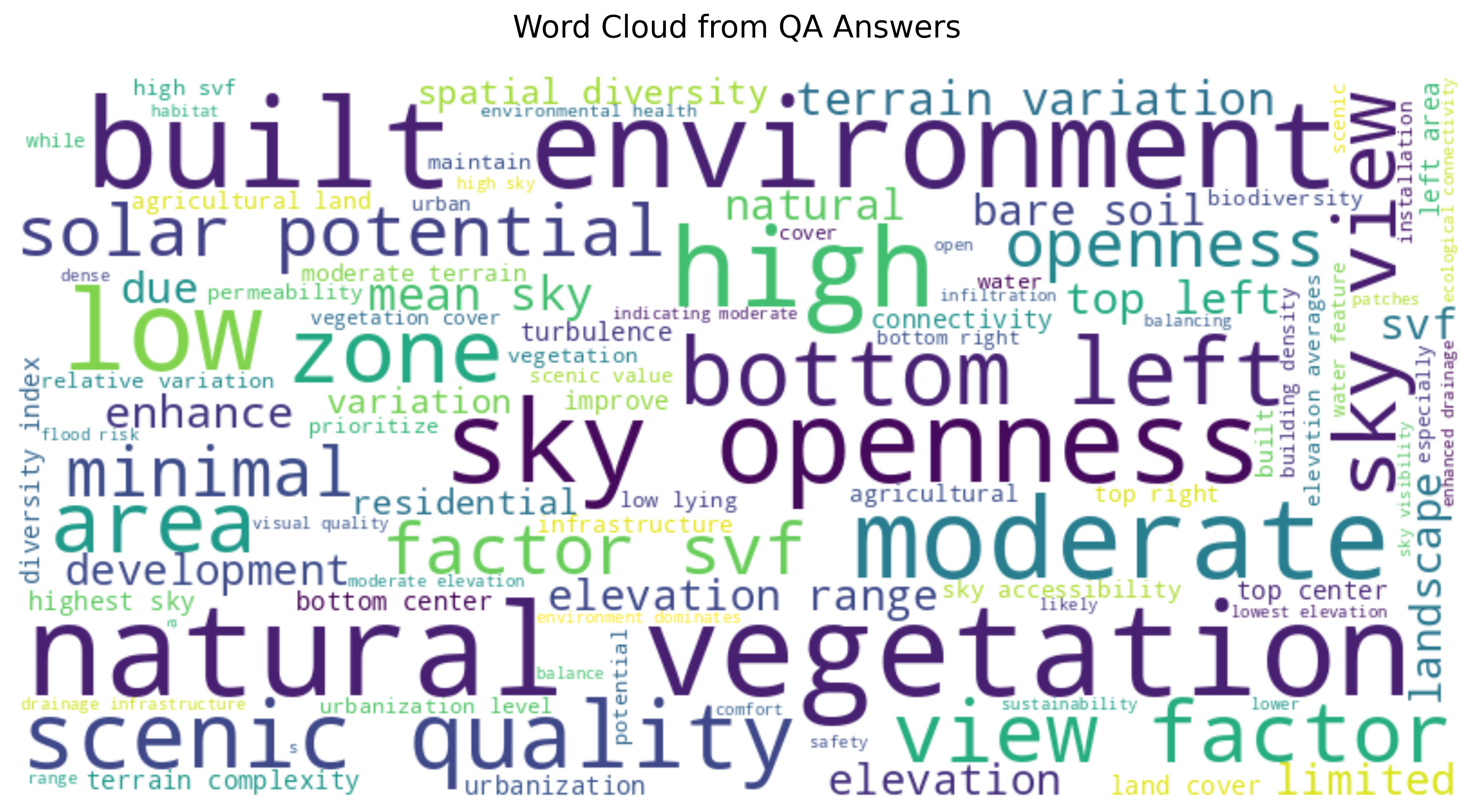}
    \caption{Word cloud of ground-truth answers for Tier-3 free-form Q\&As.}
    \label{fig:wordcloud_answers}
\end{figure}
\subsubsection{Synergistic Training Effects Analysis}
\label{sec:appendix_synergistic_training}

Table~\ref{tab:qwen_comprehensive_comparison_app} demonstrates the synergistic effects of combining short-answer and free-form QA training, showing systematic improvements with notable gains in the overall accuracy. 

\subsubsection{Short-answer results: v1 vs v2}
\label{sec:appendix_shortanswer_v1_v2}
To ensure transparency, we report a short-answer summary table with both v2 (recalculated) and v1 (original) results in Table~\ref{tab:shortanswer_v1_v2_summary}. Values outside parentheses are v2 and values in parentheses are v1. Only the land-cover-related evaluation and its aggregation were affected.

\begin{table}[t]
    \centering
    \scriptsize
    \begingroup\setlength{\tabcolsep}{3.2pt}\renewcommand{\arraystretch}{0.9}
    \resizebox{\columnwidth}{!}{%
    \begin{tabular}{@{}l|cccc|c@{}}
    \toprule
    \textbf{Model} & \textbf{SVF} & \textbf{Height} & \textbf{Land Use/LC} & \textbf{Multi} & \textbf{Overall} \\
    \midrule
    *o4-mini & 20.2 & 11.7 & 23.4 (15.3) & 23.7 & 20.5 (19.2) \\
    GPT-4o & 30.7 & 27.8 & 32.3 (17.5) & 32.4 & 31.0 (28.6) \\
    GPT-4.1-mini & 28.5 & 23.7 & 43.3 (25.1) & 31.5 & 31.0 (28.1) \\
    Gemini-2.5-Flash & 32.9 & 19.4 & 45.9 (40.0) & 35.6 & 33.8 (33.0) \\
    TEOChat & 17.5 & 21.4 & 10.1 (4.0) & 31.5 & 20.3 (19.3) \\
    GeoChat & 20.2 & 21.0 & 26.5 (22.1) & 25.4 & 22.6 (21.9) \\
    LLaVA-OneVision & 22.4 & 24.5 & 26.1 (21.6) & 19.3 & 22.4 (21.8) \\
    InternVL3-8B & 21.8 & 24.9 & 32.9 (20.7) & 22.0 & 24.0 (22.1) \\
    Qwen2.5-VL-3B & 21.3 & 22.5 & 30.2 (33.0) & 24.4 & 23.6 (24.2) \\
    Qwen2.5-VL-7B Base & 23.5 & 21.3 & 45.3 (34.5) & 22.6 & 26.4 (24.8) \\
    Qwen2.5-VL-7B FT (10K) & 34.24 & 33.56 & 65.1 (71.25) & 39.30 & 40.4 (41.43) \\
    Qwen2.5-VL-7B FT (100K) & 44.61 & 41.07 & 71.3 (77.09) & 45.45 & 48.7 (49.63) \\
    \bottomrule
    \end{tabular}%
    }
    \endgroup
    \caption{Short-answer summary results for v2 (outside parentheses) and v1 (in parentheses).}
    \label{tab:shortanswer_v1_v2_summary}
\end{table}

\begin{table}[h]
    \centering
    \footnotesize
    \begin{tabular}{@{}l|cccc@{}}
    \toprule
    \textbf{Category} & \textbf{Base} & \textbf{10K+free} & \textbf{100K} & \textbf{100K+Free} \\
    \midrule
    \multicolumn{1}{l|}{\textit{SVF-aware inference}} & & & & \\
    SVF value & 6.8 & 27.9 & 42.5 & 42.7 \\
    region ranking & 19.5 & 29.6 & 37.8 & 39.7 \\
    regional SVF variability & 34.0 & 39.4 & 53.3 & 51.4 \\
    sun exposure & 28.6 & 35.3 & 42.3 & 43.9 \\
    \midrule
    \multicolumn{1}{l|}{\textit{Height-aware inference}} & & & & \\
    height average & 9.9 & 34.8 & 35.7 & 42.2 \\
    highest region & 32.7 & 32.3 & 40.2 & 39.9 \\
    \midrule
    \multicolumn{1}{l|}{\textit{Multi-factor inference}} & & & & \\
    spatial openness & 32.2 & 40.6 & 46.5 & 47.6 \\
    sky visibility & 13.1 & 48.5 & 56.0 & 55.3 \\
    building density & 39.8 & 45.0 & 49.7 & 50.4 \\
    visibility range & 22.0 & 26.8 & 32.5 & 32.7 \\
    \midrule
    \multicolumn{1}{l|}{\textit{Major categories}} & & & & \\
    SVF inference & 23.5 & 34.2 & 43.9 & 44.6 \\
    Height inference & 21.3 & 33.6 & 38.0 & 41.1 \\
    Land cover & 45.3 & 65.1 & 72.0 & 71.3 \\
    Multi-factor & 22.6 & 39.3 & 45.5 & 45.5 \\
    \midrule
    \textbf{Overall} & \textbf{26.4} & \textbf{40.4} & \textbf{48.0} & \textbf{48.7} \\
    \bottomrule
    \end{tabular}
    \vspace{0.2em}
    \parbox{\columnwidth}{\footnotesize This table demonstrates the synergistic effects of combining short-answer and free-form QA training. Key observations: (1) The 100 K+Free configuration achieves the best overall accuracy; (2) Most categories improve with larger short-answer supervision; (3) Free-form augmentation yields a modest overall gain from 100 K to 100 K+Free.}
    \caption{Comprehensive comparison of Qwen2.5-VL-7B variants showing synergistic effects of combining short-answer and free-form training. All values show accuracy (\%). `100 K' here means purely 100 K short-answer Q\&As, and 10 K+free and 100 K+Free are `10 K' and `100 K' recipes respectively, mixed with 1 K free-form Q\&As.}
    \label{tab:qwen_comprehensive_comparison_app}
\end{table}

\subsubsection{Format Error Analysis}
\label{sec:appendix_format_errors}

\noindent \textbf{Output formatting and decoding behavior.} 
The short-answer evaluation pipeline first analyzes raw model outputs to design task-specific parsing rules that follow each model's typical output tendencies (e.g., stripping chain-of-thought prefixes, extracting the last line as the final answer, or canonicalizing comma-separated lists). We counted an output as a formatting error only when no valid answer could be recovered under these tolerant rules. Quantitatively, o4-mini exhibited a high overall rate of invalid outputs, whereas Gemini-2.5-Flash maintained a lower overall error rate but showed pronounced spikes in specific categories (e.g., \textit{SVF value}). The o4-mini often repeats questions without answers for difficult questions, whereas Gemini-2.5-Flash sometimes exceeds the total token limits (including thinking tokens), and increasing the limits leads to redundant answers without following answer instructions. For such ``thinking-only'' responses that never commit to a final answer string, we classify the outputs as formatting errors in our statistics. Both models had large error rates in the SVF value, which estimates the absolute average SVF value in the region, likely because absolute inference is difficult. The combined summary is presented in Table~\ref{tab:format_errors_combined}. The per-category error counts in this table sum to the overall error counts (e.g., 4833/10232 for o4-mini). 

\begin{table}[h]
    \centering
    \footnotesize
    \begin{tabular}{@{}p{0.35\columnwidth}|p{0.12\columnwidth}p{0.12\columnwidth}|p{0.12\columnwidth}p{0.12\columnwidth}@{}}
    \toprule
    \multirow{2}{*}{\textbf{Category}} & \multicolumn{2}{c|}{\textbf{o4-mini}} & \multicolumn{2}{c}{\textbf{Gemini-2.5-Flash}} \\
     & \textbf{Errors} & \textbf{Rate (\%)} & \textbf{Errors} & \textbf{Rate (\%)} \\
    \midrule
    Overall & 4833/10232 & 47.23 & 1666/10232 & 16.28 \\
    SVF\_value & 692 & 69.27 & 811 & 81.18 \\
    region\_ranking & 640 & 64.06 & 36 & 3.60 \\
    height\_average & 299 & 44.89 & 305 & 45.80 \\
    highest\_region & 369 & 55.41 & 128 & 19.22 \\
    top\_land\_uses & 151 & 22.67 & 4 & 0.60 \\
    landcover\_type & 575 & 57.91 & 85 & 8.56 \\
    spatial\_openness & 443 & 44.34 & 50 & 5.01 \\
    regional\_svf\_variability & 110 & 16.62 & 14 & 2.11 \\
    sky\_visibility & 436 & 43.64 & 100 & 10.01 \\
    sun\_exposure & 356 & 35.64 & 66 & 6.61 \\
    building\_density & 225 & 38.46 & 20 & 3.42 \\
    visibility\_range & 537 & 53.75 & 47 & 4.70 \\
    \bottomrule
    \end{tabular}
    \caption{Format error counts and rates across categories for o4-mini and Gemini-2.5-Flash.}
    \label{tab:format_errors_combined}
\end{table}

\noindent \textbf{Performance analysis insights.} Among the non-fine-tuned models, Gemini-2.5-Flash attained the highest overall accuracy (33.8\%), likely because of its hybrid reasoning architecture with thinking capabilities. Without fine-tuning, Gemini-2.5-Flash performs strongly across short-answer categories, except for height inference. We assume that pre-existing 2D geospatial datasets and deliberate option-by-option comparisons achieved by reasoning conventions improved selection accuracy among candidates.

\section{Spatial Metrics and Scientific Methodology}
\label{sec:appendix_metrics}

\subsubsection{Scientific Rationale for Spatial Metrics}
\label{sec:appendix_scientific_rationale}

\noindent \textbf{Rationale and weight selection.} We derived the weighting schemes for our composite indices from established geospatial and environmental design literature. Primary physical determinants dominate each score, whereas proxy and auxiliary indicators contribute only marginally.

\noindent \textbf{Urban density} assessment relies on coverage and volume based metrics. Building coverage ratio (BCR) and floor area ratio (FAR) serve as fundamental descriptors of built form~\cite{pont2023spacematrix}. We assign BCR the highest weight (0.5) because surface occupancy directly shapes street level compactness. FAR receives a weight of 0.25 because it exhibits less perceptual salience despite its importance in planning contexts.

Secondary indicators complement these primary metrics. The SVF complement captures geometric enclosure~\cite{grimmond2001rapid,chen2012sky} with a moderate weight (0.15). Edge density reflects morphological fragmentation, while local daytime RGB brightness and shadowing provide only a weak proxy for built intensity. Dense built-up areas tend to appear darker during daytime imagery due to cast shadows, higher impervious surface ratios, and reduced vegetation cover~\cite{morabito2017urban,bonafoni2018land,krayenhoff2016daytime}. Both indicators therefore receive minimal weights (0.05), given their strong dependence on illumination conditions, sensor viewing geometry, and surface properties rather than urban form alone.

\noindent \textbf{Spatial openness} assessment emphasizes geometric enclosure over surface coverage. The topographic Openness Index~\cite{yokoyama2002visualizing} dominates the score (0.5) because it directly captures elevation based enclosure. SVF complements this geometric descriptor (0.25) by quantifying sky access from surrounding structures and vegetation~\cite{oke1981canyon}.

Because building density correlates with SVF, we assign it a lower weight (0.15) to avoid double counting occlusion effects. Terrain flatness (0.05) accounts for slope's influence on viewsheds~\cite{fisher1993algorithm}, while visual simplicity (0.05) reflects evidence that spatial complexity reduces perceived openness~\cite{ewing2009measuring}.

\noindent \textbf{Sky visibility} assessment relies on SVF to determine radiative exchange and vertical openness~\cite{oke1981canyon}. However, it can overestimate perceived openness when horizontal obstructions and near field barriers are significant~\cite{miao2020review,middel2018sky,dirksen2019sky}. We therefore assign SVF the highest coefficient (0.7) while explicitly combining it with building penalties (0.3). These penalties encode canyon geometry effects and local building occupancy that constrain sky access and urban climate processes~\cite{grimmond2001rapid,Daramola2019Analysis,Xia2021Sky}.

Edge penalties remain small (0.05 to 0.025) because skyline roughness affects the subjective continuity of sky views but contributes less to the physical sky fraction~\cite{stamps2005enclosure}.

These hierarchical weighting reflects both geometric theory and empirical evidence. Primary indices receive weights of at least 0.5, and secondary complements range from approximately 0.15 to 0.3, whereas proxies receive weights of 0.05 or less. This structure emphasizes the theoretical dominance of geometric descriptors while incorporating human perception and urban climate evidence.

\noindent \textbf{Supplementary Explanation for the Scientific Rationale and the Weighting Schemes.}

\noindent \textbf{Sky View Factor and Sky Visibility Methodology.} The Sky View Factor (SVF) quantifies the hemispherical sky fraction visible from a given point. This geometric metric has become foundational in urban climatology~\cite{liang2017automatic,Bohong_Zheng_2022} with applications covering energy efficiency~\cite{Jaehyun_Ha_2016,Joseph_Appelbaum_2022} and solar potential analysis~\cite{hodul2016estimation}. Our computational methodology follows established geometric algorithms for hemispherical sky visibility calculations, incorporating digital surface model data for accurate obstruction modeling.

However, SVF captures primarily vertical openness. Recent reviews note that it may overestimate perceived openness when lateral obstructions are significant~\cite{miao2020review,middel2018sky,dirksen2019sky}. Street canyon studies link lower SVF values to stronger horizon obstruction and heat island intensity~\cite{unger2004intra,oke1981canyon}. Yet lateral barriers such as building walls, tree belts, and near field structures can block sky access even where SVF remains high.

We therefore model sky visibility as a composite measure governed by SVF but modulated by local land cover and building configuration. This approach better approximates perceptual sky access by combining SVF with land cover derived building occupancy and edge based penalties.

To consider perceptual sky visibility, we introduce two stabilizing terms. First, a \emph{building penalty} based on normalized building occupancy in a local window approximates sub hemispherical occlusions from nearby massing, consistent with obstruction aware SVF estimators~\cite{zeng2018fast, gong2018mapping}. Second, an \emph{edge penalty} discourages boundary adjacent points where horizon estimation and classification transitions are unstable due to façade edges and material transitions.

The resulting score \( V = 0.7\,\mathrm{SVF} - 0.3\,\mathrm{BuildingPenalty} - w_e\,\mathrm{EdgePenalty} \) favors locations that jointly maximize sky access and minimize the immediate occlusion risk. We use a small edge penalty coefficient (\(w_e\in\{0.05,\,0.025\}\)) to keep skyline roughness as a secondary modifier consistent with the weighting scheme above.

\noindent \textbf{Spatial openness (Openness Assessment).} Perceived spatial openness reflects the balance of sky access, low obstruction, gentle terrain, and low visual clutter. We first derive an \emph{OpennessIndex} from the mean SVF and its variability. This index serves as the dominant geometric descriptor in the openness score.

We then combine this index with the mean SVF, the complement of building density, terrain flatness, and a visual simplicity term derived from edge variance. These components use the weights described above. Prior work connects SVF and obstruction to perceived spaciousness and comfort in streetscapes~\cite{Jaehyun_Ha_2016, li2020two} and relates terrain smoothness to functional and visual quality~\cite{hoechstetter2008effects}. Our exponential penalties for height variability and edge variance reflect the diminishing openness under rugged relief and textural clutter.

\noindent \textbf{Visibility range.} The visibility range extends beyond local openness to incorporate the geometric sightline length across the terrain. Standard viewshed analysis on DSMs models horizon blocking and line of sight constraints. This approach is widely used for environmental monitoring and path planning~\cite{pan2020novel}.

Landscape visibility surveys emphasize that horizon structure and local relief shape visible extent and perceived view quality~\cite{wu2023visibilitysurvey}. These studies recommend multi criteria formulations that combine viewshed extent with openness proxies. We therefore mix the viewshed (60\%), local SVF context (25\%), and terrain roughness (15\%) to balance the raw sightline reach, occlusion likelihood, and terrain comfort.

\noindent \textbf{Urban density.} Urban density is characterized by coverage and volume based metrics. The building coverage ratio (BCR) and floor area ratio (FAR) are canonical in urban planning and daylight regulation. We complement these with an SVF component (\(1-\overline{\mathrm{SVF}}\)) that captures enclosure from local massing. This term is clamped to zero in nature dominant regions to avoid conflating canopy cover with urban density.

Prior studies have connected building density, enclosure, and visual or environmental quality~\cite{liu2023interpretable, abarca2019urban, frank2013aesthetics}. Edge density and RGB brightness modestly increase the density score as proxies for fragmentation and surface shadowing. Both receive small weights, consistent with their role as weak modifiers in the index.

\subsubsection{Metric Computation Framework}

Each category employs explicit scoring functions based on standard geospatial metrics:
\begin{itemize}
\item \textbf{Urban Density}: \( D = 0.5 \, \mathrm{BCR} + 0.25 \, \mathrm{FAR} + 0.15 \, \mathrm{SVF\_comp} + 0.05 \, \mathrm{EdgeDensity} + 0.05 \, (1-\mathrm{Brightness}) \), where \(\mathrm{SVF\_comp}=1-\overline{\mathrm{SVF}}\) with a correction to 0 under nature-dominant regions
\item \textbf{Openness Assessment}: \( O = 0.5 \, \mathrm{OpennessIndex} + 0.25 \, \overline{\mathrm{SVF}} + 0.15 \, (1-\mathrm{BuildingDensity}) + 0.05 \, \mathrm{TerrainFlatness} + 0.05 \, \mathrm{VisualSimplicity} \)
\item \textbf{Sky Visibility}: \( V = 0.7 \, \mathrm{SVF} - 0.3 \, \mathrm{BuildingPenalty} - w_e \, \mathrm{EdgePenalty} \), with \(w_e\in\{0.05,\,0.025\}\) for standard/hard settings
\end{itemize}

\noindent To evaluate the reliability of these composite metrics and the derived results, we combined statistical validation with multimodal visual inspection. We performed correlation and sensitivity analyses between the indices and their constituent SVF, DSM, and land cover components. We also manually inspected overlaid RGB, DSM, SVF, and segmentation maps for representative scenes to verify that high or low scores corresponded to intuitively open or enclosed locations.

 \paragraph{Qualitative weight sensitivity.}
\label{sec:appendix_weight_sensitivity}
To assess the robustness of these handcrafted weights, we ran a small sensitivity study on the \textit{sky\_visibility} metric by perturbing individual coefficients by \(\pm 0.1\) (e.g., the SVF weight, building-penalty weight, window size normalization, and edge-penalty weights) over 30 randomly sampled questions. Across all perturbations, the top-1 region changed in at most 6.7\% of the sampled cases (with many settings exhibiting 0\% change), indicating that the discrete argmax decisions were qualitatively stable under modest weight variations.

Parameterization and derived terms
\begin{itemize}
\item \(\mathrm{OpennessIndex} = (\overline{\mathrm{SVF}} + 0.5\,\mathrm{std(SVF)}) / 1.25\)
\item \(\mathrm{TerrainFlatness} \approx \exp(-\mathrm{std(height)}/5.0)\), \quad \(\mathrm{VisualSimplicity} \approx \exp(-\mathrm{edge\_var}/500)\)
\item \(\mathrm{BuildingPenalty} = 0.3\, \mathrm{building\_ratio\_in\_window} \times \mathrm{norm}\) (window normalization)
\item \(\mathrm{Viewshed}\) is the mean of max line-of-sight distances over 8 directions normalized by an upper bound; \(\mathrm{TerrainRoughness} \approx \exp(- (\mathrm{std(height)}-10)^2/(2\cdot 10^2))\)
\item \(\mathrm{FAR} \approx \mathrm{BCR} \times \mathrm{avg\_floors}\), \(\mathrm{avg\_floors} \approx \min(20, \mathrm{height}/3.5\,\mathrm{m})\), normalized by 5
\end{itemize}

The mapping from height (in meters) to the approximate number of stories adopts a representative 3.5\,m per floor, which is commonly used in urban analytics for large-scale building-height approximation~\cite{Usui2022Comparison}.
\subsubsection{Task Category Definitions}

We summarize the benchmark categories according to the three-tier taxonomy. 

\noindent \textbf{Tier 1 (T1): Single-Feature Analysis}

\textbf{Only SVF Categories:}
\begin{itemize}
\item \textbf{Sun Exposure}: Identification of locations with highest solar exposure potential based on SVF values
\item \textbf{Region Ranking}: Arrangement of regions by their openness levels from highest to lowest
\item \textbf{Regional SVF variability}: Identification of regions with highest SVF standard deviation
\item \textbf{Average SVF value}: Calculation of precise average SVF values for specific areas (1-decimal precision)
\end{itemize}

\textbf{Only landcover map Categories:}
\begin{itemize}
\item \textbf{Landcover type}: Identification of land-use types present in the image using landcover maps
\item \textbf{Land Use}: Analysis of predominant landcover usage in specific areas
\end{itemize}

\textbf{Only DSM Categories:}
\begin{itemize}
\item \textbf{Height inference}: Calculation of average height in images using DSM data (10 m precision)
\item \textbf{Highest Region}: Identification of highest elevation locations using DSM
\end{itemize}

\noindent \textbf{Tier 2 (T2): Multi-Feature Analysis}

\textbf{SVF + landcover map Categories:}
\begin{itemize}
\item \textbf{Sky Visibility}: Assessment of `perceived' unobstructed sky view using SVF and landcover data
\end{itemize}
\textbf{SVF + DSM Categories:}
\begin{itemize}
\item \textbf{Visibility Range}: Determination of locations with longest visibility range using multiple modalities
\end{itemize}
\textbf{SVF + DSM + landcover map Categories:}
\begin{itemize}
\item \textbf{Spatial Openness}: Assessment of most expansive open locations considering SVF, terrain, and landcover
\item \textbf{Building Density}: Evaluation of building concentration using SVF penalties and building height data
\end{itemize}

\noindent \textbf{Tier 3 (T3): Free-form Caption}
\begin{itemize}
\item \textbf{Urban Development/land use Application}: Analysis of urban development potential and recommendations for improving scenic quality, safety, and human–natural coexistence
\item \textbf{Renewable Energy Installation}: Assessment of potential for solar panel and wind power generation installation
\item \textbf{Landscape Analysis}: Comprehensive analysis of landscape characteristics including sky visibility, terrain, and landcover types
\item \textbf{Water accumulation}: Analysis of water accumulation risk based on terrain and ground characteristics
\end{itemize}

\begin{table*}[t]
\centering
\footnotesize
\begin{tabular}{@{}p{3.5cm}p{5cm}p{3.5cm}p{3cm}p{2.5cm}@{}}
\toprule
\textbf{Category} & \textbf{Scoring Function} & \textbf{Question Type} & \textbf{Modalities} & \textbf{ID} \\
\midrule
\multicolumn{5}{c}{\textbf{Tier 1}} \\
\midrule
Sun Exposure & 100\% SVF value & Multiple choice & SVF only & (sun\_exposure) \\
Average SVF value & Mean SVF of a region & Numeric (1-dec.) & SVF only & (SVF\_value) \\
Region Ranking & Multi-region SVF ranking & near multiple choice (format specified) & SVF only & (region\_ranking) \\
Regional SVF variability & Regional SVF std/coverage & Multiple choice & SVF only & (regional\_svf\_variability) \\
Land cover type & N/A (categorical) & Multi-label & Segmentation only & (landcover\_type) \\
Land use & N/A (categorical) & Multi-label & Segmentation only & (top\_land\_uses) \\
Height inference & Mean height (10 m bins) & Numeric ("X m") & DSM only & (height\_average) \\
Highest Region & Max elevation & Multiple choice & DSM only & (highest\_region) \\
\midrule
\multicolumn{5}{c}{\textbf{Tier 2}} \\
\midrule
Sky Visibility & \(0.7\,\mathrm{SVF} - 0.3\,\mathrm{BuildingPenalty} - w_e\,\mathrm{EdgePenalty}\) & Multiple choice & SVF + Segmentation & (sky\_visibility) \\
Spatial Openness & \(0.5\,\mathrm{OpennessIndex} + 0.25\,\overline{\mathrm{SVF}} + 0.15\,(1-\mathrm{BuildingDensity}) + 0.05\,\mathrm{TerrainFlatness} + 0.05\,\mathrm{VisualSimplicity}\) & Multiple choice & SVF + DSM + Segmentation & (spatial\_openness) \\
Visibility Range & Viewshed (60\%) + SVF (25\%) + terrain (15\%) & Multiple choice & SVF + DSM & (visibility\_range) \\
Building Density & \(0.5\,\mathrm{BCR} + 0.25\,\mathrm{FAR} + 0.15\,\mathrm{SVF\_comp} + 0.05\,\mathrm{EdgeDensity} + 0.05\,(1-\mathrm{Brightness})\) & Multiple choice & SVF + DSM + Segmentation & (building\_density) \\
\bottomrule
\end{tabular}
\caption{Category taxonomy with scoring functions, output formats, modality requirements, and identifiers.}
\label{tab:category_taxonomy_details}
\end{table*}

\subsection{Evaluation Templates and Coordinate System}
\label{sec:appendix_eval_prompts}

We standardized the evaluation prompts across all models. All coordinates were normalized to a percentage in the range [0,100].

\noindent \textbf{Coordinate Guide:}
\begin{itemize}
  \item Point: (x\%, y\%) where (0,0) is top-left. x is horizontal and y is vertical.
  \item Region: [xmin\%, ymin\%, xmax\%, ymax\%] defining a rectangle; require xmin $<$ xmax, ymin $<$ ymax.
\end{itemize}

\begin{figure*}[t]
  \centering
  \includegraphics[width=\textwidth]{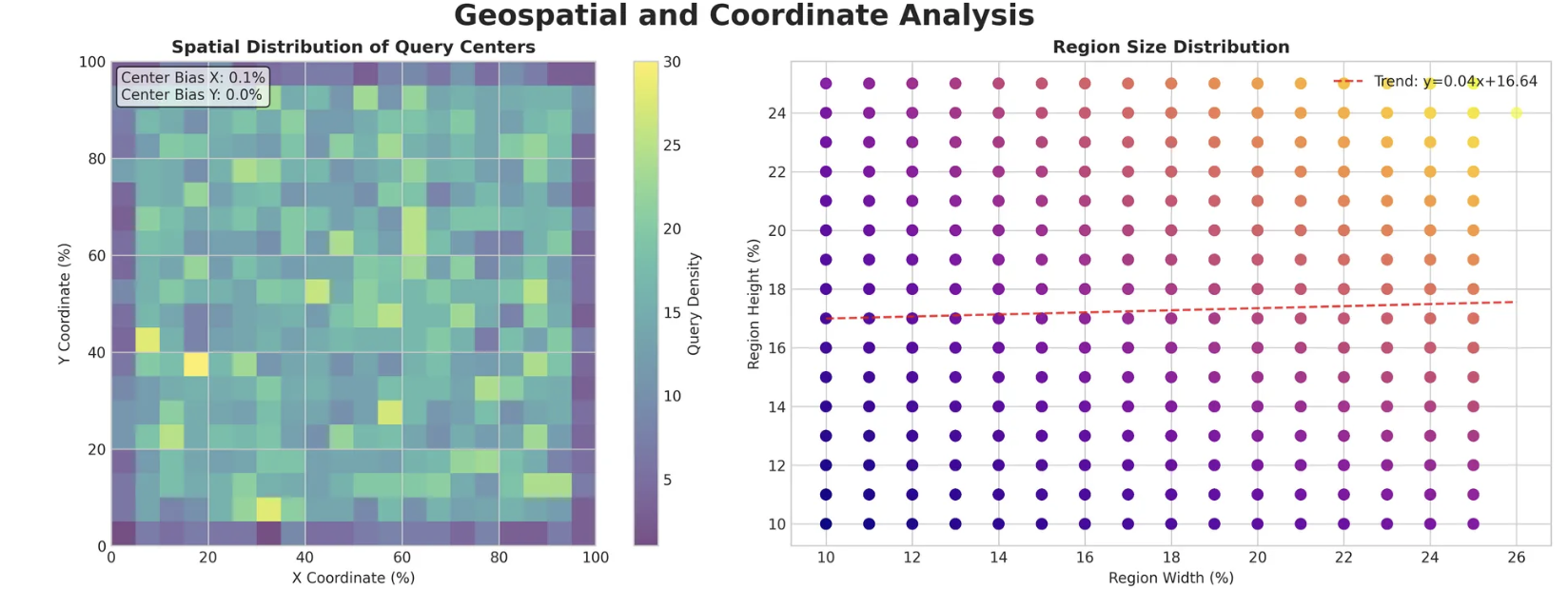}
  \caption{Coordinates of the options were uniformly sampled in the two-dimensional space, and the bounding boxes were selected to cover a wide range of scales. Points are denoted as $(x\%, y\%)$ with $(0,0)$ at the top-left corner; regions are represented as $[x_{\min}\%, y_{\min}\%, x_{\max}\%, y_{\max}\%]$ with $x_{\min} < x_{\max}$ and $y_{\min} < y_{\max}$.}
  \label{fig:vis_coordinates}
\end{figure*}

\noindent \textbf{Prompt-embedded scoring definitions.} For all Tier 2 categories (\textit{visibility\_range}, \textit{sky\_visibility}, \textit{spatial\_openness}, and \textit{building\_density}), we explicitly embedded the scoring method and coordinate system description in the user prompt to remove ambiguity. The scoring definitions are exactly aligned with those in Table~\ref{tab:category_taxonomy_details}.

\noindent \textbf{Format-Constrained Answers.} We specified strict target formats to minimize invalid outputs, but the evaluation used a tolerant parser rather than treating each deviation as an error. The parser first normalizes whitespace and case, canonicalizes comma-separated lists, and strips common reasoning prefixes or suffixes to recover the final answer string. Outputs were only counted as formatting errors when no valid answer could be reliably extracted (e.g., missing numeric value, no region/point selected, or responses that contained only meta-reasoning without a final answer).
\begin{itemize}
  \item Numeric (SVF mean): X.X in [0.0,1.0], 1 decimal.
  \item Height: "X m" with 10-meter increments.
  \item Category: exact label strings from the provided vocabulary (case-sensitive).
  \item Ranking: "Region X, Region Y, Region Z" (comma-separated, exact case).
  \item Region choice: "Region A"/"Region B"/"Region C"/"Region D" only.
  \item Point choice: "Point (x.x\%, y.y\%)" exactly as listed.
  \item Balanced choices: correct-option indices are balanced across A/B/C/D and choices are shuffled without positional bias.
  \item Diversity: candidate selection enforces minimal similarity to avoid ambiguous ties.
\end{itemize}

\subsubsection{Prompt paraphrasing robustness}
\label{sec:appendix_prompt_paraphrasing}

To evaluate robustness to natural language variations, we created a paraphrased version of the benchmark in which GPT-4 rewrote each question based on the same task templates.
We manually verified a random subset of $\sim$100 items to ensure that the paraphrases preserved the original semantics and the answer labels.
Using this paraphrased set, we re-ran the evaluation for our best RGB-only model, Qwen2.5-VL-7B fine-tuned with the 100K+free recipe. 

Table~\ref{tab:prompt_paraphrasing_robustness} summarizes the major-category accuracies for Qwen2.5-VL-7B Base and the fine-tuned 100K+free model under the original vs paraphrased prompts.
Paraphrasing moderately degraded the fine-tuned model (approximately 9.0 pp drop in overall accuracy), but instruction tuning still yielded large gains over the baseline model under the original templates (approximately +20--40 pp across SVF, height, LULC, multi-factor, and overall).
Even under paraphrased prompts, the fine-tuned model remained clearly above the baseline model evaluated using the standard templates.
The format-error rate remains negligible (0.09\% overall; 0.91\% only in \textit{landcover\_type}), indicating that our tolerant parser and format constraints generalize well to linguistically diverse prompts.

\begin{table}[t]
    \centering
    \footnotesize
    \begingroup\setlength{\tabcolsep}{3pt}\renewcommand{\arraystretch}{0.9}
    \begin{tabular}{@{}l|ccccc@{}}
    \toprule
    \textbf{Setup} & \textbf{SVF} & \textbf{Height} & \textbf{LULC} & \textbf{Multi} & \textbf{Overall} \\
    \midrule
    Qwen Base (std) & 23.5 & 21.3 & 45.3 & 22.6 & 26.4 \\
    FT (100K+free, std) & 44.6 & 41.1 & 71.3 & 45.5 & 48.7 \\
    FT (100K+free, para) & 34.5 & 34.5 & 64.2 & 38.9 & 39.7 \\
    FT(para) $-$ FT(std) & -10.1 & -6.6 & -7.1 & -6.6 & -9.0 \\
    \bottomrule
    \end{tabular}
    \endgroup
    \caption{Prompt paraphrasing robustness for Qwen2.5-VL-7B FT (100K+free).
    Accuracy (\%) for major-category summaries and overall accuracy under the original evaluation templates vs GPT-4-paraphrased prompts. "std" stands for standard templates, "para" stands for paraphrased prompts.
    LULC denotes the combined land use/land cover major category and Multi denotes the multi-factor major category (cf.\ Table~\ref{tab:qwen_comprehensive_comparison_app}).}
    \label{tab:prompt_paraphrasing_robustness}
\end{table}

\section{Extended Discussion and Worked Examples}
\label{sec:appendix_examples}

We provide extended discussions and worked examples of the benchmark tasks.

\subsubsection{Feasibility and Scope of RGB-to-3D Reasoning}
\label{sec:appendix_feasibility_extended}

\noindent \textbf{Feasibility.}
While precise metric 3D estimation from monocular RGB is ill-posed owing to scale ambiguity, prior work has shown that \emph{coarse} height and relative elevation cues can be learned from 2D indicators such as shadow geometry, occlusion boundaries, perspective distortions, and structural regularities~\cite{Cai2020Monocular,Lu2021SGTBN,Lourenço2021Intel,liu2020im2elevation,liasis2016satellite}. Geo3DVQA targets this feasible regime.

\noindent \textbf{Scope of the benchmark.}
We evaluate height-aware reasoning at decision-oriented granularity rather than full 3D reconstruction.
\begin{itemize}
  \item \textbf{Categorical height distinctions:} accuracy is calculated on 10 m bins.
  \item \textbf{Relative comparisons:} which region is higher; which point has more sky access.
  \item \textbf{SVF pattern recognition:} open vs. enclosed spaces from RGB-visible urban density cues.
  \item \textbf{Multi-feature integration:} combining height, openness, and land cover for composite spatial metrics.
\end{itemize}

\noindent \textbf{Practical applications.}
This coarse level supports (i) urban heat risk screening (tall + low SVF zones), (ii) solar siting pre-screening via sky openness, (iii) evacuation planning through open corridors and cluster identification, and (iv) accessibility planning via coarse terrain opennesswithout requiring specialized sensors. These applications prioritize actionable stratification over centimeter-level accuracies.

\noindent \textbf{Geographic generalization and scalable deployment.} As acknowledged in the main text limitation, our evaluation is currently limited to North Rhine-Westphalia (NRW) in Germany, which constrains generalization claims. However, the framework's design principles support scalable deployment and appropriate geographic expansion. The current GeoNRW limitation reflects data availability constraints rather than fundamental methodological limitations. For scalable real-world deployment, the following requirements should be addressed: (1) \emph{Geographic diversity}: training data should encompass multiple regions with varying urban morphologies, terrain types, and climate zones to reduce geographic bias; (2) \emph{Seasonal and temporal variation}: incorporating multi-temporal imagery across different seasons would improve robustness to phenological changes and weather conditions; and (3) \emph{Adaptive fine-tuning}: region-specific fine-tuning or few-shot adaptation could bridge domain gaps when deploying to new geographic areas. Within the current NRW constraint, the framework has already demonstrated practical utility for prescreening, coarse stratification, and initial risk assessment tasks that do not require centimeter-level precision. The scalable deployment pathway involves the systematic expansion of training data across diverse geographic contexts, which is a natural next step for production systems rather than a fundamental limitation of the approach.




\subsubsection{Dataset Statistics and Visualization}
\label{sec:appendix_dataset_statistics}

\begin{figure*}[t]
    \centering
    \begin{subfigure}[t]{0.48\textwidth}
        \centering
        \includegraphics[width=\linewidth]{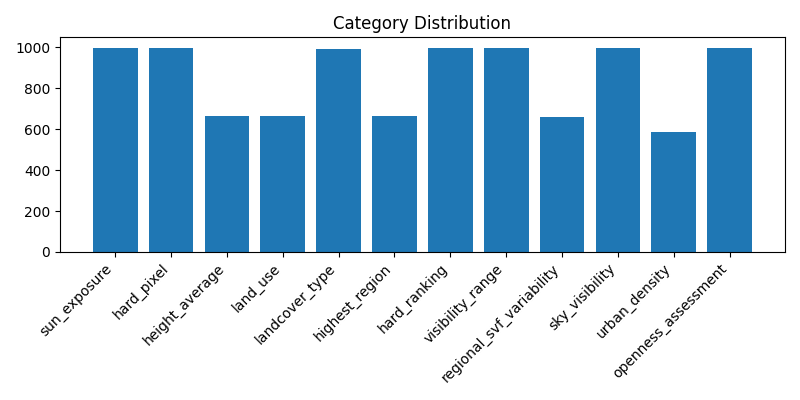}
        \caption{Per-category question counts (train/test).}
        \label{fig:stat_categories}
    \end{subfigure}\hfill
    \begin{subfigure}[t]{0.48\textwidth}
        \centering
        \includegraphics[width=\linewidth]{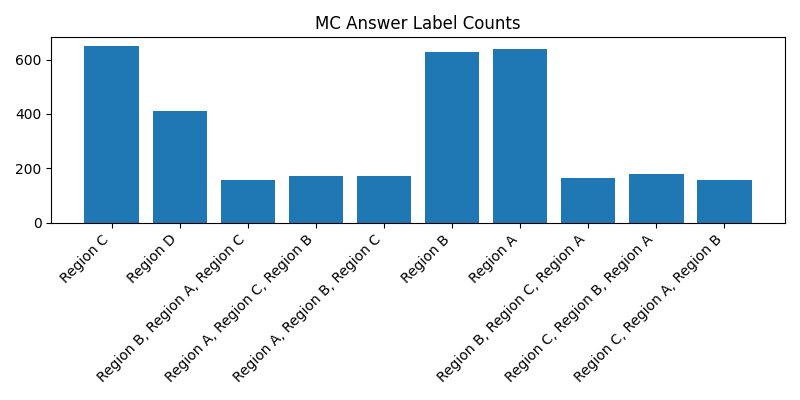}
        \caption{Correct label frequencies in multiple-choice.}
        \label{fig:stat_mc_label}
    \end{subfigure}

    \vspace{0.5cm}

    \begin{subfigure}[t]{0.48\textwidth}
        \centering
        \includegraphics[width=\linewidth]{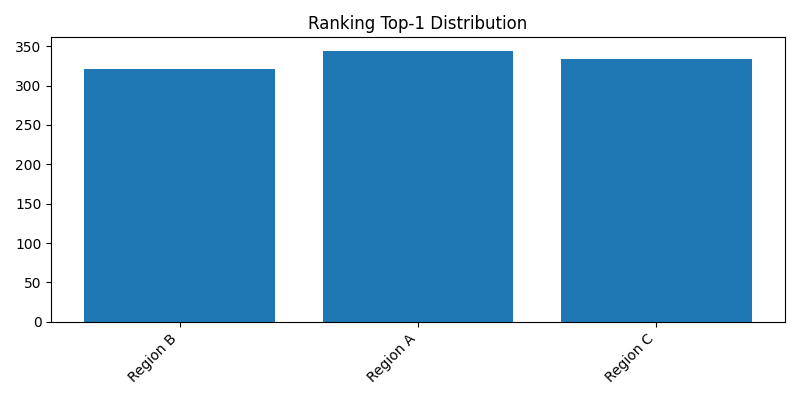}
        \caption{Top-1 region distribution in ranking tasks.}
        \label{fig:stat_ranking_top1}
    \end{subfigure}

    \vspace{0.35cm}

    \caption{Geo3DVQA dataset statistics showing task-level dataset properties: (a) per-category question counts, (b) label frequencies in multiple-choice, (c) Top-1 region in ranking tasks.}
    \label{fig:dataset_statistics}
\end{figure*}

\subsubsection{Model Size Constraints and Future Scaling}
\label{sec:appendix_model_constraints}

All local experiments were bounded to models with at most 10 B parameters owing to server memory limits for both training and inference. This constraint influences our architectural choices and may understate the attainable ceiling for elevation- and SVF-aware reasoning. Future work will investigate the scaling behavior with respect to free-form coherence, format compliance under constrained decoding, and multi-image conditioning for improved coordinate fidelity.

\subsubsection{Input/Output Normalization and Examples}

\noindent \textbf{Input/Output Normalization Rules.}
\label{sec:appendix_io_rules}
\begin{itemize}
  \item Coordinates and Regions: See the Coordinate Guide for normalized coordinate definitions and region format.
  \item Output labels: fixed vocabulary for LU/LC; case-sensitive matching.
  \item Numerical outputs: clipped to valid ranges and rounded to specified precision.
\end{itemize}

\section{Feasibility and Learnability Analysis}
\label{sec:appendix_feasibility_learnability}

\subsection{Feasibility and Learnability}
\noindent \textbf{Chance Level vs. Performance.}
While an overall accuracy of $\sim$50\% might seem moderate, it must be contextualized against the random-chance level. For 4-choice questions, chance is 25\%, and for multi-label or ranking tasks, it is significantly lower. The performance of the fine-tuned model (approx. double the chance level) confirms that it has learned meaningful geospatial patterns rather than relying on hallucinations.

\noindent \textbf{Relative vs. Absolute Reasoning.}
Our analysis shows that the models perform better on relative tasks (e.g., Ranking, Highest Region) than on absolute metric regression. This aligns with the inherent ambiguities of monocular vision. For urban planning applications, relative accuracy (e.g., ``Area A is denser than Area B'') is often sufficient for initial screening, supporting the practical feasibility of the approach despite the limitations of absolute metric precision.

\noindent \textbf{Semantic-Geometric Integration.}
A key advantage of VLMs is their ability to leverage semantic context. For instance, a pure depth model could have difficulty distinguishing a flat gray roof from gray pavement if height cues are ambiguous. However, a VLM can recognize an object as a "school building" based on contextual cues, such as shape and the presence of a playground, and infer that it must be elevated. This semantic-geometric integration allows the model to resolve ambiguities that purely geometric methods cannot, which further supports the end-to-end paradigm.

\noindent \textbf{Worked Examples.}


\begingroup
\setlength{\belowcaptionskip}{0pt}
\captionsetup{skip=4pt}
\begin{figure*}[!t]
\centering
\begin{subfigure}{0.48\textwidth}
  \includegraphics[width=\textwidth]{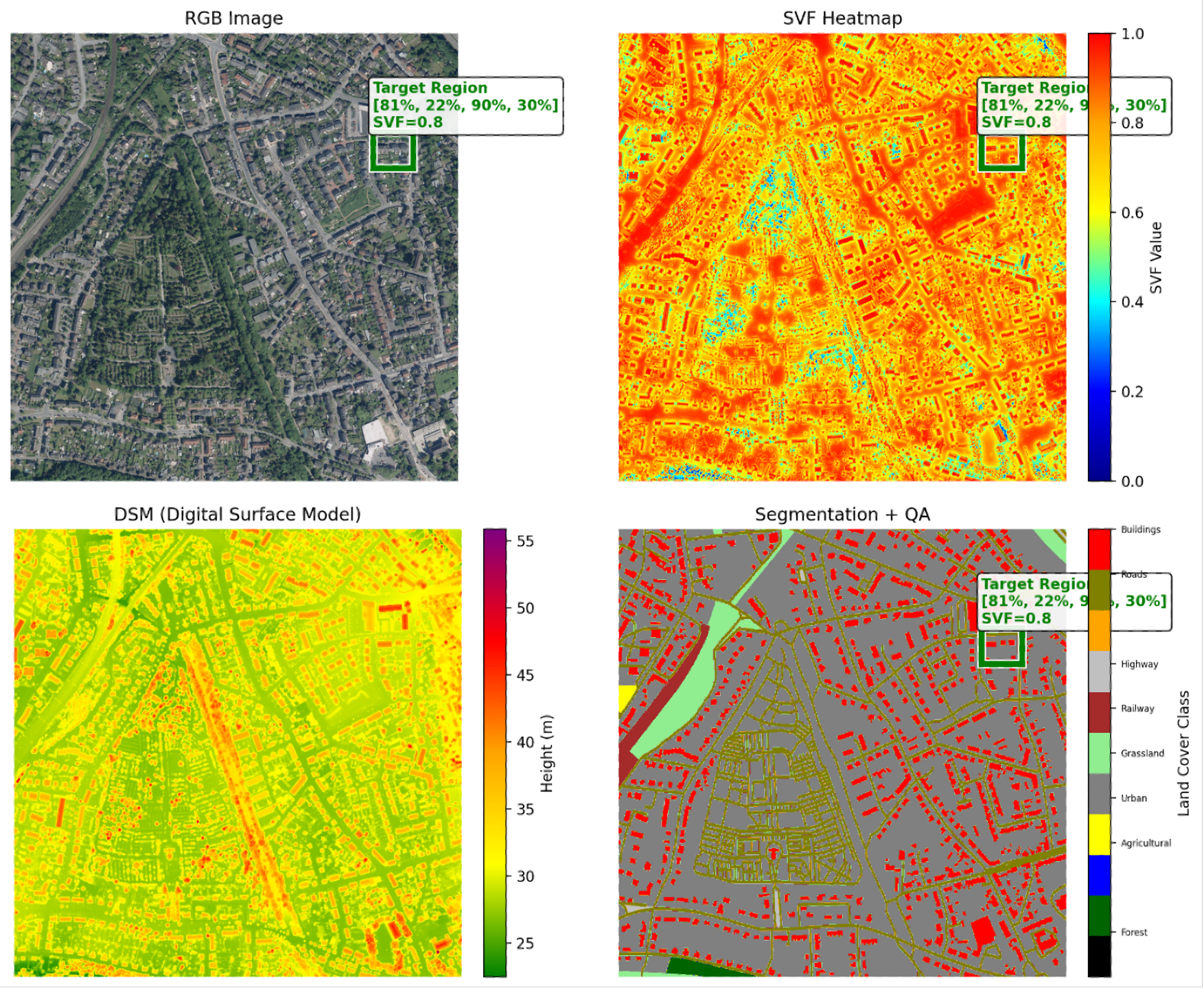}
  \caption{\textbf{Q30 Mean SVF (Region)}. Q: Calculate the mean SVF within \textcolor{darkgreen}{[81\%, 22\%, 90\%, 30\%]}. A: 0.8}
  \label{fig:appendix_q030}
\end{subfigure}
\hfill
\begin{subfigure}{0.48\textwidth}
\includegraphics[width=\textwidth]{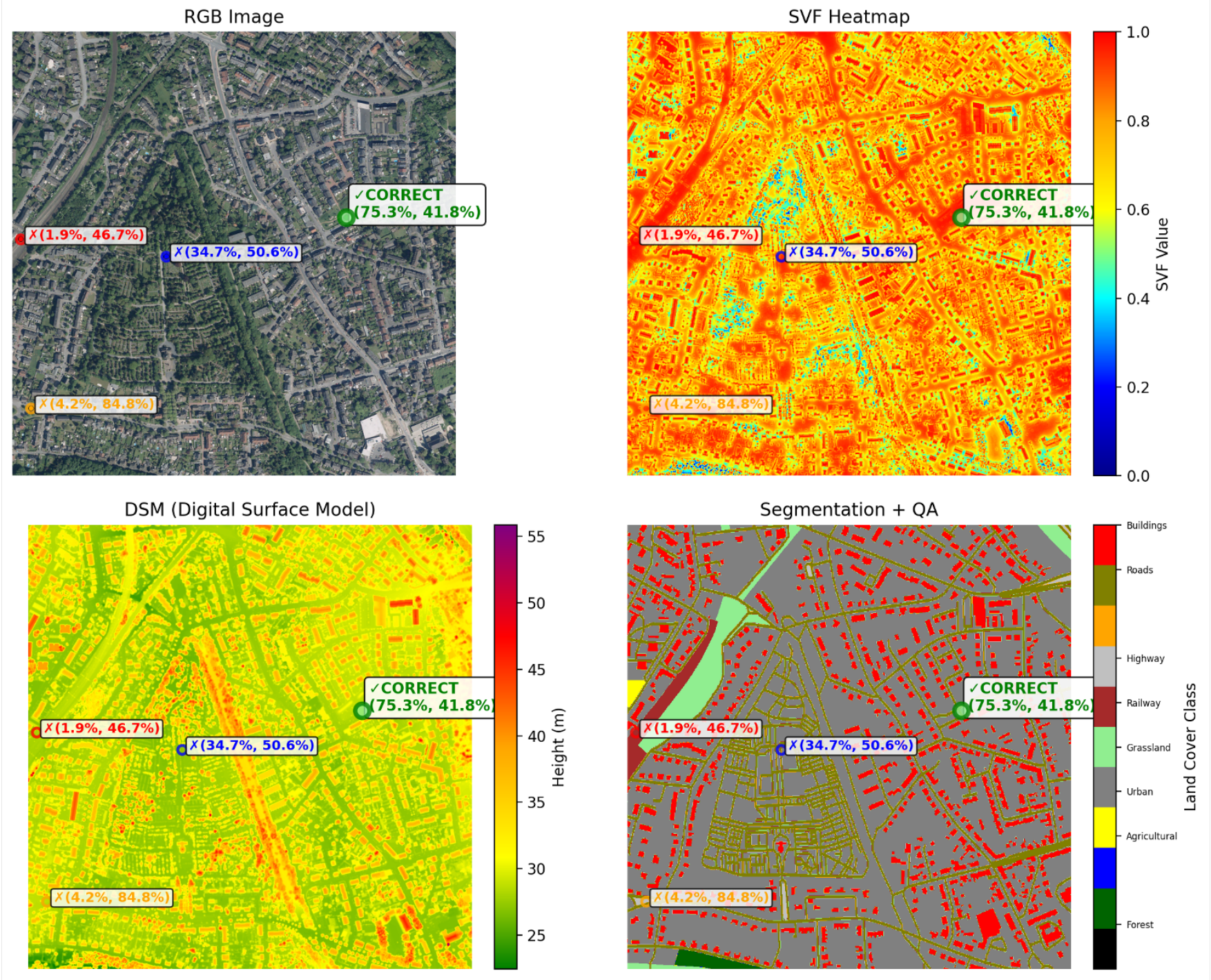}
\caption{\textbf{Q8 Visibility Range}. Q: Which location has the most comprehensive sightlines? A: \textcolor{darkgreen}{Point (75.3\%, 41.8\%)}}
\label{fig:appendix_q008}
\end{subfigure}
\begin{subfigure}{0.48\textwidth}
\includegraphics[width=\textwidth]{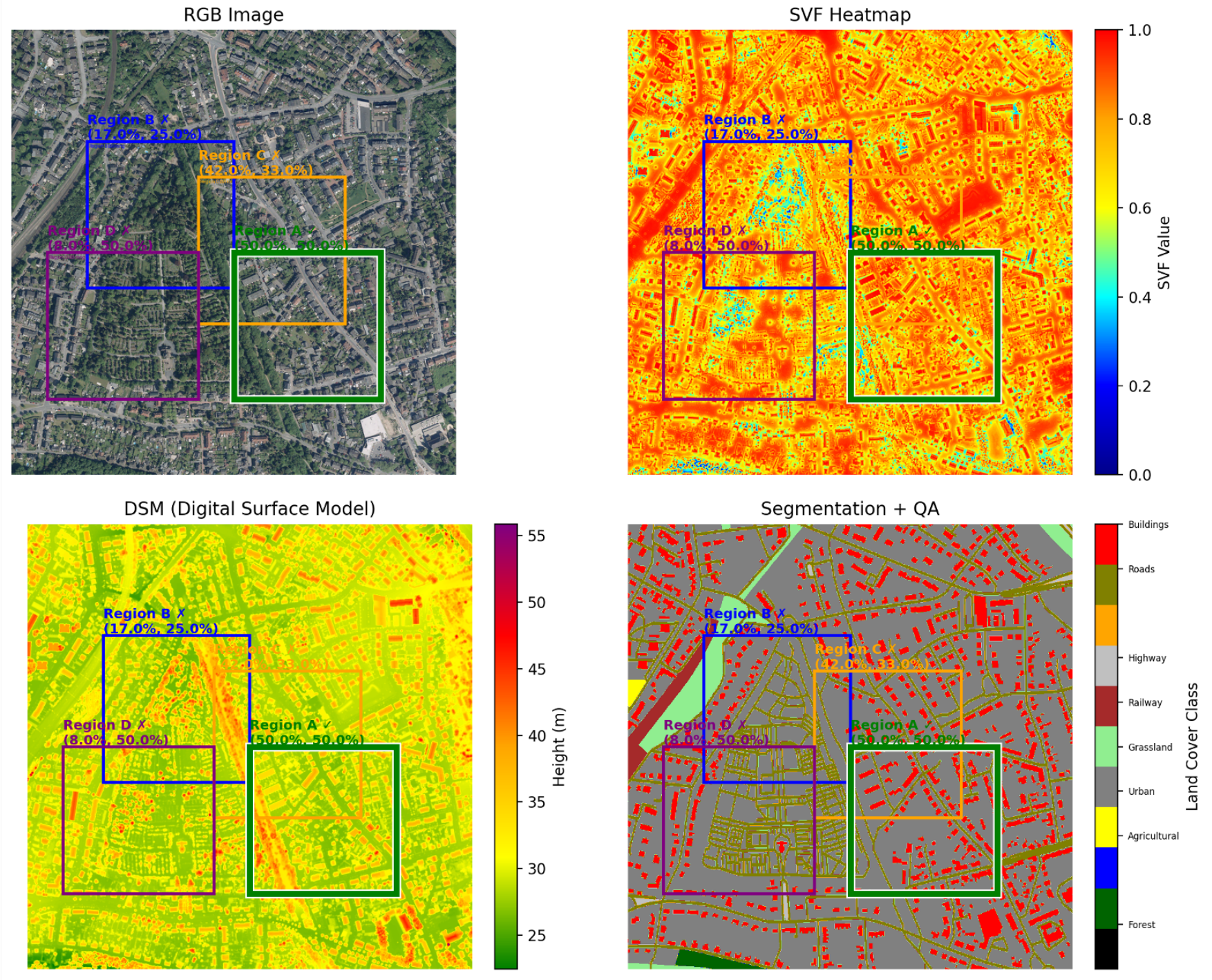}
\caption{\textbf{Q18 Urban Density}. Q: Which area has the most crowded urban layouts? A: \textcolor{darkgreen}{Region A}}
\label{fig:appendix_q018}
\end{subfigure}
\hfill
\begin{subfigure}{0.48\textwidth}
\includegraphics[width=\textwidth]{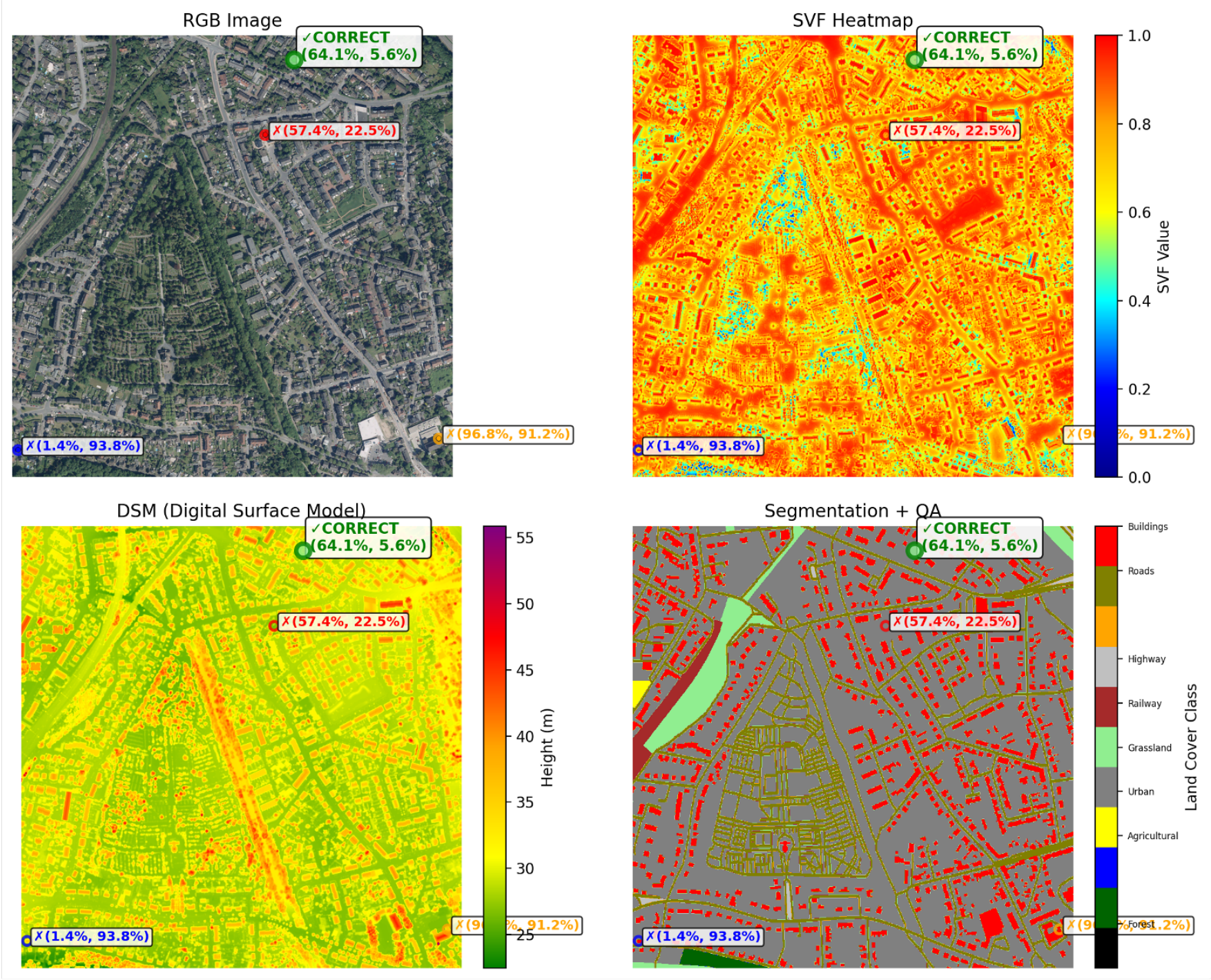}
\caption{\textbf{Q22 Sky Visibility}. Q: Where is sky access the most unrestricted? A: \textcolor{darkgreen}{Point (64.1\%, 5.6\%)}}
\label{fig:appendix_q022}
\end{subfigure}
\begin{subfigure}{0.46\textwidth}
\includegraphics[width=\textwidth]{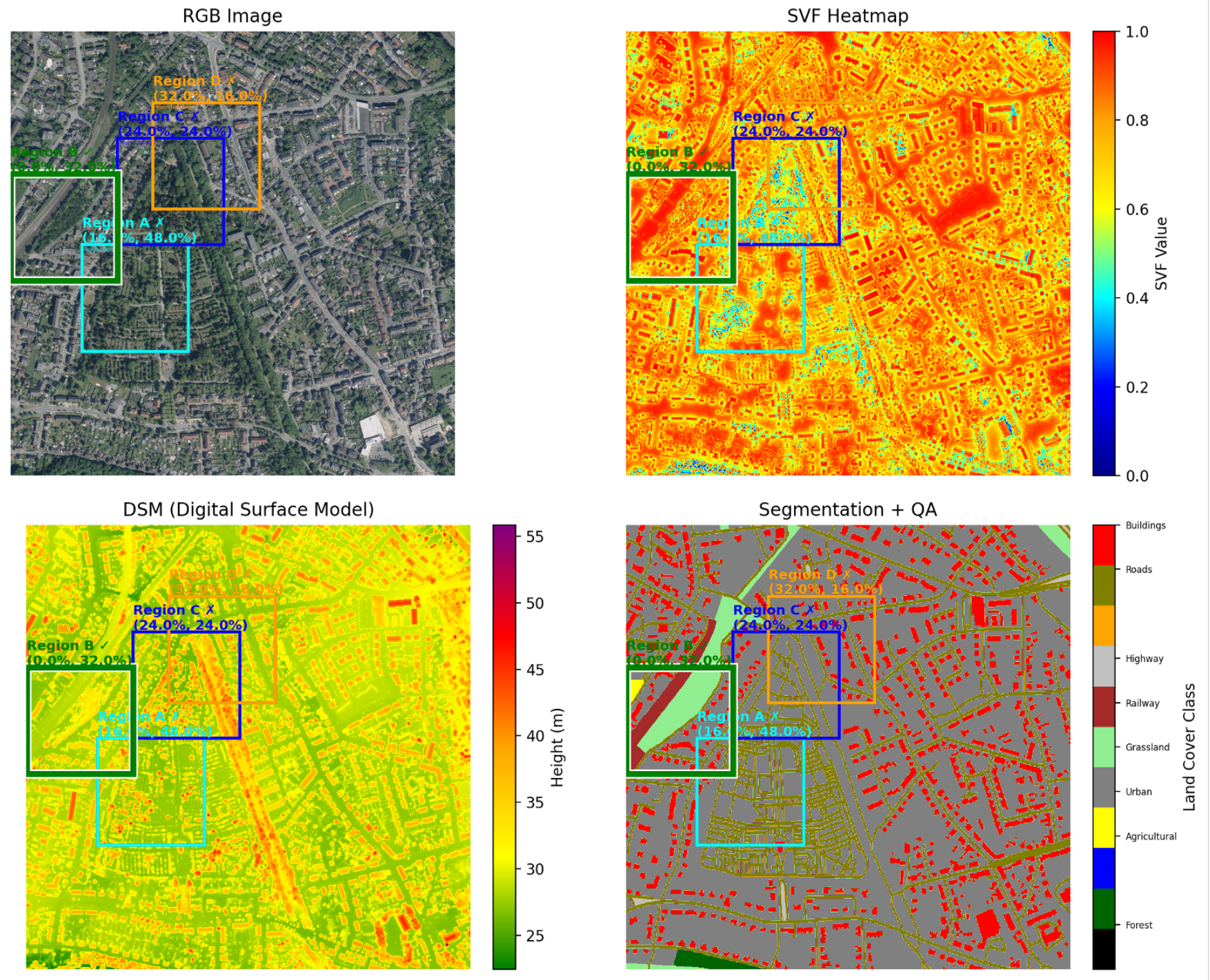}
\caption{\textbf{Q23 Spatial Openness}. Q: Which area demonstrates maximum openness with minimal obstruction? A: \textcolor{darkgreen}{Region B}}
\label{fig:appendix_q023}
\end{subfigure}
\hfill
\begin{subfigure}{0.46\textwidth}
  \includegraphics[width=\textwidth]{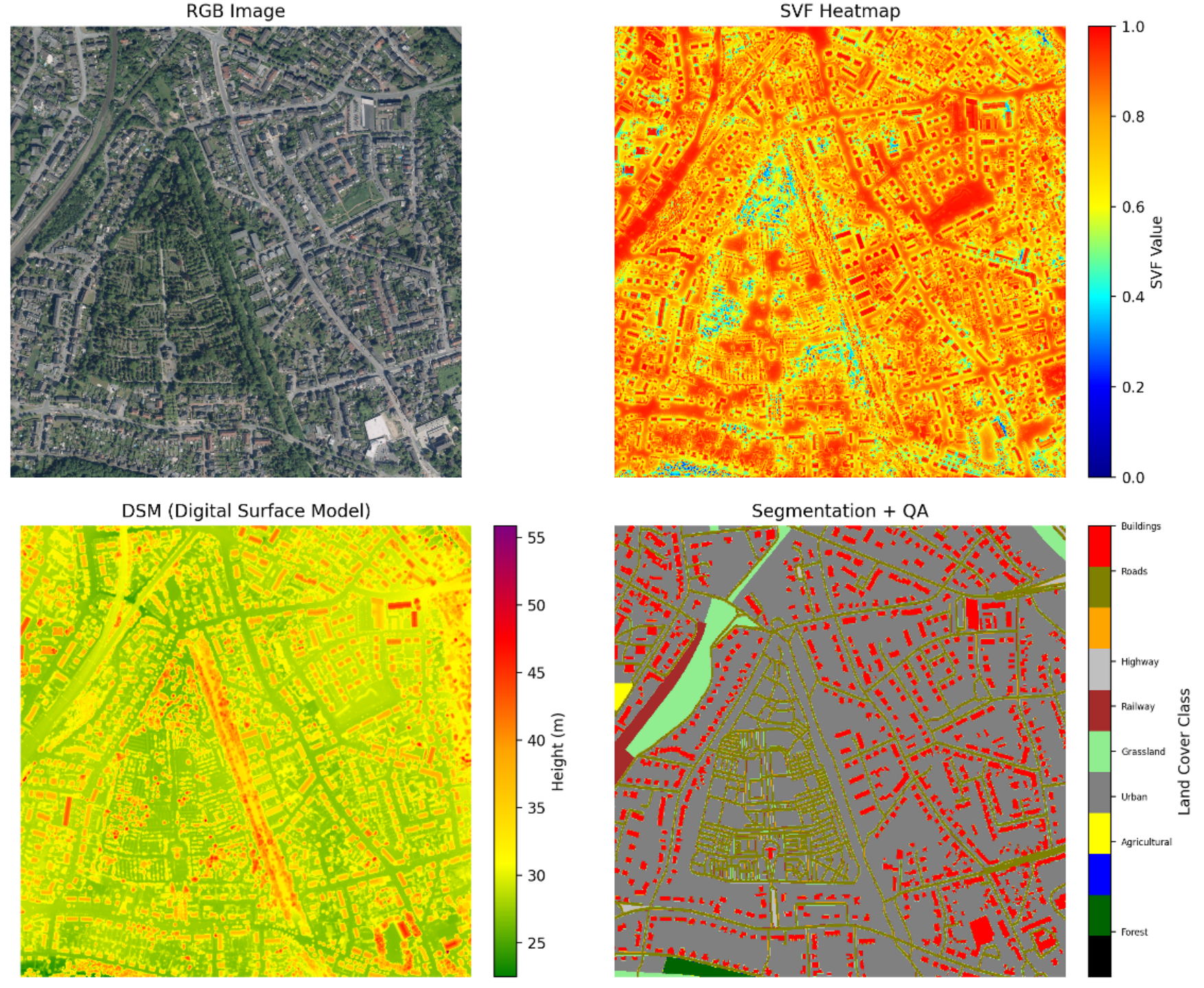}
  \caption{\textbf{Q3 Height Average}. Q: Calculate the mean elevation within [19\%, 47\%, 25\%, 53\%]. A: 30 m}
  \label{fig:appendix_q003}
\end{subfigure}
\vspace{-0.1cm}
\caption{Visualized QA pairs for a sample scene. Each subfigure shows RGB, DSM, SVF, and segmentation overlays. Subcaptions include the question summary and the ground-truth answer.}
\label{fig:appendix_qa_examples_338_5698}
\end{figure*}
\endgroup

\clearpage
\onecolumn
\section*{Prompt and Output Listings}

\begin{lstlisting}[style=geocode,caption={Example statistical input JSON},label={lst:stat_input}]
{
  "scene_id": "geonrw_sample_001",
  "svf_statistics": {
    "mean": 0.4237,
    "std": 0.2841,
    "min": 0.0023,
    "max": 0.9876,
    "quartiles": [0.1892, 0.3945, 0.6234]
  },
  "height_statistics": {
    "mean": 12.34,
    "std": 8.92,
    "building_coverage_ratio": 0.31,
    "max_height": 45.2
  },
  "land_cover_statistics": {
    "building_ratio": 0.28,
    "vegetation_ratio": 0.35,
    "road_ratio": 0.23,
    "water_ratio": 0.02,
    "other_ratio": 0.12
  },
  "spatial_statistics": {
    "edge_density": 0.156,
    "patch_count": 23,
    "largest_patch_ratio": 0.187
  },
  "rgb_statistics": {
    "brightness_mean": 127.3,
    "contrast_std": 31.2,
    "dominant_colors": ["#4A5D23", "#8B7355", "#A3A3A3"]
  },
  "derived_metrics": {
    "urban_density_score": 0.67,
    "openness_index": 0.43,
    "solar_potential": 0.58
  }
}
\end{lstlisting}

\begin{lstlisting}[style=geocode,caption={FT example: Urban Development (QID 1: koeln\_364\_5633\_dem)},label={lst:ft_example_q1}]
Question ID: 1 (koeln_364_5633_dem)
Question (excerpt): Analyze the potential of this area for urban development.

Ground truth sections (excerpt):
<OBSERVATION>Built environment coverage was 20.8%, with vegetation dominating at 66.5% (agricultural 57.4%, natural vegetation 9.2%). Mean SVF is 0.915. Elevation ranges 43.4-76.8 m (delta 33.4 m), relative variation 0.60. Urbanization level was 2/5 and the spatial diversity index was 1.320. The highest sky accessibility was in the middle center, and the best scenic quality was in the middle right.</OBSERVATION>
<CONCLUSION>Low building density and high SVF support comfortable expansion; prioritize top-center infrastructure; preserve middle-right scenic zones; leverage elevation variation for diversified fabric; maintain vegetation >50% with green corridors.</CONCLUSION>

Model sections (FT prediction):
<OBSERVATION> The built environment coverage was 21.5%, with residential areas dominating at 18.7%. Agricultural land covers 69.1% of the area, whereas natural vegetation is minimal at 0.1%. Mean SVF is 0.851. Elevation ranges 48.1-60.1 m (delta 12.0 m), terrain complexity 0.24. The highest sky accessibility top-left; bottom-right offers the best development suitability.</OBSERVATION>
<CONCLUSION>Maintain low building density to preserve scenic quality and sky openness. Prioritize bottom-right development, enhance green corridors by increasing natural vegetation, and improve infrastructure connectivity while maintaining a safe elevation.</CONCLUSION>

Validation summary:
Word count: 131; Scores: Total 3.0, Obs. 3, Conc. 3, SVF 4, LC 3, Elev. 3, Logic 4
\end{lstlisting}

\begin{lstlisting}[style=geocode,caption={FT example: Landscape Analysis (QID 7: guetersloh\_454\_5748\_dem)},label={lst:ft_example_q7}]
Question ID: 7 (guetersloh_454_5748_dem)
Question (excerpt): Analyze the overall landscape in terms of sky visibility, terrain, and land cover.

Ground truth sections (excerpt):
<OBSERVATION>Natural vegetation 32.5%, built 7.8%, forest 28.8%, agriculture 56.7%. Elevation 67.9-100.7 m (mean 76.5 m, delta 32.8 m, var. 0.43). The mean sky openness was 0.783 (0.011-1.000); highest top-right. Spatial diversity index: 1.174; urbanization: 1/5; scenic quality: 0.727.</OBSERVATION>
<CONCLUSION>Natural-agricultural balance with low built density, moderate relief supports connectivity, top-right openness enhances solar and visual quality, conserve forest patches, and focus development top-left to minimize fragmentation.</CONCLUSION>

Model sections (FT prediction):
<OBSERVATION>Natural vegetation 12.5% (forest 11.6%); agriculture 79.8%; built 0.1%. Elevation 100.1-141.1 m (mean 116.1 m, delta 41.0 m, var. 0.35). The mean sky openness was 0.811 (0.011-1.000); highest top-left. The Shannon diversity was 0.641. Scenic quality 0.711.</OBSERVATION>
<CONCLUSION>Agriculture-dominant landscape with limited urbanization; terrain variation supports connectivity; high sky openness enhances visual quality; conserve forest patches to maintain biodiversity.</CONCLUSION>

Validation summary:
Word count: 129; Scores: Total 3.0, Obs. 2, Conc. 3, SVF 4, LC 2, Elev. 3, Logic 3
\end{lstlisting}

\begin{lstlisting}[style=geocode,caption={Worked example (Qwen Base)},label={lst:worked_example}]
Question ID: 1 (kleve_306_5742_dem)
Ground truth observation (excerpt): mean SVF 0.942; elevation 12.8-43.3 m; built 7.0%; bottom-right suitable; middle-left scenic.
Model sections:
<OBSERVATION></OBSERVATION>
<ANALYSIS> ... general discussion without coordinates ... </ANALYSIS>
<CONCLUSION> ... vague recommendations ... </CONCLUSION>
Scores: Total 2.0, Obs. 1, Anal. 3, Conc. 2, SVF 2, LC 2, Elev. 1, Logic 3
\end{lstlisting}

\begin{lstlisting}[style=geocode,caption={Numeric (Region mean SVF)},label={lst:numeric_svf}]
System: You are a precise geospatial assistant. Answer concisely.
User: Consider an RGB image. Answer using the specified format only.
Question: Calculate the mean Sky View Factor within the area
         [65%, 81%, 79%, 95%].
Note: Coordinates are the percentages of the image size [xmin%, ymin%, xmax%, and ymax%].
Answer format: X.X (a single number in [0.0, 1.0], rounded to 1 decimal)
\end{lstlisting}

\begin{lstlisting}[style=geocode,caption={Point Selection (Sky visibility / Visibility range)},label={lst:point_selection}]
System: You are a precise geospatial assistant. Answer concisely.
User: Which location has the highest sky visibility?
Hint: Areas with fewer obstacles have a higher sky view factor.
Please choose from:
Point (89.7%, 20.6%)
Point (81.7%, 37.0%)
Point (11.3%, 3.9%)
Point (57.8%, 9.7%)
Answer format: Point (x.x%, y.y%)
\end{lstlisting}

\begin{lstlisting}[style=geocode,caption={Point Selection (Visibility range with scoring and coordinates)},label={lst:visibility_range_scored}]
System: You are a precise geospatial assistant. Answer concisely.
User: Which location has the most comprehensive sight lines?
Hint: Areas with good visibility typically have a high sky view factor and fewer obstacles in the line of sight.
Scoring method: Locations were scored solely based on viewshed distance analysis (60%), Sky View Factor (25%), and terrain roughness variation (15%). Higher scores indicate a better visibility range with longer line-of-sight distances.
Coordinate system: Each point is specified by (x, y) coordinates as percentages of the image dimensions, where (0, 0) represents the top-left corner. 'x' represents the horizontal position (from left to right), and 'y' represents the vertical position (from top to bottom).
Please choose from:
Point (75.3%, 41.8%)
Point (1.9%, 46.7%)
Point (34.7%, 50.6%)
Point (4.2%, 84.8%)
Answer format: Point (x.x%, y.y%)
\end{lstlisting}

\begin{lstlisting}[style=geocode,caption={Region Ranking},label={lst:region_ranking}]
System: You are a precise geospatial assistant. Answer concisely.
User: Compare the SVF values of the three regions. Reply with
      The order from highest to lowest was labeled A/B/C.
Regions (x1%, y1%, x2%, y2%):
  A: (46%, 60%, 65%, 75%)
  B: (44%, 48%, 63%, 63%)
  C: (28%, 32%, 46%, 48%)
Answer format:
"Region A, Region B, Region C"
\end{lstlisting}

\begin{lstlisting}[style=geocode,caption={Multi-Choice Region Selection},label={lst:multi_choice}]
System: You are a precise geospatial assistant. Answer concisely.
User: Which region best matches this criterion? Choose one.

A: [22%, 67%, 41%, 87%]
B: [77%, 6%, 98%, 27%]
C: [70%, 46%, 91%, 67%]
D: [75%, 12%, 95%, 32%]

Please choose from:
Region A
Region B
Region C
Region D
Answer format:
Region X
\end{lstlisting}

\begin{lstlisting}[style=geocode,caption={Land Use / Landcover (Multi-Label)},label={lst:land_use}]
System: You are a precise geospatial assistant. Answer concisely.
User: Which land use types are most frequent in the region
      (16%, 48%, 40%, 72%)? Choose from the list.
Choices: residential, agricultural, forest, grassland,
         railways, roads, bare_soil, buildings, water, other
Answer format:
Comma-separated labels in lowercase.
\end{lstlisting}

\twocolumn

\endgroup

\section{Data and Code Availability}
\label{sec:appendix_data_code}

\noindent \textbf{Release scope and licensing constraints.}
We are committed to maximizing reproducibility within the GeoNRW data licensing restrictions. Upon publication, we plan to release the following:

\noindent \textit{Publicly released components (unrestricted access):}
\begin{itemize}
  \item Question–answer pairs with image identifiers (110k items) and splits
  \item Evaluation scripts and rubric-based assessors for free-form scoring
  \item Fine-tuned model checkpoints (Qwen2.5-VL-7B) and training code
  \item Documentation of task definitions, metrics, and validation procedures
\end{itemize}

\noindent \textit{Access-restricted components (require GeoNRW license):}
\begin{itemize}
  \item Raw aerial RGB imagery (1 m) and raw DSM/segmentation rasters
\end{itemize}

A project page with links to the repository and model weights will be announced on our GitHub repository.

%% file: references.bib
@article{liu2023visual,
  title         = {{Visual instruction tuning}},
  author        = {Liu, Haotian and Li, Chunyuan and Wu, Qingyang and Lee, Yong Jae},
  journal       = {Advances in neural information processing systems},
  volume        = {36},
  pages         = {34892--34916},
  year          = {2023}
}

@article{li2024llava,
  title={{LLaVA-OneVision}: Easy visual task transfer},
  author={Li, Bo and Zhang, Yuanhan and Guo, Dong and Zhang, Renrui and Li, Feng and Zhang, Hao and Zhang, Kaichen and Zhang, Peiyuan and Li, Yanwei and Liu, Ziwei and others},
  journal={arXiv preprint arXiv:2408.03326},
  year={2024}
}

@article{bai2025qwen2,
  title={{Qwen2.5-VL technical report}},
  author={Bai, Shuai and Chen, Keqin and Liu, Xuejing and Wang, Jialin and Ge, Wenbin and Song, Sibo and Dang, Kai and Wang, Peng and Wang, Shijie and Tang, Jun and others},
  journal={arXiv preprint arXiv:2502.13923},
  year={2025}
}

@article{arias2021climate,
  title={{Climate Change 2021: the physical science basis. Contribution of Working Group I to the Sixth Assessment Report of the Intergovernmental Panel on Climate Change; technical summary}},
  author={Arias, Paola and Bellouin, Nicolas and Coppola, Erika and Jones, Richard and Krinner, Gerhard and Marotzke, Jochem and Naik, Vaishali and Palmer, Matthew and Plattner, Gian-Kaspar and Rogelj, Joeri and others},
  year={2021}
}

@article{unger2004intra,
  title={{Intra-urban relationship between surface geometry and urban heat island: review and new approach}},
  author={Unger, J{\'a}nos},
  journal={Climate research},
  volume={27},
  number={3},
  pages={253--264},
  year={2004}
}

@article{middel2018sky,
  title={{Sky View Factor footprints for urban climate modeling}},
  author={Middel, Ariane and Lukasczyk, Jonas and Maciejewski, Ross and Demuzere, Matthias and Roth, Matthias},
  journal={Urban Climate},
  volume={25},
  pages={120--134},
  year={2018},
  publisher={Elsevier}
}

@article{Li2020Automatic,
  title={{Automatic Indoor as-Built Building Information Models Generation by Using Low-Cost RGB-D Sensors}},
  author={Li, Yaxin and Li, Wenbin and Tang, Shengjun and Darwish, Walid and Hu, Yuling and Chen, Wu},
  journal={Sensors},
  volume={20},
  number={1},
  pages={293},
  year={2020},
  publisher={MDPI}
}

@article{Lu2021SGTBN,
  title={{SGTBN}: Generating Dense Depth Maps From Single-Line LiDAR},
  author={Hengjie Lu and Shugong Xu and Shan Cao},
  journal={IEEE Sensors Journal},
  year={2021},
  volume={21},
  pages={19091-19100},
  doi={10.1109/JSEN.2021.3088308}
}

@article{Takhtkeshha2024Multispectral,
  title={Multispectral light detection and ranging technology and applications: a review},
  author={Takhtkeshha, Narges and Mandlburger, Gottfried and Remondino, Fabio and Hyypp{\"a}, Juha},
  journal={Sensors},
  volume={24},
  number={5},
  pages={1669},
  year={2024},
  publisher={Multidisciplinary Digital Publishing Institute}
}

@inproceedings{Cai2020Monocular,
  title={{Monocular 3D Object Detection with Decoupled Structured Polygon Estimation and Height-Guided Depth Estimation}},
  author={Cai, Yingjie and Li, Buyu and Jiao, Zeyu and Li, Hongsheng and Zeng, Xingyu and Wang, Xiaogang},
  booktitle={Proceedings of the AAAI Conference on Artificial Intelligence},
  volume={34},
  number={07},
  pages={10478--10485},
  year={2020}
}

@inproceedings{Lourenço2021Intel,
  title={{Intel RealSense SR305, D415 and L515: Experimental Evaluation and Comparison of Depth Estimation}},
  author={Louren{\c{c}}o, Francisco and Araujo, Helder},
  booktitle={VISIGRAPP (4: VISAPP)},
  pages={362--369},
  year={2021}
}

@article{liasis2016satellite,
  title={{Satellite images analysis for shadow detection and building height estimation}},
  author={Liasis, Gregoris and Stavrou, Stavros},
  journal={ISPRS Journal of Photogrammetry and Remote Sensing},
  volume={119},
  pages={437--450},
  year={2016},
  publisher={Elsevier}
}

@article{liu2020im2elevation,
  title={{IM2ELEVATION: Building height estimation from single-view aerial imagery}},
  author={Liu, Chao-Jung and Krylov, Vladimir A and Kane, Paul and Kavanagh, Geraldine and Dahyot, Rozenn},
  journal={Remote Sensing},
  volume={12},
  number={17},
  pages={2719},
  year={2020},
  publisher={MDPI}
}

@article{lindberg2018umep,
title = {{Urban Multi-scale Environmental Predictor (UMEP): An integrated tool for city-based climate services}},
journal = {Environmental Modelling {\&} Software},
volume = {99},
pages = {70-87},
year = {2018},
issn = {1364-8152},
doi = {https://doi.org/10.1016/j.envsoft.2017.09.020},
url = {https://www.sciencedirect.com/science/article/pii/S1364815217304140},
author = {Fredrik Lindberg and C.S.B. Grimmond and Andrew Gabey and Bei Huang and Christoph W. Kent and Ting Sun and Natalie E. Theeuwes and Leena Järvi and Helen C. Ward and I. Capel-Timms and Yuanyong Chang and Per Jonsson and Niklas Krave and Dongwei Liu and D. Meyer and K. Frans G. Olofson and Jianguo Tan and Dag Wästberg and Lingbo Xue and Zhe Zhang},
}

@article{lobry2020rsvqa,
  title={{RSVQA: Visual question answering for remote sensing data}},
  author={Lobry, Sylvain and Marcos, Diego and Murray, Jesse and Tuia, Devis},
  journal={IEEE Transactions on Geoscience and Remote Sensing},
  volume={58},
  number={12},
  pages={8555--8566},
  year={2020},
  publisher={IEEE}
}

@article{li2024hrvqa,
  title={{HRVQA: A Visual Question Answering benchmark for high-resolution aerial images}},
  author={Li, Kun and Vosselman, George and Yang, Michael Ying},
  journal={ISPRS Journal of Photogrammetry and Remote Sensing},
  volume={214},
  pages={65--81},
  year={2024},
  publisher={Elsevier}
}

@article{li2024vrsbench,
  title={{VRSBench}: A versatile vision-language benchmark dataset for remote sensing image understanding},
  author={Li, Xiang and Ding, Jian and Elhoseiny, Mohamed},
  journal={Advances in Neural Information Processing Systems},
  volume={37},
  pages={3229--3242},
  year={2024}
}

@article{hu2025rsgpt,
  title={{RSGPT}: A {Remote} {Sensing} {Vision} {Language} {Model} and {Benchmark}},
  author={Hu, Yuan and Yuan, Jianlong and Wen, Congcong and Lu, Xiaonan and Liu, Yu and Li, Xiang},
  journal={ISPRS Journal of Photogrammetry and Remote Sensing},
  volume={224},
  pages={272--286},
  year={2025},
  publisher={Elsevier}
}

@article{wang2025disasterm3,
  title={{DisasterM3: A Remote Sensing Vision-Language Dataset for Disaster Damage Assessment and Response}},
  author={Wang, Junjue and Xuan, Weihao and Qi, Heli and Liu, Zhihao and Liu, Kunyi and Wu, Yuhan and Chen, Hongruixuan and Song, Jian and Xia, Junshi and Zheng, Zhuo and others},
  journal={arXiv preprint arXiv:2505.21089},
  year={2025}
}

@article{xuan2025dynamicvl,
  title={{DynamicVL: Benchmarking Multimodal Large Language Models for Dynamic City Understanding}},
  author={Xuan, Weihao and Wang, Junjue and Qi, Heli and Chen, Zihang and Zheng, Zhuo and Zhong, Yanfei and Xia, Junshi and Yokoya, Naoto},
  journal={arXiv preprint arXiv:2505.21076},
  year={2025}
}

@inproceedings{irvin2024teochat,
  title={{TEOChat}: A Large {Vision}-{Language} Assistant for Temporal Earth Observation Data},
  author={Irvin, Jeremy Andrew and Liu, Emily Ruoyu and Chen, Joyce Chuyi and Dormoy, Ines and Kim, Jinyoung and Khanna, Samar and Zheng, Zhuo and Ermon, Stefano},
  booktitle={International Conference on Learning Representations},
  year={2025}
}

@inproceedings{Kuckreja2024GeoChat,
  author = {Kuckreja, Kartik and Danish, Muhammad Sohail and Naseer, Muzammal and Das, Abhijit and Khan, Salman and Khan, Fahad Shahbaz},
  booktitle = {2024 {IEEE}/{CVF} {Conference} on {Computer} {Vision} and {Pattern} {Recognition} ({CVPR})},
  doi = {10.1109/cvpr52733.2024.02629},
  year = {2024},
  month = {jun 16},
  pages = {27831--27840},
  organization = {IEEE},
  title = {{GeoChat}: Grounded {Large} {Vision}-{Language} {Model} for {Remote} {Sensing}},
  url = {http://dx.doi.org/10.1109/CVPR52733.2024.02629},
}

@inproceedings{wang2024earthvqa,
  title         = {{EarthVQA: Towards queryable earth via relational reasoning-based remote sensing visual question answering}},
  author        = {Wang, Junjue and Zheng, Zhuo and Chen, Zihang and Ma, Ailong and Zhong, Yanfei},
  booktitle     = {{Proceedings of the AAAI Conference on Artificial Intelligence}},
  volume        = {38},
  number        = {6},
  pages         = {5481--5489},
  year          = {2024}
}

@InProceedings{Danish_2025_ICCV,
    author    = {Danish, Muhammad and Munir, Muhammad Akhtar and Shah, Syed Roshaan Ali and Kuckreja, Kartik and Khan, Fahad Shahbaz and Fraccaro, Paolo and Lacoste, Alexandre and Khan, Salman},
    title     = {{GEOBench-VLM}: Benchmarking {Vision}-{Language} Models for Geospatial Tasks},
    booktitle = {Proceedings of the IEEE/CVF International Conference on Computer Vision (ICCV)},
    month     = {October},
    year      = {2025},
    pages     = {7132-7142}
}

@article{mou2018im2height,
  title={{IM2HEIGHT: Height estimation from single monocular imagery via fully residual convolutional-deconvolutional network}},
  author={Mou, Lichao and Zhu, Xiao Xiang},
  journal={arXiv preprint arXiv:1802.10249},
  year={2018}
}

@article{liang2017automatic,
  title={{Automatic sky view factor estimation from street view photographs—A big data approach}},
  author={Liang, Jianming and Gong, Jianhua and Sun, Jun and Zhou, Jieping and Li, Wenhang and Li, Yi and Liu, Jin and Shen, Shen},
  journal={Remote Sensing},
  volume={9},
  number={5},
  pages={411},
  year={2017},
  publisher={MDPI}
}

@article{hodul2016estimation,
  title={{Estimation of continuous urban sky view factor from landsat data using shadow detection}},
  author={Hodul, Matus and Knudby, Anders and Ho, Hung Chak},
  journal={Remote Sensing},
  volume={8},
  number={7},
  pages={568},
  year={2016},
  publisher={MDPI}
}

@misc{baier2020geonrw,
  doi           = {10.21227/s5xq-b822},
  url           = {https://dx.doi.org/10.21227/s5xq-b822},
  author        = {Gerald Baier and Antonin Deschemps and Michael Schmitt and Naoto Yokoya},
  publisher     = {IEEE Dataport},
  title         = {{GeoNRW}},
  year          = {2020}
}

@article{Guth2021Digital,
  title={{Digital elevation models: Terminology and definitions}},
  author={Guth, Peter L and Van Niekerk, Adriaan and Grohmann, Carlos H and Muller, Jan-Peter and Hawker, Laurence and Florinsky, Igor V and Gesch, Dean and Reuter, Hannes I and Herrera-Cruz, Virginia and Riazanoff, Serge and others},
  journal={Remote Sensing},
  volume={13},
  number={18},
  pages={3581},
  year={2021},
  publisher={Multidisciplinary Digital Publishing Institute}
}

@article{Algancı2018Accuracy,
  title={Accuracy assessment of different digital surface models},
  author={Alganci, Ugur and Besol, Baris and Sertel, Elif},
  journal={ISPRS International Journal of Geo-Information},
  volume={7},
  number={3},
  pages={114},
  year={2018},
  publisher={MDPI}
}

@article{Jiao2019Evaluation,
title={Evaluation of Four Sky View Factor Algorithms Using Digital Surface and Elevation Model Data},
author={Zhonghu Jiao and H. Ren and X. Mu and J. Zhao and Tianxing Wang and Jiaji Dong},
journal={Earth and Space Science},
year={2019},
volume={6},
pages={222 - 237},
doi={10.1029/2018ea000475}}

@article{dirksen2019sky,
  title         = {{Sky view factor calculations and its application in urban heat island studies}},
  author        = {Dirksen, M and Ronda, RJ and Theeuwes, NE and Pagani, GA},
  journal       = {Urban Climate},
  volume        = {30},
  pages         = {100498},
  year          = {2019},
  publisher     = {Elsevier}
}

@article{miao2020review,
  title={{Review of methods used to estimate the sky view factor in urban street canyons}},
  author={Miao, Chunping and Yu, Shuai and Hu, Yuanman and Zhang, Huiwen and He, Xingyuan and Chen, Wei},
  journal={Building and Environment},
  volume={168},
  pages={106497},
  year={2020},
  publisher={Elsevier}
}

@article{Daramola2019Analysis,
title={Analysis of the urban surface thermal condition based on sky-view factor and vegetation cover},
author={M. Daramola and I. Balogun},
journal={Remote Sensing Applications: Society and Environment},
year={2019},
doi={10.1016/J.RSASE.2019.100253}}

@article{Xia2021Sky,
title={Sky view factor estimation from street view images based on semantic segmentation},
author={Yixi Xia and N. Yabuki and T. Fukuda},
journal={Urban Climate},
year={2021},
doi={10.1016/j.uclim.2021.100999}}

@article{bradley2001method,
  title={A method to assess the variation of urban canyon geometry from sky view factor transects},
  author={Bradley, AV and Thornes, JE and Chapman, L},
  journal={Atmospheric Science Letters},
  volume={2},
  number={1-4},
  pages={155--165},
  year={2001},
  publisher={Wiley Online Library}
}

@article{Jaehyun_Ha_2016,
  author        = {Jaehyun Ha and Sugie Lee and Cheolyeong Park},
  doi           = {10.3390/SU8090895},
  journal       = {Sustainability},
  number        = {9},
  publication_type = {article},
  title         = {{Temporal Effects of Environmental Characteristics on Urban Air Temperature: The Influence of the Sky View Factor}},
  url           = {https://www.mdpi.com/2071-1050/8/9/895/pdf},
  volume        = {8},
  year          = {2016}
}

@article{li2020two,
  title={{A two-level nested model for extracting positive and negative terrains combining morphology and visualization indicators}},
  author={Li, Jingxin and Zhang, Hongqi and Xu, Erqi},
  journal={Ecological Indicators},
  volume={109},
  pages={105842},
  year={2020},
  publisher={Elsevier}
}

@article{hoechstetter2008effects,
  title={{Effects of topography and surface roughness in analyses of landscape structure-A proposal to modify the existing set of landscape metrics}},
  author={Hoechstetter, Sebastian and Walz, Ulrich and Thinh, Nguyen Xuan and others},
  journal={Landscape Online},
  pages={3--3},
  year={2008}
}

@article{zhu2025internvl3,
  title={{InternVL3: Exploring advanced training and test-time recipes for open-source multimodal models}},
  author={Zhu, Jinguo and Wang, Weiyun and Chen, Zhe and Liu, Zhaoyang and Ye, Shenglong and Gu, Lixin and Tian, Hao and Duan, Yuchen and Su, Weijie and Shao, Jie and others},
  journal={arXiv preprint arXiv:2504.10479},
  year={2025}
}

@article{hurst2024gpt,
  title={{GPT-4o System Card}},
  author={Hurst, Aaron and Lerer, Adam and Goucher, Adam P and Perelman, Adam and Ramesh, Aditya and Clark, Aidan and Ostrow, AJ and Welihinda, Akila and Hayes, Alan and Radford, Alec and others},
  journal={arXiv preprint arXiv:2410.21276},
  year={2024}
}

@article{Jiao2023DilateFormer:,
  title={{DilateFormer}: Multi-Scale Dilated Transformer for Visual Recognition},
  author={Jiayu Jiao and Yuyao Tang and Kun-Li Channing Lin and Yipeng Gao and Jinhua Ma and Yaowei Wang and Wei-Shi Zheng},
  journal={IEEE Transactions on Multimedia},
  year={2023},
  volume={25},
  pages={8906-8919},
  doi={10.1109/TMM.2023.3243616}
}

@article{tang2025data,
  title={Data-efficient multi-scale fusion vision transformer},
  author={Tang, Hao and Liu, Dawei and Shen, Chengchao},
  journal={Pattern Recognition},
  volume={161},
  pages={111305},
  year={2025},
  publisher={Elsevier}
}

@article{Zhang2023Vision-Language,
  title={{Vision-Language Models for Vision Tasks: A Survey}},
  author={Jingyi Zhang and Jiaxing Huang and Sheng Jin and Shijian Lu},
  journal={IEEE Transactions on Pattern Analysis and Machine Intelligence},
  year={2023},
  volume={46},
  pages={5625-5644},
  doi={10.1109/TPAMI.2024.3369699}
}

@article{hou2023enhancing,
  title={{Enhancing monocular height estimation from aerial images with street-view images}},
  author={Hou, Xiaomou and Gan, Wanshui and Yokoya, Naoto},
  journal={arXiv preprint arXiv:2311.02121},
  year={2023}
}

@article{ouyang2025empirical,
  title={An empirical study of the non-determinism of chatgpt in code generation},
  author={Ouyang, Shuyin and Zhang, Jie M and Harman, Mark and Wang, Meng},
  journal={ACM Transactions on Software Engineering and Methodology},
  volume={34},
  number={2},
  pages={1--28},
  year={2025},
  publisher={ACM New York, NY}
}

@inproceedings{renze2024effect,
  title={The effect of sampling temperature on problem solving in large language models},
  author={Renze, Matthew},
  booktitle={Findings of the association for computational linguistics: EMNLP 2024},
  pages={7346--7356},
  year={2024}
}

@misc{scikitlearnJaccard_score,
	author = {},
	title = {{jaccard\_score --- scikit-learn.org}},
	howpublished = {\url{https://scikit-learn.org/stable/modules/generated/sklearn.metrics.jaccard_score.html}},
	year = {},
	note = {[Accessed 16-09-2025]},
}

@article{everingham2010pascal,
  title={{The pascal visual object classes (voc) challenge}},
  author={Everingham, Mark and Van Gool, Luc and Williams, Christopher KI and Winn, John and Zisserman, Andrew},
  journal={International journal of computer vision},
  volume={88},
  number={2},
  pages={303--338},
  year={2010},
  publisher={Springer}
}

@inproceedings{lin2014microsoft,
  title={{Microsoft} coco: Common objects in context},
  author={Lin, Tsung-Yi and Maire, Michael and Belongie, Serge and Hays, James and Perona, Pietro and Ramanan, Deva and Doll{\'a}r, Piotr and Zitnick, C Lawrence},
  booktitle={{European conference on computer vision}},
  pages={740--755},
  year={2014},
  organization={Springer}
}

@article{rodriguez2006global,
  title={{A global assessment of the SRTM performance}},
  author={Rodriguez, Ernesto and Morris, Charles S and Belz, J Eric},
  journal={Photogrammetric Engineering \& Remote Sensing},
  volume={72},
  number={3},
  pages={249--260},
  year={2006},
  publisher={American Society for Photogrammetry and Remote Sensing}
}

@article{lang2022global,
  title={Global canopy height regression and uncertainty estimation from {GEDI} {LIDAR} waveforms with deep ensembles},
  author={Lang, Nico and Kalischek, Nikolai and Armston, John and Schindler, Konrad and Dubayah, Ralph and Wegner, Jan Dirk},
  journal={Remote sensing of environment},
  volume={268},
  pages={112760},
  year={2022},
  publisher={Elsevier}
}

@article{hyndman2006another,
  title={{Another look at measures of forecast accuracy}},
  author={Hyndman, Rob J and Koehler, Anne B},
  journal={International journal of forecasting},
  volume={22},
  number={4},
  pages={679--688},
  year={2006},
  publisher={Elsevier}
}

@book{pont2023spacematrix,
  title={{Spacematrix: Space, Density and Urban Form-revised edition}},
  author={Pont, Meta Berghauser and Haupt, Per},
  year={2023},
  publisher={TU Delft OPEN Publishing}
}

@article{grimmond2001rapid,
  title={{Rapid methods to estimate sky-view factors applied to urban areas}},
  author={Grimmond, CSB and Potter, SK and Zutter, HN and Souch, C},
  journal={International Journal of Climatology: A Journal of the Royal Meteorological Society},
  volume={21},
  number={7},
  pages={903--913},
  year={2001},
  publisher={John Wiley \& Sons, Ltd. Chichester, UK}
}

@article{chen2012sky,
  title={Sky view factor analysis of street canyons and its implications for daytime intra-urban air temperature differentials in high-rise, high-density urban areas of {Hong Kong}: a {GIS}-based simulation approach},
  author={Chen, Liang and Ng, Edward and An, Xipo and Ren, Chao and Lee, Max and Wang, Una and He, Zhengjun},
  journal={International Journal of Climatology},
  volume={32},
  number={1},
  pages={121--136},
  year={2012},
  publisher={John Wiley \& Sons, Ltd. Chichester, UK}
}

@article{morabito2017urban,
  title={Urban imperviousness effects on summer surface temperatures nearby residential buildings in different urban zones of Parma},
  author={Morabito, Marco and Crisci, Alfonso and Georgiadis, Teodoro and Orlandini, Simone and Munaf{\`o}, Michele and Congedo, Luca and Rota, Patrizia and Zazzi, Michele},
  journal={Remote Sensing},
  volume={10},
  number={1},
  pages={26},
  year={2017},
  publisher={MDPI}
}

@article{bonafoni2018land,
  title={Land surface temperature and urban density: Multiyear modeling and relationship analysis using MODIS and Landsat data},
  author={Bonafoni, Stefania and Keeratikasikorn, Chaiyapon},
  journal={Remote Sensing},
  volume={10},
  number={9},
  pages={1471},
  year={2018},
  publisher={MDPI}
}

@article{krayenhoff2016daytime,
  title={Daytime thermal anisotropy of urban neighbourhoods: Morphological causation},
  author={Krayenhoff, E Scott and Voogt, James A},
  journal={Remote Sensing},
  volume={8},
  number={2},
  pages={108},
  year={2016},
  publisher={MDPI}
}

@article{yokoyama2002visualizing,
  title={{Visualizing topography by openness: A new application of image processing to digital elevation models}},
  author={Yokoyama, Ryuzo and Shirasawa, Michio and Pike, Richard J},
  journal={Photogrammetric engineering and remote sensing},
  volume={68},
  number={3},
  pages={257--266},
  year={2002},
  publisher={ASPRS AMERICAN SOCIETY FOR PHOTOGRAMMETRY AND}
}

@article{oke1981canyon,
  title={{Canyon geometry and the nocturnal urban heat island: comparison of scale model and field observations}},
  author={Oke, Tim R},
  journal={Journal of climatology},
  volume={1},
  number={3},
  pages={237--254},
  year={1981},
  publisher={Wiley Online Library}
}

@article{fisher1993algorithm,
  title={{Algorithm and implementation uncertainty in viewshed analysis}},
  author={Fisher, Peter F},
  journal={International Journal of Geographical Information Science},
  volume={7},
  number={4},
  pages={331--347},
  year={1993},
  publisher={Taylor \& Francis}
}

@article{ewing2009measuring,
  title={{Measuring the unmeasurable: Urban design qualities related to walkability}},
  author={Ewing, Reid and Handy, Susan},
  journal={Journal of Urban design},
  volume={14},
  number={1},
  pages={65--84},
  year={2009},
  publisher={Taylor \& Francis}
}

@article{stamps2005enclosure,
  title={{Enclosure and safety in urbanscapes}},
  author={Stamps III, Arthur E},
  journal={Environment and behavior},
  volume={37},
  number={1},
  pages={102--133},
  year={2005},
  publisher={Sage Publications Sage CA: Thousand Oaks, CA}
}

@article{Bohong_Zheng_2022,
  author        = {Bohong Zheng and Jiayu Li},
  doi           = {10.3390/buildings12060787},
  journal       = {Buildings},
  number        = {6},
  pages         = {787--787},
  publication_type = {article},
  title         = {{Evaluating the Annual Effect of the Sky View Factor on the Indoor Thermal Environment of Residential Buildings by ENVI-met}},
  url           = {https://www.mdpi.com/2075-5309/12/6/787/pdf?version=1654760717},
  volume        = {12},
  year          = {2022}
}

@article{Joseph_Appelbaum_2022,
  author        = {Joseph Appelbaum and A. Aronescu},
  doi           = {10.3390/en15228742},
  journal       = {Energies},
  number        = {22},
  pages         = {8742--8742},
  publication_type = {article},
  title         = {{View Factors of Flat Collectors, Including Photovoltaics, Visible to Partial Sky}},
  url           = {https://www.mdpi.com/1996-1073/15/22/8742/pdf?version=1669020955},
  volume        = {15},
  year          = {2022}
}

@article{zeng2018fast,
  title={{A fast approach for large-scale Sky View Factor estimation using street view images}},
  author={Zeng, Liyue and Lu, Jun and Li, Wuyan and Li, Yongcai},
  journal={Building and Environment},
  volume={135},
  pages={74--84},
  year={2018},
  publisher={Elsevier}
}

@article{gong2018mapping,
  title={{Mapping sky, tree, and building view factors of street canyons in a high-density urban environment}},
  author={Gong, Fang-Ying and Zeng, Zhao-Cheng and Zhang, Fan and Li, Xiaojiang and Ng, Edward and Norford, Leslie K},
  journal={Building and Environment},
  volume={134},
  pages={155--167},
  year={2018},
  publisher={Elsevier}
}

@article{pan2020novel,title={A Novel Rapid Method for Viewshed Computation on DEM through Max-Pooling and Min-Expected Height},author={Zhibin Pan and Jin Tang and T. Tjahjadi and Zhihu Wu and Xiaoming Xiao},journal={ISPRS Int. J. Geo Inf.},year={2020},volume={9},pages={633},doi={10.3390/ijgi9110633}}

@article{wu2023visibilitysurvey,
  title={{A survey of the landscape visibility analysis tools and technical improvements}},
  author={Wu, Zhiqiang and Wang, Yuankai and Gan, Wei and Zou, Yixuan and Dong, Wen and Zhou, Shiqi and Wang, Mo},
  journal={International Journal of Environmental Research and Public Health},
  volume={20},
  number={3},
  pages={1788},
  year={2023},
  publisher={MDPI}
}

@article{liu2023interpretable,
  title={{An interpretable machine learning framework for measuring urban perceptions from panoramic street view images}},
  author={Liu, Yunzhe and Chen, Meixu and Wang, Meihui and Huang, Jing and Thomas, Fisher and Rahimi, Kazem and Mamouei, Mohammad},
  journal={iScience},
  volume={26},
  number={3},
  year={2023},
  publisher={Elsevier}
}

@article{abarca2019urban,
  title={{Urban shape and built density metrics through the analysis of European urban fabrics using artificial intelligence}},
  author={Abarca-Alvarez, Francisco Javier and Campos-S{\'a}nchez, Francisco Sergio and Osuna-P{\'e}rez, Fernando},
  journal={Sustainability},
  volume={11},
  number={23},
  pages={6622},
  year={2019},
  publisher={MDPI}
}

@article{frank2013aesthetics,
  title={{Assessment of landscape aesthetics—Validation of a landscape metrics-based assessment by visual estimation of the scenic beauty}},
  author={Frank, S. and F{"u}rst, C. and Koschke, L. and Witt, A. and Makeschin, F.},
  journal={Ecological Indicators},
  volume={32},
  pages={222--231},
  year={2013}
}

@article{Usui2022Comparison,
  title={{Comparison of precise and approximated building height: Estimation from number of building storeys and spatial variations in the Tokyo metropolitan region}},
  author={Usui, Hiroyuki},
  journal={Environment and Planning B: Urban Analytics and City Science},
  volume={50},
  pages={487--499},
  year={2022},
  doi={10.1177/23998083221116117}
}
